%% file: main.tex
\documentclass[10pt,twocolumn,letterpaper]{article}

\usepackage[final]{main/cvpr}

\input{main/packages}

\definecolor{cvprblue}{rgb}{0.21,0.49,0.74}
\usepackage[pagebackref,breaklinks,colorlinks,citecolor=cvprblue]{hyperref}

\def\paperID{12774}
\def\confName{CVPR}
\def\confYear{2024}

\title{Open-vocabulary object 6D pose estimation}

\author{
\begin{minipage}{0.19\textwidth}
    \vspace*{-7mm}
    \centering
    Jaime Corsetti$^{1,2}$
\end{minipage}
\begin{minipage}{0.19\textwidth}
    \vspace*{-7mm}
    \centering
    Davide Boscaini$^1$
\end{minipage}
\begin{minipage}{0.19\textwidth}
    \vspace*{-6mm}
    \centering
    Changjae Oh$^3$
\end{minipage}
\begin{minipage}{0.22\textwidth}
    \vspace*{-7mm}
    \centering
    Andrea Cavallaro$^{4,5}$
\end{minipage}
\begin{minipage}{0.16\textwidth}
    \vspace*{-7mm}
    \centering
    Fabio Poiesi$^1$
\end{minipage} \\
\begin{minipage}[t]{0.17\textwidth}
    \hspace*{0.02cm} $^1$Fondazione \\ Bruno Kessler
\end{minipage}
\begin{minipage}[t]{0.13\textwidth}
    $^2$University \\ \hspace*{0.1cm} of Trento
\end{minipage}
\begin{minipage}[t]{0.27\textwidth}
    $^3$Queen Mary University \\ \hspace*{1cm} of London
\end{minipage}
\begin{minipage}[t]{0.26\textwidth}
    \vspace*{-0.1cm}
    $^4$Idiap Research Institute
\end{minipage}
\begin{minipage}[t]{0.10\textwidth}
    \vspace*{-0.1cm}
    $^5$EPFL
\end{minipage} \\
\begin{minipage}{0.36\textwidth}
    \vspace*{4mm}
    \centering
    \tt \small \{jcorsetti,dboscaini,poiesi\}@fbk.eu
\end{minipage}
\begin{minipage}{0.24\textwidth}
    \vspace*{4mm}
    \centering
    \tt\small c.oh@qmul.ac.uk
\end{minipage}
\begin{minipage}{0.28\textwidth}
    \vspace*{4mm}
    \centering
    \tt \small andrea.cavallaro@epfl.ch
\end{minipage}
}
\begin{document}

\include{main/commands}

\maketitle

\input{main/sections/0_abstract}
\input{main/sections/1_intro}

\input{main/sections/2_related}
\input{main/sections/3_method}

\input{main/sections/4_results}
\input{main/sections/5_conclusion}

{
    \small
    \bibliographystyle{main/ieeenat_fullname}
    \bibliography{main}
}


\end{document}


\include{supp/commands}

\maketitle

\input{supp/sections/0_introduction}
\input{supp/sections/1_comparison}
\input{supp/sections/2_details}
\input{supp/sections/3_dataset}

\input{supp/sections/4_metrics}
\input{supp/sections/5_prompts}
\input{supp/sections/6_qualitative}

{
    \small
    \bibliographystyle{supp/ieeenat_fullname}
    \bibliography{supp}
}


%% file: main/packages.tex
\usepackage[dvipsnames]{xcolor}
\usepackage{amsmath,amssymb,amsfonts}
\usepackage{multirow,multicol}
\usepackage{natbib}
\usepackage{algorithmic}
\usepackage{graphicx}
\usepackage{textcomp}
\usepackage{xspace}
\usepackage{url}
\usepackage{dsfont}
\usepackage{booktabs}
\usepackage{balance}
\usepackage{soul}
\usepackage{todonotes}
\usepackage{pifont}
\usepackage{colortbl}
\usepackage{overpic}
\usepackage{enumitem}
\usepackage{nccmath}
\usepackage{courier}
\usepackage{array}
\usepackage{tabularray}
\usepackage{pifont}
\usepackage[accsupp]{axessibility} 

%% file: main/commands.tex
\newcommand{\fabiocomment}[1]{\todo[color=purple!20, inline, author=Fabio]{#1}}
\newcommand{\davidecomment}[1]{\todo[color=blue!20, inline, author=Davide]{#1}}

\newcommand{\fabio}[1]{\textbf{\textcolor{purple!75}{#1}}}
\newcommand{\davide}[1]{\textbf{\textcolor{blue!75}{#1}}}
\newcommand{\changjae}[1]{\textbf{\textcolor{olive!75}{#1}}}
\newcommand{\andrea}[1]{\textbf{\textcolor{orange!75}{#1}}}
\newcommand{\warning}[1]{\textbf{\textcolor{red!75}{#1}}}

\newcommand{\imgencoder}[0]{$\phi_V$\xspace}
\newcommand{\textencoder}[0]{$\phi_T$\xspace}
\newcommand{\guidance}[0]{$\phi_G$\xspace}
\newcommand{\decoder}[0]{$\phi_D$\xspace}
\newcommand{\fusion}[0]{$\phi_{TV}$\xspace}

\newcommand{\learnedprompts}[0]{$\mathbf{D}$\xspace}
\newcommand{\learnedembs}[0]{$\mathbf{E}^D$\xspace}

\newcommand{\clipfeats}[1]{$\mathbf{E}^{#1}$\xspace}
\newcommand{\costvolume}[1]{$\mathbf{V}^{#1}$\xspace}
\newcommand{\costemb}[1]{$\mathbf{C}^{#1}$\xspace}
\newcommand{\finalfeats}[1]{$\mathbf{F}^{#1}$\xspace}
\newcommand{\promptemb}[0]{$\mathbf{e}^T$\xspace}

\newcommand{\onedim}[1]{$\in \mathbb{R}^{#1}$\xspace}
\newcommand{\twodim}[2]{$\in \mathbb{R}^{#1\times #2}$\xspace}
\newcommand{\threedim}[3]{$\in \mathbb{R}^{#1\times #2\times #3}$\xspace}

\newcommand{\rgb}[1]{RGB$^{#1}$\xspace}
\newcommand{\depth}[1]{D$^{#1}$\xspace}
\newcommand{\pcd}[1]{P$^{#1}$\xspace}
\newcommand{\predmask}[1]{$\mathbf{M}^{#1}$\xspace}
\newcommand{\query}[0]{$Q$\xspace}
\newcommand{\anchor}[0]{$A$\xspace}
\newcommand{\object}[0]{$O$\xspace}
\newcommand{\prompt}[0]{$T$\xspace}

\newcommand{\objectpcd}[0]{$\mathbf{O}$\xspace}
\newcommand{\predpose}[0]{$\mathbf{T}$\xspace}
\newcommand{\gtpose}[0]{$\mathbf{\hat{T}}$\xspace}

\newcommand{\lowerbetter}[0]{{\color{black!50}{$\,\downarrow$}}}
\newcommand{\higherbetter}[0]{{\color{black!50}{$\,\uparrow$}}}
\newcommand{\oracle}[1]{\textcolor{gray}{#1}}

\definecolor{visual}{HTML}{3399FF}
\definecolor{text}{HTML}{97D077}

\definecolor{nocsbottle}{RGB}{31,119,180}
\definecolor{nocsbowl}{RGB}{255,127,14}
\definecolor{nocscamera}{RGB}{44,160,44}
\definecolor{nocscan}{RGB}{214,39,40}
\definecolor{nocslaptop}{RGB}{148,103,189}
\definecolor{nocsmug}{RGB}{140,86,75}
\newcommand{\impp}[1]{{\textcolor{Green}{+#1}}}
\newcommand{\impn}[1]{{\textcolor{BrickRed}{-#1}}}

\newcommand{\correct}[1]{{\textcolor{Green}{#1}}}
\newcommand{\wrong}[1]{{\textcolor{BrickRed}{#1}}}

\newcommand{\acronym}{Oryon\xspace}
\definecolor{myazure}{rgb}{0.8509,0.8980,0.9412}
\newcommand{\cmark}{\ding{51}}%
\newcommand{\xmark}{\ding{55}}%

\newcommand{\captionprompt}[3]{
    \centering \texttt{#1} \\
    \centering \texttt{\correct{#2}} \\
    \centering \texttt{\wrong{#3}}
}

%% file: main/sections/0_abstract.tex
\begin{abstract}
We introduce the new setting of open-vocabulary object 6D pose estimation, in which a textual prompt is used to specify the object of interest.
In contrast to existing approaches, in our setting
(i) the object of interest is specified solely through the textual prompt,
(ii) no object model (e.g.,~CAD or video sequence) is required at inference, and
(iii) the object is imaged from two RGBD viewpoints of different scenes.
To operate in this setting, we introduce a novel approach that leverages a Vision-Language Model to segment the object of interest from the scenes and to estimate its relative 6D pose.
The key of our approach is a carefully devised strategy to fuse object-level information provided by the prompt with local image features, resulting in a feature space that can generalize to novel concepts.
We validate our approach on a new benchmark based on two popular datasets, REAL275 and Toyota-Light, which collectively encompass 34 object instances appearing in four thousand image pairs. 
The results demonstrate that our approach outperforms both a well-established hand-crafted method and a recent deep learning-based baseline in estimating the relative 6D pose of objects in different scenes.
Code and dataset are available at \url{https://jcorsetti.github.io/oryon}.
\end{abstract}

%% file: main/sections/1_intro.tex
\input{main/figures/teaser}

\section{Introduction}\label{sec:intro}

Accurate 6D pose estimation is fundamental for a wide range of applications, such as robot manipulation~\cite{grasping1}, autonomous driving~\cite{autodriving1} and augmented reality~\cite{augreality}. 
While recent works achieve high degree of pose estimation accuracy, even in cluttered scenes~\cite{geometricaware6d, e2ek, fcgf6d} and with challenging symmetric objects~\cite{surfemb}, most are \textit{instance-level} pose estimation models, which are trained and tested with the same set of objects~\cite{zebrapose,geometricaware6d}.
\textit{Generalizable} or \textit{one-shot}~\cite{ove6d, onepose, osop} pose estimation methods remove the above limitation by training a model on a diverse set of objects using recent large-scale datasets (e.g.~ShapeNet6D~\cite{he2022fs6d} and MegaPose~\cite{labbe2022megapose}), and test on novel objects without constraints about shape or category.  
However, \textit{model-based} methods still require a model of the novel object at test time~\cite{osop,ove6d,labbe2022megapose}, while methods defined as \textit{model-free} require a set of reference views~\cite{xiao2022fewshot,kuznetsova2016exploiting} or a video~\cite{gen6d,onepose,oneposepp} of the novel object, and additional preprocessing. 
Such methods are unsuitable when novel objects are not physically available, as a video sequence cannot be acquired.

In this paper, we introduce a new 6D pose estimation formulation that changes the assumptions of previous approaches (Fig.~\ref{fig:teaser}), by using a textual prompt to identify the object of interest.
To accomplish this, we integrate in the architecture a Vision-Language Model (VLM) to identify the object of interest from the two scenes and to estimate its relative 6D pose.
We name our approach \textit{\acronym} (\textbf{O}pen-vocabula\textbf{ry} \textbf{o}bject pose estimatio\textbf{n}).
The proposed new formulation uses textual information (prompt) provided by the user to not only localize the object of interest in a cluttered scene, but also to guide our VLM to focus on points which are more specific to the target object.
We define this formulation as \textit{Open-Vocabulary} as we do not put any constraint on the input prompt.
While previous works use language in related tasks~\cite{language6dpose}, their contribution is limited to the localization of the object of interest.
Instead, we show that textual prompts can provide rich semantic information to guide the VLM, and are fundamental to the generalization capabilities of our method. 
This setting can be applied to different scenarios.
We focus on the \textit{cross-scene} setting, in which the two RGBD images show different scenes, with one or possibly more objects in common.
We validate \acronym on a new benchmark that is based on two popular datasets, namely REAL275~\cite{nocs} and Toyota-Light~\cite{toyl}.
The first shows an high variation of object poses and scenes with mild occlusions, while the second presents challenging light conditions.
We compare \acronym against a well-established hand-crafted feature extraction method, i.e.,~SIFT~\cite{lowe1999sift}, and a recent deep learning approach specifically designed for registration of point clouds with low overlap, i.e.,~ObjectMatch~\cite{gumeli2023objectmatch}.
\acronym outperforms both SIFT and ObjectMatch on both datasets.
We carry out an extensive ablation study to validate the different components.
In summary, our contributions are as follows:
\begin{itemize}
    \item We introduce a novel setting in object 6D pose estimation, featuring a new set of assumptions. This includes specifying the object of interest via a textual prompt, as opposed to relying on a 3D model or a video sequence;
    \item We propose an architecture based on a Vision-Language Model, capable of segmenting and producing local distinctive features for object matching;
    \item We establish a new benchmark based on two popular object 6D pose estimation datasets, REAL275~\cite{nocs} and Toyota-Light~\cite{toyl}. The benchmark focuses on determining the relative pose of a text-prompted object captured from two different scenes.
\end{itemize}

%% file: main/figures/teaser.tex
\begin{figure}[t]
    \centering
    \begin{overpic}[width=\columnwidth]{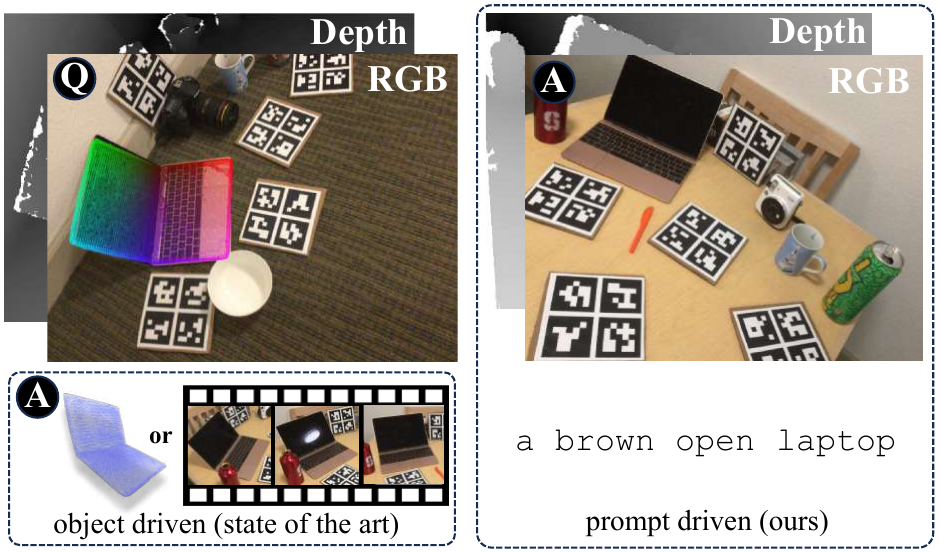}
    \end{overpic}
    \vspace{-6mm}
    \caption{
    Our open-vocabulary setting enables the estimation of the pose of an object captured in two distinct scenes. 
    State-of-the-art generalizable 6D pose estimation methods from RGB on RGBD images typically depend on the object CAD model~\cite{ove6d} or a video sequence of the object at test time~\cite{gen6d,onepose} as a reference (\anchor) to compute the object pose in the query image (\query). 
    In contrast, our method uses a textual prompt to guide the pose estimation process, and requires a single view as reference.}
    \label{fig:teaser}
\end{figure}

%% file: main/sections/2_related.tex
\section{Related work}\label{sec:related}

\noindent\textbf{Model-free pose estimation of novel objects.}~Gen6D~\cite{gen6d} represents one of the first approach to tackle the pose estimation of novel objects without CAD models, using only RGB images. 
Gen6D implements an object-agnostic detector, a viewpoint selector, and a pose refinement step. 
During evaluation, Gen6D requires the acquisition of an RGB video sequence of the novel object, upon which a Structure-from-Motion method (COLMAP~\cite{schoenberger2016sfm}) is applied to recover the camera pose for each frame. 
However, a significant limitation of Gen6D is the complexity of its evaluation procedure, which also requires manual cropping and orientation of the reconstructed point cloud.
Similarly, OnePose~\cite{onepose} demands an RGB video sequence of the novel object and a comparable procedure at test time.
OnePose++~\cite{oneposepp} extends OnePose to address pose estimation for low-textured objects, but maintains the same test-time requirements as OnePose.
In contrast, \acronym eliminates the need for SfM at test time, and only requires a textual description of the object, which can be effortlessly provided by a user without technical expertise.

\noindent\textbf{Relative pose estimation}~aims to estimate the pose of an object in a scene (\textit{query}) with respect to a reference view (\textit{anchor})~\cite{nguyen2023nope, zhang2022relpose, lin2023relpose++,park2020latentfusion}. 
NOPE~\cite{nguyen2023nope} trains an autoencoder to learn a representation conditioned on the relative orientation of the query image with the reference image. 
At test time, a set of reference features are produced, each associated with an orientation. 
The orientation of the reference feature more similar to the query one is chosen.
Similarly to \acronym, NOPE is evaluated on novel objects and does not use object models. 
However, NOPE only predicts the relative rotation and has not been tested in real-world scenarios.
LatentFusion~\cite{park2020latentfusion} shares a similar setting, but by adopting RGBD scenes instead of RGB only is capable of estimating also the translation component.
A similar task is 6D pose tracking, where several model-free methods have been recently proposed~\cite{bundletrack,wen2023bundlesdf,nguyen2022pizza}. 
In this setting, the relative pose is estimated between pairs of consecutive frames. 
Methods like BundleTrack~\cite{bundletrack} and BundleSDF~\cite{wen2023bundlesdf} rely on the information provided by previous frames by storing their representation in a memory pool. 
Instead, our reference is a single viewpoint image.
Moreover, the segmentation module allows \acronym to tackle the case in which the query and reference frames are acquired in different scenes.
See the Supplementary Material for an extended discussion of relative pose estimation methods.

\noindent\textbf{Point cloud registration.}~A related problem to RGBD-based 6D pose estimation is point cloud registration.
This task is commonly addressed by using specific methods for feature extraction~\cite{detone2018superpoint, FCGF, gedi, fcgf6d, dip} paired by matching and registration, or by end-to-end optimization~\cite{Xu2021omnet, yew2022regtr}.
A recent advancement on this field is ObjectMatch~\cite{gumeli2023objectmatch}, which tackles the problem of registration of two scene fragments captured from two 3D viewpoints with low overlap.
This is achieved by implicitly computing object matches through the detection of objects in the scenes.
Our scenario is significantly more challenging as the only overlapping parts of the scenes are the ones belonging to the object of interest.
Therefore, our problem could be considered as a particular case of point cloud registration, in which the point cloud parts to be aligned are described by the textual prompt.

%% file: main/sections/3_method.tex
\section{Our approach}
\label{sec:method}
\input{main/figures/maindiagram}

\subsection{Overview}\label{sec:overview}

The new setting is defined as follows:
(i) the object of interest is specified solely through textual prompt (no object model or video sequence is required);
(ii) the object is imaged from two different viewpoints of two different scenes; and
(iii) the object was not observed during the training phase.
From these requirements, we formulate the problem as a relative pose estimation task~\cite{nguyen2023nope, zhang2022relpose} between two scenes, the anchor \anchor and the query \query, which are represented as RGB image \rgb{A} (\rgb{Q}) and depth map \depth{A} (\depth{Q}).
To match \anchor and \query we propose \acronym, which relies on a fusion module based on cost-aggregation that relates the information of the textual prompt to the local visual feature map. 
The model generalizes to unseen objects and is trained on a large dataset of object models annotated with brief textual descriptions~\cite{he2022fs6d}.

We address the localization problem jointly with the matching task, by predicting a segmentation mask on each view (scene) to locate the object of interest. 
The final features  encode {\em semantic} information about the objects for the segmentation task and  {\em geometric} information, as the pose is derived from the relative position of keypoints.
At test time, we use the predicted masks to select points on the object of interest and match pairs of features by nearest neighbor. 
The resulting matches are projected back to the 3D domain to obtain the final relative object pose across the two scenes through point cloud registration~\cite{pointdsc}.

Fig.~\ref{fig:pipeline} shows \acronym's architecture.
We describe the object of interest with a user-provided textual prompt \prompt. 
We use \prompt to guide the extraction of keypoints and to compute the matches, as only the points which match the textual description provided by the prompt itself are considered through the use of the mask.
Given a pair of scenes (\anchor, \query), we compute a set of coordinates $\mathbf{x}^A$ and $\mathbf{x}^Q$ that form a set of matches between scenes (i.e., ~they describe the most similar locations in color and 3D structure), which are filtered by the predicted masks \predmask{A}, \predmask{Q}.

In order to jointly process \prompt and (\anchor, \query), their representations should share the same feature space.
We achieve this by leveraging a VLM trained for semantic alignment, such as CLIP~\cite{clip}.
We process \rgb{A} and \rgb{Q} with the CLIP image encoder \imgencoder, and extract the features before the final pooling layer.
Let (\clipfeats{A}, \clipfeats{Q}) \threedim{D}{H}{H} be a pair of feature maps extracted from $A$ and $Q$, where $D$ is the feature dimension, and $H$ is the spatial dimension of the feature map.
$D$ and $H$ depend on the CLIP encoder used.
Note that \rgb{A} and \rgb{Q} are processed without any prior crop, unlike other pose estimation approaches which adopt prior detectors or segmenters~\cite{gen6d, onepose, oneposepp, geometricaware6d, zebrapose}.
\prompt is encoded by the CLIP text encoder \textencoder.
To provide a rich set of representations, we adopt a set of templates to generate $N$ versions of \prompt~\cite{clip,cho2023catseg}, which are encoded by \textencoder.
Hence, we obtain the prompt features \promptemb \twodim{N}{D}, where $N$ is the number of templates.
Both CLIP encoders are frozen.

\subsection{Text-visual fusion}
\label{sec:costaggregation}

The objective of the text-visual module \fusion is to provide a visual representation which is semantically consistent with the represented object (and therefore influenced by \promptemb), but which is also representative of the object local appearance in \clipfeats{A} and \clipfeats{Q}.
The first requirement is fundamental for the segmentation task, while the second is needed to perform image matching. 
We implement the fusion module \fusion by building a cost volume, i.e.~by computing the cosine similarity among each feature map location in \clipfeats{A}, \clipfeats{Q} and \promptemb, thus obtaining a pair of matrices of shape $\mathbb{R}^{N\times H \times H}$.
The cost volume represents the cost of associating each prompt feature to each location in the visual feature maps.

Note that the cost volume obtained does not enforce a spatial consistency between neighboring patches.
In order to let the model learn these relations and refine the cost volume, we adopt a cost aggregation block based on two Transformer layers~\cite{liu2021swin}.
Both layers rely on self-attention to model the relationship between the image patches, with different modalities.
The first applies self-attention within the same window, while the second applies self-attention among patches of shifted windows.
This allows the model to perform attention both within the patch and between neighboring patches.
Objects can have different shapes and sizes in the image, therefore only applying local attention can not be sufficient to build an effective representation.
To enrich the representation provided by CLIP, we adopt guidance features~\cite{cho2023catseg} from another backbone \guidance which are concatenated to the query and keys of the Transformer layers.
The output of the fusion module is a pair of cost features \costemb{A}, \costemb{Q} \threedim{D}{H}{H}.

\subsection{Decoding}\label{sec:decoding}

The feature maps obtained in the previous step have the original low resolution of the CLIP feature map (24$\times$24 in our case).
As we pursue an image matching objective, we require a higher resolution to compute fine-grained matches. 
To this end, we adopt a decoder \decoder composed by three upsampling layers.
Note that CLIP was trained for semantic alignment of the global feature, and this property partially transfers to its visual feature map~\cite{zhou2022maskclip}.
However, in our setting we also require information based on the appearance of the object, as opposed to semantic information only.
Therefore, we find beneficial to add the guidance features of \guidance in the decoder.
Specifically, the two feature maps obtained by the guidance network are projected and concatenated to the input feature map before each upsampling layer~\cite{cho2023catseg}. 
The final layer does not use any guidance.

The resulting feature maps \finalfeats{A}, \finalfeats{Q} are used both for computing the matches and for the segmentation task.
For the latter, we add a segmentation head to compute the activations and output a pair of binary masks \predmask{A}, \predmask{Q}.

\subsection{Optimization}\label{sec:optimization}

Our formulation implies two optimization objectives.
For effective image matching, we directly optimize the feature maps by forcing low similarity between dissimilar locations and high similarity between similar locations across \anchor and \query.
As supervision, we adopt the ground-truth matches between \anchor and \query (i.e.~pair of points on \anchor and  \query which belong to the same portion of the object of interest).
In practice, we adopt a contrastive loss $\ell_F$ with hardest negative mining~\cite{FCGF,fcgf6d}.
There are two components: the first promotes the corresponding features (i.e.~the matches) to be close in the feature space, while the second increases the distance between a feature and its \emph{hardest negative}.
Given the pair of scenes \anchor, \query, the set of positive pairs is defined as $\mathcal{P} = \{ (i, j) \colon \textbf{x}^A_i \in \;\anchor, \textbf{x}^Q_j \in \;\query, \phi(\textbf{x}^A_i) = \textbf{x}^Q_j \}$, where $\phi \colon A \to Q $ is a match mapping between \anchor and \query pixels. 
We define $\mathbf{f}^A$, $\mathbf{f}^Q$ \twodim{C}{D} as the set of features extracted respectively from the feature maps \finalfeats{A}, \finalfeats{Q}, by using the ground-truth matches $\mathcal{P}$.
$C = \lvert \mathcal{P}\rvert$ is the total number of matches and $D$ is the feature dimension. 
The positive loss component $\ell_P$ is defined as
\begin{equation}\label{eq:hcpos}
\ell_P = \sum_{(i, j) \in \mathcal{P}} \frac{1}{\lvert \mathcal{P} \rvert} \left( \mathrm{dist} ( \textbf{f}^A_i, \textbf{f}^Q_j) - \mu_P \right)_+,
\end{equation}
where $\mu_P$ is a positive margin and $( \cdot )_+ = \max(0, \cdot)$.
The positive margin is an hyperparameter, and represents the desired distance in the feature space of a positive pair. 

To define negative pairs, we consider a set of features $\mathbf{f}$ and their corresponding 2D coordinates $\textbf{x}$ on the image. 
Given a single feature $\mathbf{f}_i$, we define its set of candidate negatives as $\mathcal{N}_i = \{ k \colon \textbf{x}_k \in \textbf{x}, k \ne i,  \lVert\textbf{x}_i - \textbf{x}_k\rVert\ \geq \tau\}$.
Note that this set excludes all the points which are closer than the distance $\tau$ from the reference point $\textbf{x}_i$ in the image, so that features describing the same points are not considered.
We compute the candidate negatives set for all points of $\mathbf{x}^A$ and $\mathbf{x}^Q$, and define the negative loss component $\ell_N$ as
\begin{equation}\label{eq:hcneg}
\begin{aligned}
\ell_N = \sum_{(i, j) \in \mathcal{P}} & \frac{1}{2 \lvert \mathcal{P}_i \rvert} \left( \mu_N - \min_{k \in \mathcal{N}_i} \mathrm{dist}(\textbf{f}_i, \textbf{f}_k) \right)_+ \\
& \hspace{-4.5mm}+ \frac{1}{2 \lvert \mathcal{P}_j \rvert} \left( \mu_N - \min_{k \in \mathcal{N}_j} \mathrm{dist}(\textbf{f}_j, \textbf{f}_k) \right)_+.
\end{aligned}
\end{equation}

For each $\mathbf{f}_i$, the nearest $\mathbf{f}_k$ in the feature space (i.e.~the hardest negative) is selected.
The negative margin $\mu_N$ is an hyperparameter, which defines the desired distance in the feature space of a negative pair.
In Eqs.~\eqref{eq:hcpos} and \eqref{eq:hcneg}, $\mathrm{dist}( \cdot )$ is the inverted and normalized cosine similarity.
The feature loss $\ell_F$ is thus defined as 
\begin{equation}
    \ell_F = \lambda_N \ell_N + \lambda_{P} \ell_{P},
\end{equation}
\noindent where $\lambda_N$ and $\lambda_P$ are hyperparameters.

We implement the segmentation loss $\ell_M$ as a Dice loss~\cite{sudre2017diceloss}.
We found the Dice loss to be more effective than cross entropy, as the first is specifically designed to handle high class imbalance in the foreground mask.
This is consistent with our scenario, in which the object of interest may occupy a small portion of the image.
The final loss $\ell$ is defined as 

\begin{equation}
    \ell = \lambda_M \ell_M + \ell_{F},
\end{equation}

\noindent where the weight factor $\lambda_{M}$ is an hyperparameter.

\subsection{Matching and registration}
At test time, the predicted masks \predmask{A}, \predmask{Q} are used to retain the features of \finalfeats{A}, \finalfeats{Q} belonging to the objects, thus obtaining two lists $\mathbf{F}^A_M$ \twodim{C^1}{D}, $\mathbf{F}^Q_M$ \twodim{C^2}{D}.
For each feature $\mathbf{f}^A_i \in \mathbf{F}^A_M$, we compute the nearest neighbor $\mathbf{f}^Q_i \in \mathbf{F}^Q_M$ in the feature space. 
We reject the pairs $\mathbf{f}^A_i, \mathbf{f}^Q_i$ for which $\mathrm{dist}(\mathbf{f}^A_i, \mathbf{f}^Q_i) > \mu_t$. 
We select all points belonging to a match and back-project them to 3D, by using the depth maps and the intrinsic camera parameters of $A$ and $Q$, thus obtaining two point clouds $\mathbf{P}^A, \mathbf{P}^Q$ \twodim{C}{3}.

The pose $T_{A\rightarrow Q}$ is estimated with a point cloud registration algorithm.
Due to noise in the predicted masks and possible ambiguity in the prompt, we expect spurious matches to be present.
We use PointDSC~\cite{pointdsc} for its robust matching capabilities based on spatial consistency, which allows it to reject inconsistent matches and provide precise poses.

%% file: main/figures/maindiagram.tex
\begin{figure*}[t]
    \centering
    \begin{overpic}[width=1\textwidth]{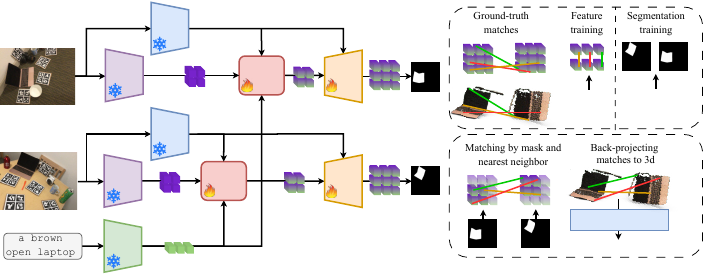}
    \put(5,19){\anchor}
    \put(5,33.5){\query}
    \put(5,7){\prompt}
    
    \put(16,13.5){\imgencoder}
    \put(16,28.5){\imgencoder}
    \put(16,4.5){\textencoder}

    \put(26.5,25.1){\clipfeats{Q}}
    \put(23,10.3){\clipfeats{A}}
    \put(24,1.5){\promptemb}

    \put(22.5,20){\guidance}
    \put(22.5,35){\guidance}
    
    \put(35.5,28.5){\fusion}
    \put(30,13.5){\fusion}
    
    \put(47.5,28.5){\decoder}
    \put(47.5,13.5){\decoder}

    \put(42,25.1){\costemb{Q}}
    \put(40.5,10.3){\costemb{A}}

    \put(53,10){\finalfeats{A}}
    \put(58,10){\predmask{A}}
    \put(53,24.5){\finalfeats{Q}}
    \put(58,24.5){\predmask{Q}}

    \put(82.5,7.9){\small{Registration}}
    \put(84,4.2){$T_{A \rightarrow Q}$}
    
    \put(82,24.5){$\ell_{F}$}    
    \put(91,24.5){$\ell_{M}$}

    \put(96,29){\rotatebox{270}{Training}}
    \put(96,10){\rotatebox{270}{Testing}}
                
    \end{overpic}
    \vspace{-7mm}
    \caption{
    The training pipeline of \acronym can be separated in four stages. 
    In the first stage, the pair of images \anchor, \query are encoded by the CLIP image encoder \imgencoder, while the prompt \prompt is encoded by the CLIP text encoder \textencoder.
    The guidance network \guidance is used to produce a rich visual representation which is used in the next stages.
    The resulting visual feature maps \clipfeats{A}, \clipfeats{Q} and text features \promptemb are used in the second stage in the fusion module \fusion.
    This outputs a pair of cost features \costemb{A},\costemb{Q} which in the third stage are upsampled to the final feature maps \finalfeats{A}, \finalfeats{Q}.
    The same feature maps are fed to a segmentation head to obtain the predicted masks \predmask{A}, \predmask{Q}.
    At train time, \finalfeats{A}, \finalfeats{Q} are optimized by a contrastive loss, while a dice loss supervises the training of the segmentation masks.
    At test time, the predicted masks are used to filter \finalfeats{A}, \finalfeats{Q}, and matches are obtained by nearest neighbor.
    Finally, the matches are projected back to the 3D domain, and a registration algorithm is used to retrieve the final pose $T_{A \rightarrow{Q}}$.
    }
    \label{fig:pipeline}
\end{figure*}

%% file: main/sections/4_results.tex
\section{Results}\label{sec:exps}

\subsection{Experimental setup}
We train our model for 20 epochs, and adopt CLIP ViT-L/14~\cite{clip} as visual and text encoder.
We use Adam~\cite{adam} with learning rate $10^{-4}$, weight decay $5\cdot10^{-4}$, and a cosine annealing scheduler~\cite{cosine} to lower the learning rate to $10^{-5}$.
Loss weights are set as $\lambda_P = \lambda_N = $ 0.5 and $\lambda_{M} = $ 1.0. 
As guidance backbone, we adopt a pretrained Swin Transformer~\cite{liu2021swin}.
As augmentations we randomly apply color jittering, horizontal flipping and vertical flipping.

We set the positive and negative margins in Eqs.~\eqref{eq:hcpos} and \eqref{eq:hcneg} as $\mu_{P}$ = 0.2 and $\mu_{N}$ = 0.9, and the excluding distance for hardest negatives as $\tau$ = 5.
At test time we set $\mu_t$ = 0.25 as maximum feature distance to identify a match, and limit the match number to $C$ = 500.
The output resolution of \finalfeats{A}, \finalfeats{Q} is $192\times192$, and the feature dimension is $F$ = 32.

\subsection{Datasets}

We use the synthetic dataset ShapeNet6D~\cite{he2022fs6d} for training, as it provides several diverse objects and scenes.
We evaluate \acronym on two real-world datasets: REAL275~\cite{nocs} and Toyota-Light~\cite{toyl}.
The standard prompt \prompt we adopt at test time is composed by the object name (e.g., ``laptop'') preceded by a brief description (e.g., ``brown open'').
For REAL275 and TOYL, we manually annotate each object instance with a textual prompt, while for ShapeNet6D we rely on its metadata.
For all datasets, we train and test on a set of scene pairs of images from the original datasets.

\noindent\textbf{ShapeNet6D} (SN6D)~\cite{he2022fs6d} is a large-scale synthetic dataset built by using ShapeNetSem~\cite{shapenetsem} object models.
For each object model, ShapeNetSem includes a name and a set of synonyms, which we use to build the textual prompt.
Note that in this dataset the prompt is only composed by the object name, as no description is provided.
To enrich the learned representations, during training we randomly substitute the object name in the prompt with one of the synonyms provided in the metadata (e.g., possible synonyms for ``television'' are ``tv'' and ``telly'').
Note that ShapeNet6D does not contain the object instances present in the two test datasets, although it may contain objects of similar category.
We train on 20K image pairs from SN6D.

\noindent\textbf{REAL275}~\cite{nocs} provides a set of RGBD images in different scenes, with a total of six object categories and three instances for each category.
REAL275 provides an high variety of viewpoints between the scenes, and also present mild occlusions.
We evaluate on 2K image pairs from the original real-world test set.

\noindent\textbf{Toyota-Light} (TOYL)~\cite{toyl} focuses on pose estimation under challenging light conditions, in which a single object is present in each scene.
This is relevant in our setting, as we process pairs of images: a significant difference in light across the two scenes poses an important challenge in the image matching task.
We evaluate on 2K image pairs.

\subsection{Evaluation metrics}
\label{sec:metrics}
We evaluate all our pose results by using the metrics proposed for the BOP Benchmark~\cite{bop-challenge}, namely AR (Average Recall), which is the average of VSD, MSSD and MSPD.
We also report ADD(S)-0.1d, as it is typically used in 6D pose estimation~\cite{lm, ffb6d, gen6d}.
ADD(S)-0.1d is a recall on the pose error: it considers a success when the error is less than one tenth of the object diameter.
Instead, VSD, MSSD and MSPD are averages of recalls on multiple thresholds.
These latter metrics are more suitable to represent performance in this challenging scenario.
Therefore, we adopt AR as main metric.
See the Supplementary Material for an extended discussion of the pose metrics.
For all experiments we also report the mean Intersection-Over-Union (mIoU) to quantify the predicted mask quality~\cite{cho2023catseg, zhou2022maskclip}.

\subsection{Comparison procedure}\label{sec:procedure}

Our setting is fundamentally different from the ones of current 6D pose estimation methods.
Gen6D~\cite{gen6d} and OnePose~\cite{onepose} both rely on video sequences of the novel objects, while methods for relative pose estimation~\cite{nguyen2023nope, zhang2022relpose} only work with RGB images and do not estimate the translation component.
Therefore, we compare \acronym with ObjectMatch~\cite{gumeli2023objectmatch}, a state-of-the-art method designed for point cloud registration with low overlap.
This is consistent with our scenario, as we assume that the only overlapping section between two scenes is the one showing the query object.
ObjectMatch is based on SuperGlue~\cite{sarlin2020superglue} to estimate the matches, and on a custom pose estimator.
We run SuperGlue on each image pair.
As in our approach, the predicted mask is used to reject all matches on the background, and the resulting matches are passed to the final ObjectMatch module that outputs the pose. 
We use the model trained on ScanNet~\cite{dai2017scannet} from the official repository. 
We also compare \acronym with a pipeline based on SIFT features~\cite{lowe1999sift} and PointDSC~\cite{pointdsc}.
SIFT is used to extract keypoints and descriptors, which are then filtered by using the segmentation prior.
As in \acronym, the resulting features are used to compute matches between \anchor and \query, which are unprojected in 3D and used to register the point clouds with PointDSC.

We report results with different segmentation priors: (i) the mask predicted by our method (Ours), (ii) the mask predicted with OVSeg~\cite{liang2023ovseg}, a method for open-vocabulary segmentation, and (iii) the ground-truth mask (Oracle).

\subsection{Quantitative results}
\label{sec:quantitative}
\input{main/tables/new_nocs}
\input{main/tables/new_toyl}

Tab.~\ref{tab:nocs} reports the results on REAL275. 
When OVSeg is used as the image segmenter, \acronym outperforms SIFT by $+8.1$ in AR and ObjectMatch by $+11.5$ in AR. 
With our segmentation head as the image segmenter, \acronym shows a performance increase over SIFT by $+7.8$ in AR and over ObjectMatch by $+9.8$ in AR. 
We attribute the smaller performance gap between \acronym and SIFT compared to that with ObjectMatch to the latter's greater sensitivity to domain shifts.
The performance gap between \acronym based on its own predicted masks and the Oracle masks is $-14.3$ AR, which indicates that a portion of the objects are not estimated correctly due to errors in the segmentation.

Tab.~\ref{tab:toyl} reports the results on TOYL. 
When OVSeg is used as segmenter, \acronym outperforms SIFT by $+3.4$ AR and ObjectMatch by $+20.0$ AR.
With our segmentation head as the image segmenter, \acronym shows a performance increase over SIFT by $+3.1$ in AR and over ObjectMatch by $+22.0$ in AR.
In this dataset, SIFT performances are closer to ours than in REAL275.
On the contrary, ObjectMatch performs much worse.
The lower performances of ObjectMatch are due to its sensitivity to the domain shift, which in TOYL is more present due to the different light conditions in the scene pair.
Another characteristic of TOYL is its high variation in poses, for which objects can have very different apparent sizes in the two images.
SIFT scale-invariant design allows it to tackle this variance in appearance and reach better results than ObjectMatch.
We also observe that the performance gap between \acronym with its own masks and the Oracle ones is narrower than in REAL275.
This suggests that \acronym does not require highly accurate segmentation masks, as PointDSC is able to reject spurious matches.

\subsection{Qualitative results}

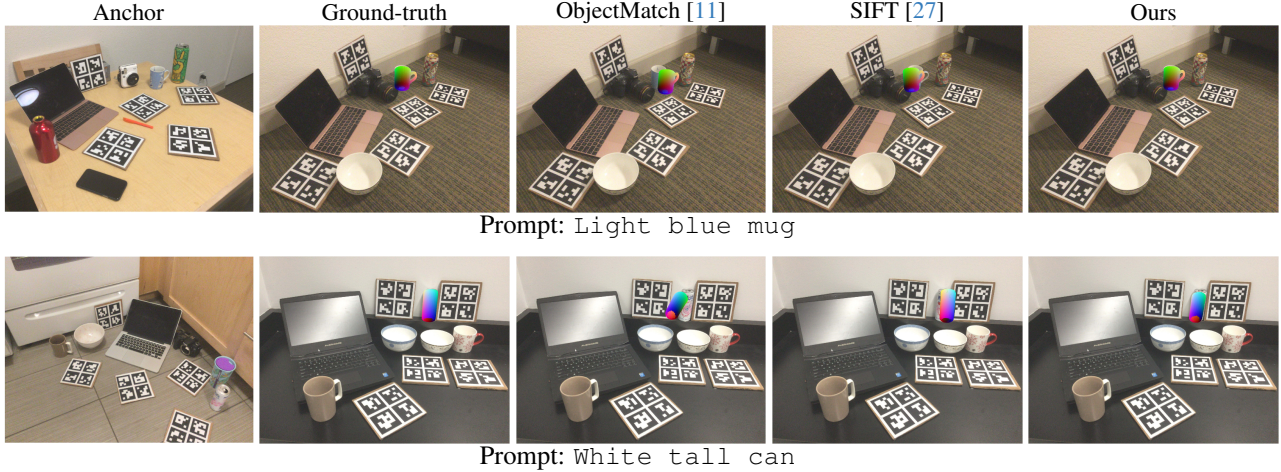
\begin{figure*}[]
    \centering
    \input{main/figures/qualitative/nocs/nocs}
    \caption{
    Examples of qualitative pose results from the REAL275~\cite{nocs} dataset.
    All the results use the segmentation mask predicted by \acronym.
    We show the object model colored my mapping its 3D coordinates to the RGB space.
    }
    \label{fig:nocs_qualitative}
\end{figure*}

\begin{figure*}[]
    \centering
    \input{main/figures/qualitative/toyl/toyl}
    \caption{
    Examples of qualitative pose results from the TOYL~\cite{toyl} dataset.
    All the results use the segmentation mask predicted by \acronym.
    We show the object model colored my mapping its 3D coordinates to the RGB space.
    }
    \label{fig:toyl_qualitative}
\end{figure*}
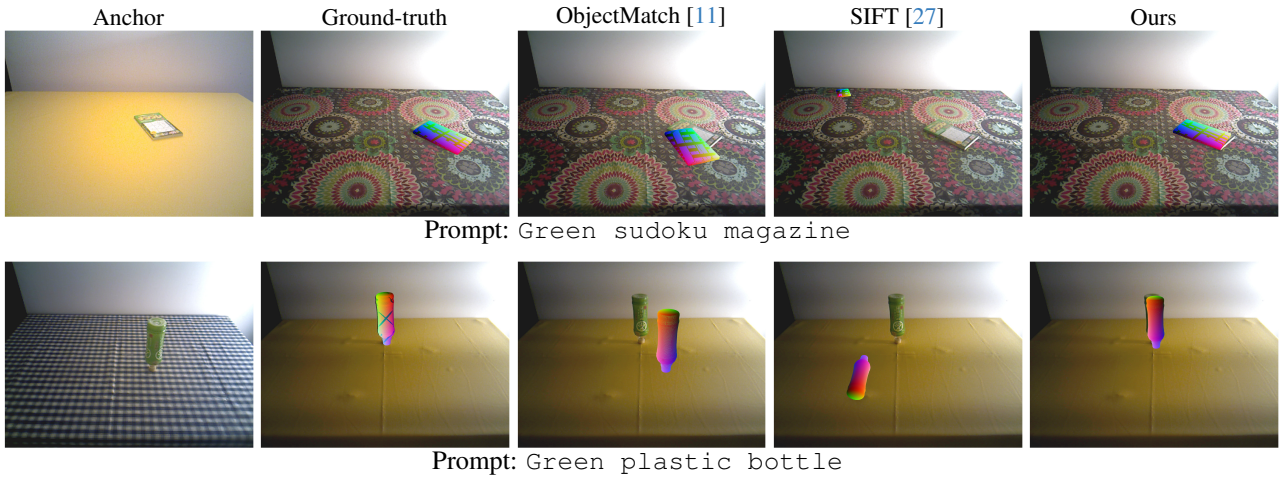

We report in Figs.~\ref{fig:nocs_qualitative}, \ref{fig:toyl_qualitative} some qualitative results on REAL275~\cite{nocs} and TOYL~\cite{toyl}, respectively. 
The predicted mask correctly localizes the objects, but ObjectMatch fails at estimating an accurate pose, due to translation (Fig.~\ref{fig:nocs_qualitative}, top) or rotation errors, (Fig.~\ref{fig:nocs_qualitative}, bottom).
Due to the small objects size, this scenario is challenging for ObjectMatch, while \acronym's higher resolution allows it to retrieve a correct pose also in this case.
Despite small errors in the translation component, SIFT reaches a good degree of accuracy.

On TOYL the results show that our method can handle different light conditions between \anchor and \query.
On the other hand, we observe that the translation component is not as accurate as in REAL275 (see Fig.~\ref{fig:toyl_qualitative}, bottom).
Both ObjectMatch and SIFT present large errors in this challenging dataset. 
SIFT in particular shows large errors in translation, which suggests that such hand-crafted features are unsuitable for this scenario with different light conditions.

\subsection{Ablation study}

We report in Tab.~\ref{tab:ablation_arch} the results obtained on REAL275 when we remove one of the modules of our architecture.
In rows 1, 2 we remove the feature guidance respectively from the fusion module \fusion and from the decoder \decoder.
Both changes result in a significant drop both in pose estimation performance ($-7.4$ and $-11.0$ in AR respectively) and in the segmentation ($-8.0$ and $-12.9$ in mIoU respectively).
This confirms the intuition that adding the guidance from a general-purpose image encoder \guidance \cite{liu2021swin} can specialize the semantic features of CLIP~\cite{clip} for our task.
Removing the upsampling layer (row 3) results in a worse performance on pose estimation, while the impact on the segmentation is small ($-7.1$ and $-1.5$ in AR and mIoU respectively).
This highlights the importance of a higher-resolution feature map to compute the matches, as only using two upsampling layers result in a feature map of size $96\times96$ instead of our default $192\times192$.
In row 4, we replace PointDSC with a RANSAC-based algorithm from a state-of-the-art pose estimation method~\cite{geometricaware6d}.
This causes an important drop in AR ($-16.2$), which shows the importance of a registration algorithm capable of dealing with spurious matches~\cite{pointdsc}.

In Tab.~\ref{tab:ablation_segm} we report the results of two experiments on REAL275 in which we train \acronym for a single task only (segmentation in rows 1-2, and matching in row 3).
When training only with the segmentation loss $\ell_M$ the features learned by \acronym led to worse pose estimation performance than our baseline, as the AR changes by $-7.4$ when using the predicted mask (row 1 vs row 5).
Also the segmentation performance is lower ($-2.2$ mIoU): this suggests that jointly training \acronym for the two tasks can benefit the segmentation performance.
In row 3 we only train \acronym with the feature loss $\ell_F$.
In this case, the AR drops by $-6.2$ when comparing with the same segmenter (row 3 vs row 4).
This confirms our intuition that jointly training a model for pose estimation and segmentation can benefit both tasks.

In Tab.~\ref{tab:ablation_prompt} we report the results of the evaluation on REAL275 with different prompts.
In rows 1-2 the object name is replaced by ``object'', in order to simulate the possible behaviour of a user describing an unknown object, for which a name cannot be provided.
In this case, the segmentation completely fails, and subsequently the predicted poses are poor ($-29.2$ AR and $-63.5$ mIoU with respect to the baseline).
In rows 3-4 we input a misleading prompt to the network, by keeping the original object name but changing the brief description (e.g., ``a white closed laptop'' for a laptop that appears open and brown).
This results in a change of $-6.8$ AR and $-10.1$ mIoU with respect to the baseline (row 8).
However, the model does not fail completely.
This suggests that \acronym has some capabilities of selecting the most useful information from the prompt given the appearance on the scenes.
When the object description is removed and only the object name is kept (rows 5-6), the drop in performance with respect to the baseline is less severe ($-2.2$ in AR, $-3.1$ in mIoU).
REAL275 presents some scenes with multiple objects with the same name (e.g., Fig.~\ref{fig:nocs_qualitative} at the bottom shows two ``bowl'' objects).
In these scenes, the description can be decisive to solve the ambiguity, while in the other cases it is less important.
See the Supplementary Material for the complete list of prompts.
 
\input{main/tables/ablation_arch}
\input{main/tables/ablation_segm}
\input{main/tables/ablation_prompt}

\subsection{Sensitivity analysis}
\label{sec:err}
We report in Fig.~\ref{fig:errors} the distribution of AR with segmentation quality and camera distance between \anchor and \query across all samples of the REAL275 test set.
We observe in Fig.~\ref{fig:errors}(a) that AR is low before reaching about 0.4 IoU, after which there is a sharp increase in pose quality.
On the other hand, AR appears to reach a plateau after 0.7 IoU.
This confirms our intuition that \acronym does not require an high segmentation accuracy to successfully estimate the pose.
However, there is not a direct dependency relation between the two metrics, as there are many samples with low AR and high mIoU.
On the other hand, the relation between pose quality and camera distance in Fig.~\ref{fig:errors}(b) is more linear, as the latter is lower when AR is higher.
Intuitively, an high camera distance implies a lower number of matches, and therefore lowers the pose quality.

\input{main/figures/errors}

%% file: main/tables/new_nocs.tex
\begin{table}[t!]
\centering
\tabcolsep 3pt
\caption{
Results on REAL275~\cite{nocs}.
We report in \textbf{bold} font and \underline{underlined} respectively the best and second best result when using our predicted mask, and \colorbox{myazure}{highlight} our main result. 
The $\Delta$ score is the difference between \acronym and the nearest competitor.
Key: Obj.Mat.: ObjectMatch, ADD: ADD(S)-0.1d.
}
\vspace{-3mm}
\label{tab:nocs}
 \resizebox{\columnwidth}{!}{%
    \begin{tabular}{cllcccccc}
        \toprule
        & Method & Segm. prior & AR\higherbetter & VSD\higherbetter & MSSD\higherbetter & MSPD\higherbetter & ADD\higherbetter & mIoU\higherbetter \\
        \midrule
        {\color{gray} \small 1} & $\,\,$ \multirow{3}{*}{\rotatebox{23}{SIFT~\cite{lowe1999sift}}} & \oracle{Oracle} & \oracle{34.1} & \oracle{16.5} & \oracle{37.9} & \oracle{48.0} & \oracle{16.4} & \oracle{100.0} \\ 
        {\color{gray} \small 2} & & OVSeg~\cite{liang2023ovseg} & 18.3 & 8.6 & 19.9 & 26.5 & 7.4 & 56.4 \\ 
        {\color{gray} \small 3} & & Ours & \underline{24.4} & 12.2 & 27.3 & \underline{33.8} & 12.8 & 66.5 \\ 
        \midrule
        {\color{gray} \small 4} & \multirow{3}{*}{\rotatebox{23}{Obj.Mat.~\cite{gumeli2023objectmatch}}} & \oracle{Oracle} & \oracle{26.0} & \oracle{15.5} & \oracle{31.7} & \oracle{30.8} & \oracle{13.4} & \oracle{100.0} \\
        {\color{gray} \small 5} & & OVSeg~\cite{liang2023ovseg} & 14.9 & 9.1 & 18.8 & 16.8 & 7.8 & 56.4 \\ 
        {\color{gray} \small 6} & & Ours & 22.4 & \underline{14.1} & \underline{27.9} & 25.2 & \underline{13.2} & 66.5\\
        \midrule
        {\color{gray} \small 7} & $\,\,\,$ \multirow{3}{*}{\rotatebox{23}{\acronym}} & \oracle{Oracle} & \oracle{46.5} & \oracle{32.1} & \oracle{50.9} & \oracle{56.7} & \oracle{34.9} & \oracle{100.0} \\
        {\color{gray} \small 8} & & OVSeg~\cite{liang2023ovseg} & 26.4 & 18.3 & 29.4 & 31.5 & 17.2 & \underline{56.4} \\
        {\color{gray} \small 9} & & \cellcolor{myazure} Ours & \cellcolor{myazure} \textbf{32.2} & \cellcolor{myazure} \textbf{23.6} & \cellcolor{myazure} \textbf{36.6} & \cellcolor{myazure} \textbf{36.4} & \cellcolor{myazure} \textbf{24.3} & \cellcolor{myazure} \textbf{66.5}\\
        \midrule 
        {\color{gray} \small 10} & $\Delta$ score & & \impp{7.8} & \impp{9.5} & \impp{8.7} & \impp{2.6} & \impp{11.1} & \impp{10.1} \\
        \bottomrule
    \end{tabular}
}
\end{table}

%% file: main/tables/new_toyl.tex
\begin{table}[t!]
\centering
\tabcolsep 3pt
\caption{
Results on Toyota-Light~\cite{toyl}.
We report in \textbf{bold} and \underline{underlined} respectively the best and second best result when using our predicted mask, and \colorbox{myazure}{highlight} our main result.
The $\Delta$ score is the difference between \acronym and the nearest competitor.
Key: Obj.Mat.: ObjectMatch, ADD: ADD(S)-0.1d.
}
\vspace{-3mm}
\label{tab:toyl}
\resizebox{\columnwidth}{!}{%
\begin{tabular}{cllcccccc}
    \toprule
    
    & Method & Segm. prior & AR\higherbetter & VSD\higherbetter & MSSD\higherbetter & MSPD\higherbetter & ADD\higherbetter & mIoU\higherbetter \\
    \midrule
    {\color{gray} \small 1} & $\,\,$ \multirow{3}{*}{\rotatebox{23}{SIFT~\cite{lowe1999sift}}} & \oracle{Oracle} & \oracle{30.3} & \oracle{7.3} & \oracle{39.6} & \oracle{44.1} & \oracle{14.1} & \oracle{100.0} \\
    {\color{gray} \small 2} & & OVSeg~\cite{liang2023ovseg} & 25.8 & 6.4 & 34.2 & 36.9 & 11.8 & 75.5 \\   
    {\color{gray} \small 3} & & Ours & \underline{27.2} & \underline{5.7} & \underline{35.4} & \underline{40.6} & \underline{9.9} & 68.1 \\ 
    \midrule
    {\color{gray} \small 4} & \multirow{3}{*}{\rotatebox{23}{Obj.Mat.~\cite{gumeli2023objectmatch}}} & \oracle{Oracle} & \oracle{9.8} & \oracle{2.4} & \oracle{13.0} & \oracle{14.0} & \oracle{5.4} & \oracle{100.0} \\
    {\color{gray} \small 5} & & OVSeg~\cite{liang2023ovseg} & 9.2 & 2.6 & 12.1 & 13.0 & 5.3 & 75.5 \\   
    {\color{gray} \small 6} & & Ours & 8.3 & 2.2 & 10.5 & 12.1 & 3.8 & 68.1 \\ 
    \midrule
    {\color{gray} \small 7} & $\,\,\,$ \multirow{3}{*}{\rotatebox{23}{\acronym}} & \oracle{Oracle} & \oracle{34.1} & \oracle{13.9} & \oracle{42.9} & \oracle{45.5} & \oracle{22.9} & \oracle{100.0} \\
    {\color{gray} \small 8} & & OVSeg~\cite{liang2023ovseg} & 29.2 & 11.9 & 36.8 & 38.9 & 18.9 & \textbf{75.5} \\
    {\color{gray} \small 9} & & \cellcolor{myazure} Ours & \cellcolor{myazure} \textbf{30.3} & \cellcolor{myazure} \textbf{12.1} & \cellcolor{myazure} \textbf{37.5} & \cellcolor{myazure} \textbf{41.4} & \cellcolor{myazure} \textbf{20.9} & \cellcolor{myazure} \underline{68.1} \\
    \midrule
    {\color{gray} \small 10} & $\Delta$ score & & \impp{3.1} & \impp{6.4} & \impp{2.1} & \impp{0.8} & \impp{11.0} & \impn{7.4} \\
    \bottomrule
    
\end{tabular}
}
\end{table}

%% file: main/figures/qualitative/nocs/nocs.tex
\hspace*{0.00mm}
\begin{minipage}{0.19\textwidth}
    \centering{\small{Anchor}}
\end{minipage}
\begin{minipage}{0.19\textwidth}
    \centering{\small{Ground-truth}}
\end{minipage}
\begin{minipage}{0.19\textwidth}
    \centering{\small{ObjectMatch~\cite{gumeli2023objectmatch}}}
\end{minipage}
\begin{minipage}{0.19\textwidth}
    \centering{\small{SIFT~\cite{lowe1999sift}}}
\end{minipage}
\begin{minipage}{0.19\textwidth}
    \centering{\small{Ours}}
\end{minipage}

\hspace*{0.00mm}
\begin{minipage}{0.19\textwidth}
    \begin{overpic}[width=1\textwidth]{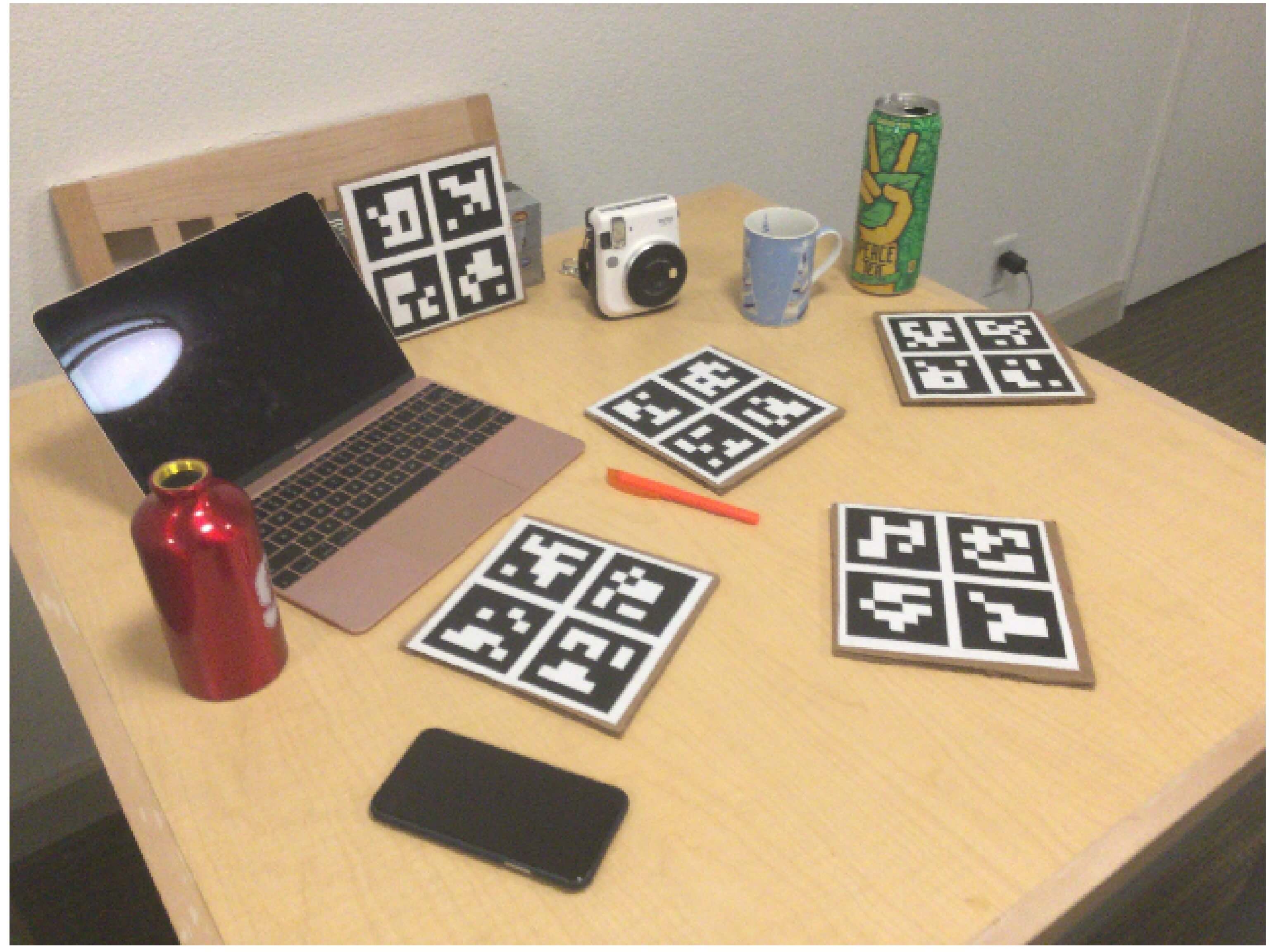}
    \end{overpic}
\end{minipage}
\begin{minipage}{0.19\textwidth}
    \begin{overpic}[width=1\textwidth]{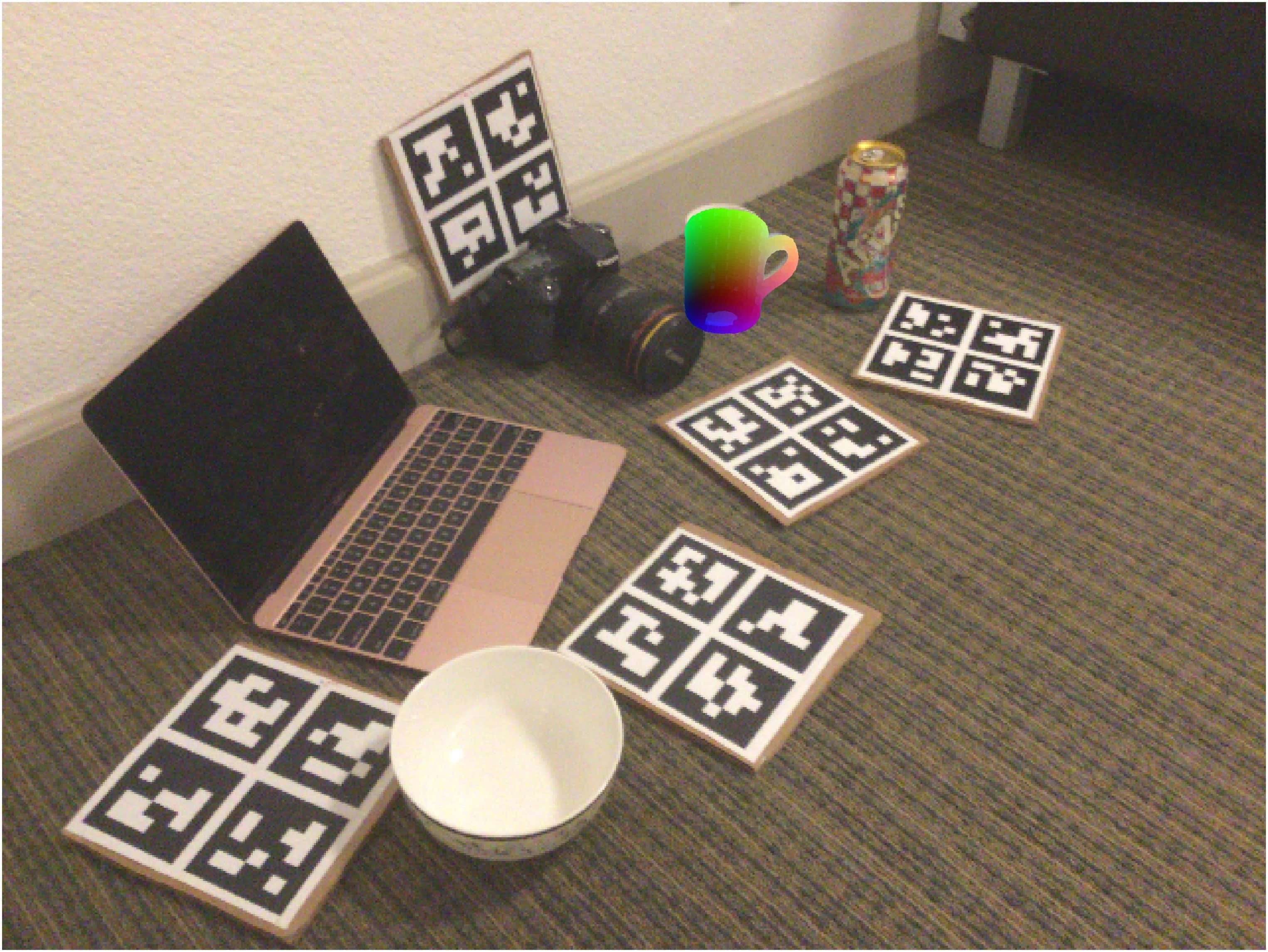}
    \end{overpic}
\end{minipage}
\begin{minipage}{0.19\textwidth}
    \begin{overpic}[width=1\textwidth]{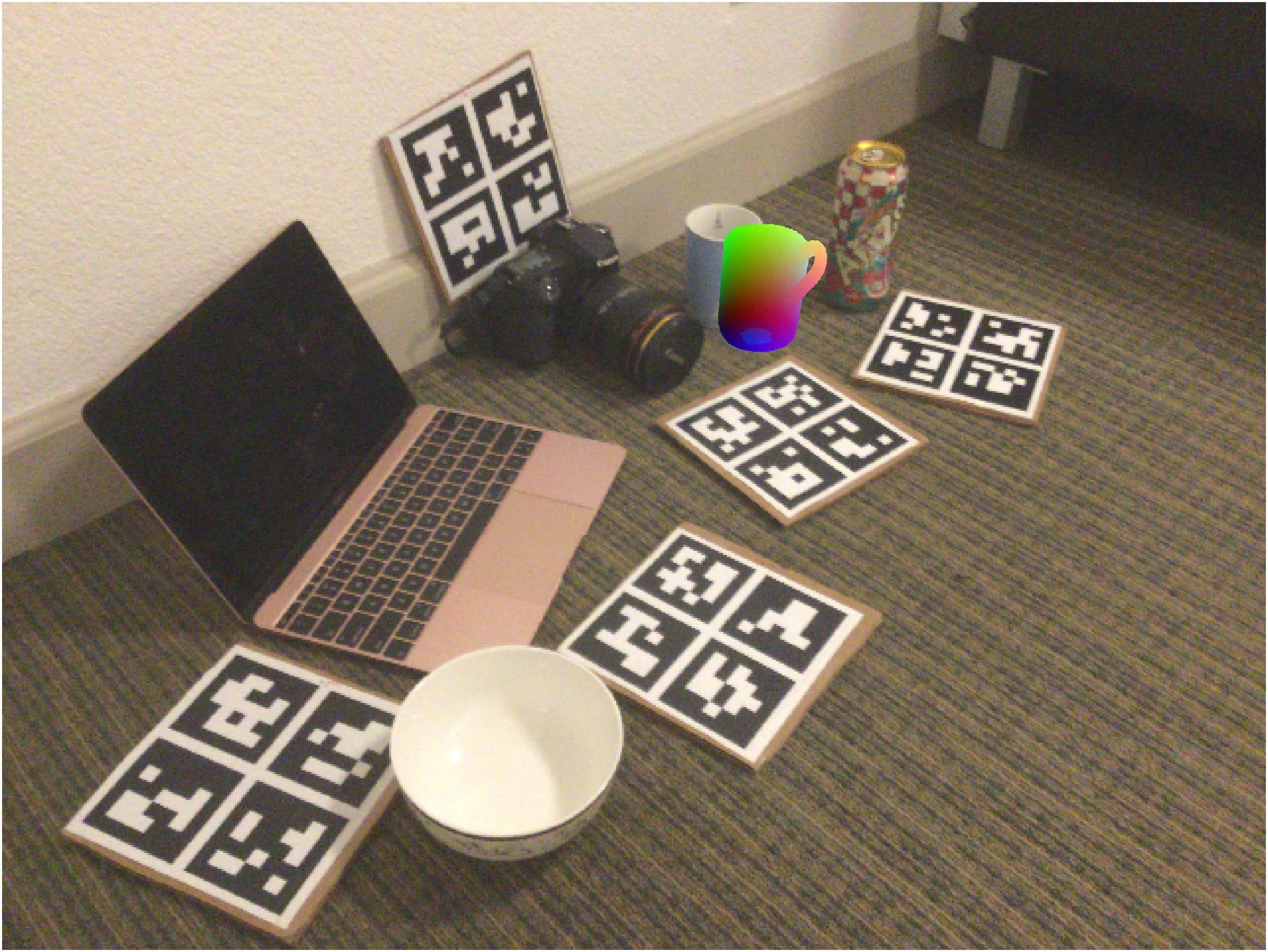}
    \end{overpic}
\end{minipage}
\begin{minipage}{0.19\textwidth}
    \begin{overpic}[width=1\textwidth]{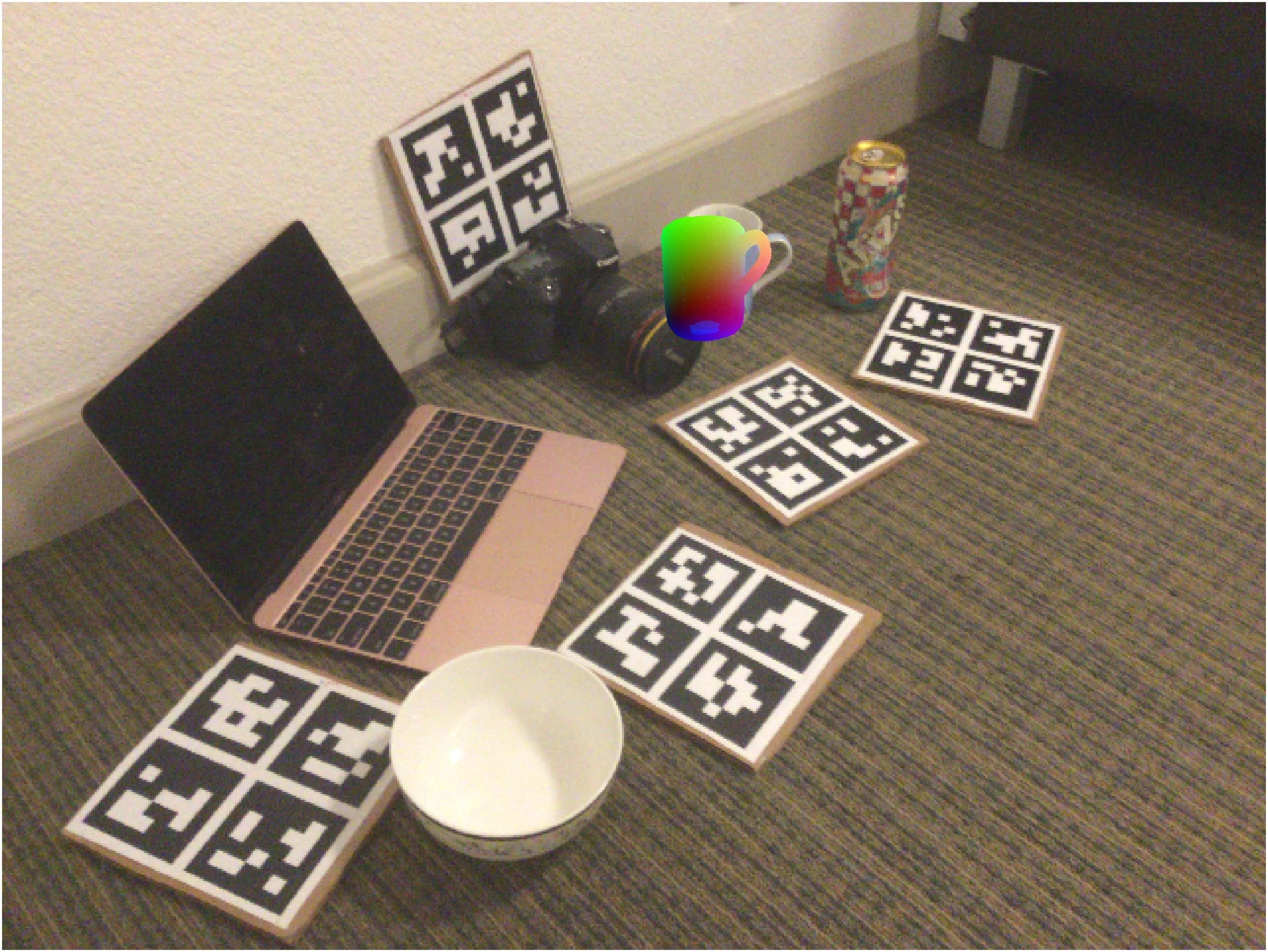}
    \end{overpic}
\end{minipage}
\begin{minipage}{0.19\textwidth}
    \begin{overpic}[width=1\textwidth]{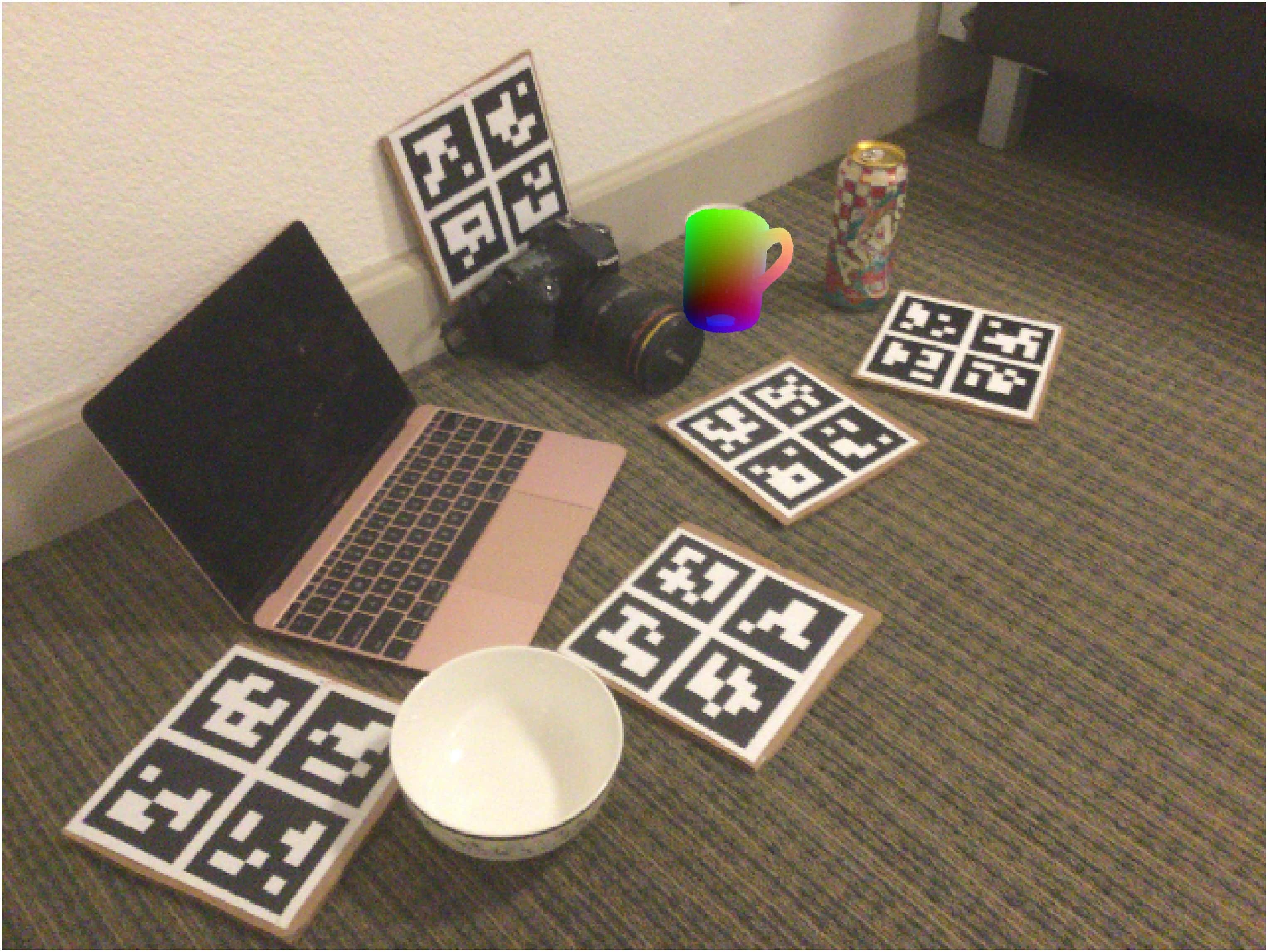}
    \end{overpic}
\end{minipage}
\vspace*{2mm}
\begin{minipage}{\textwidth}
    \centering{Prompt: \texttt{Light blue mug}}
\end{minipage}
\hspace*{0.00mm}
\begin{minipage}{0.19\textwidth}
    \begin{overpic}[width=1\textwidth]{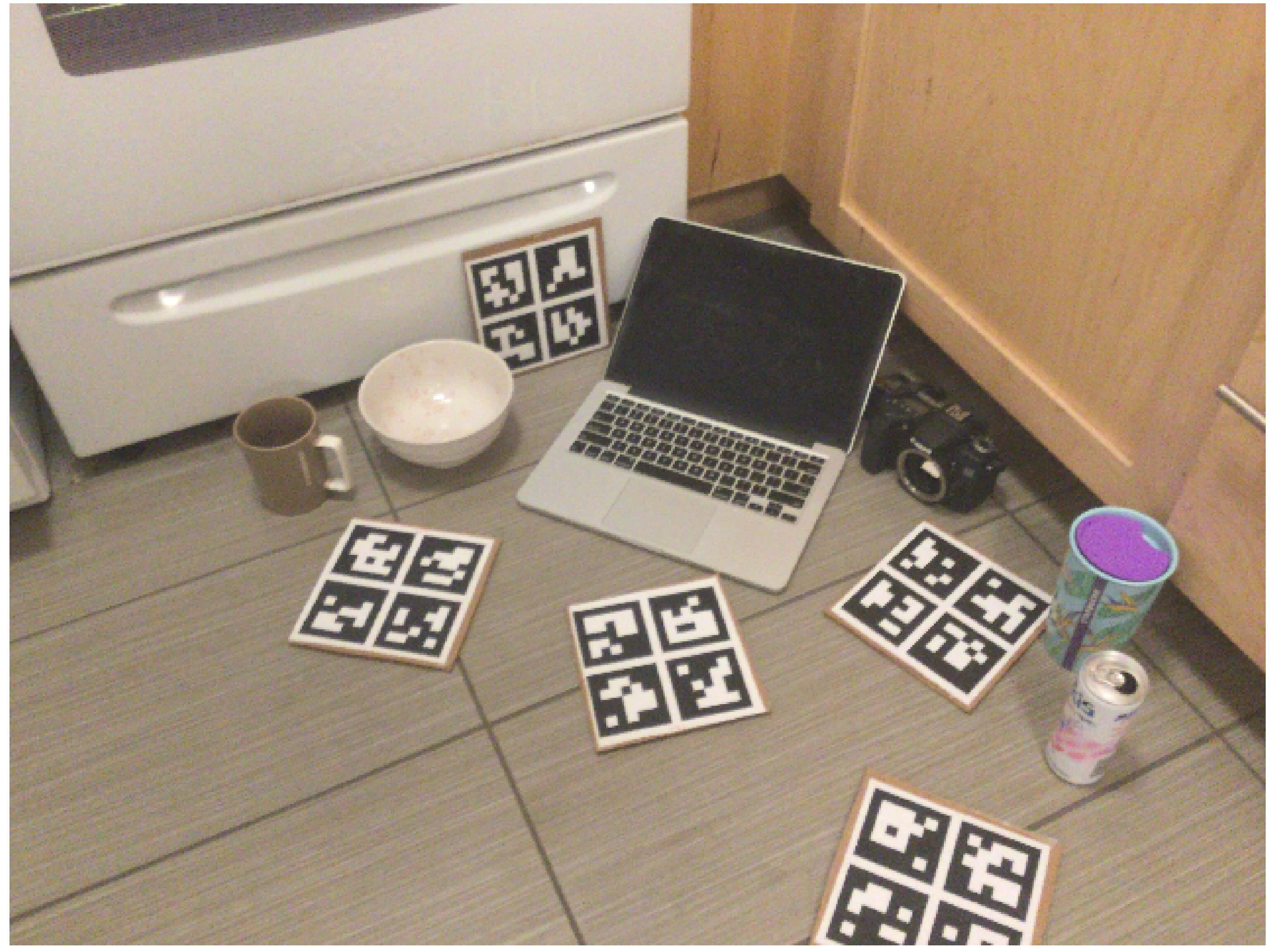}
    \end{overpic}
\end{minipage}
\begin{minipage}{0.19\textwidth}
    \begin{overpic}[width=1\textwidth]{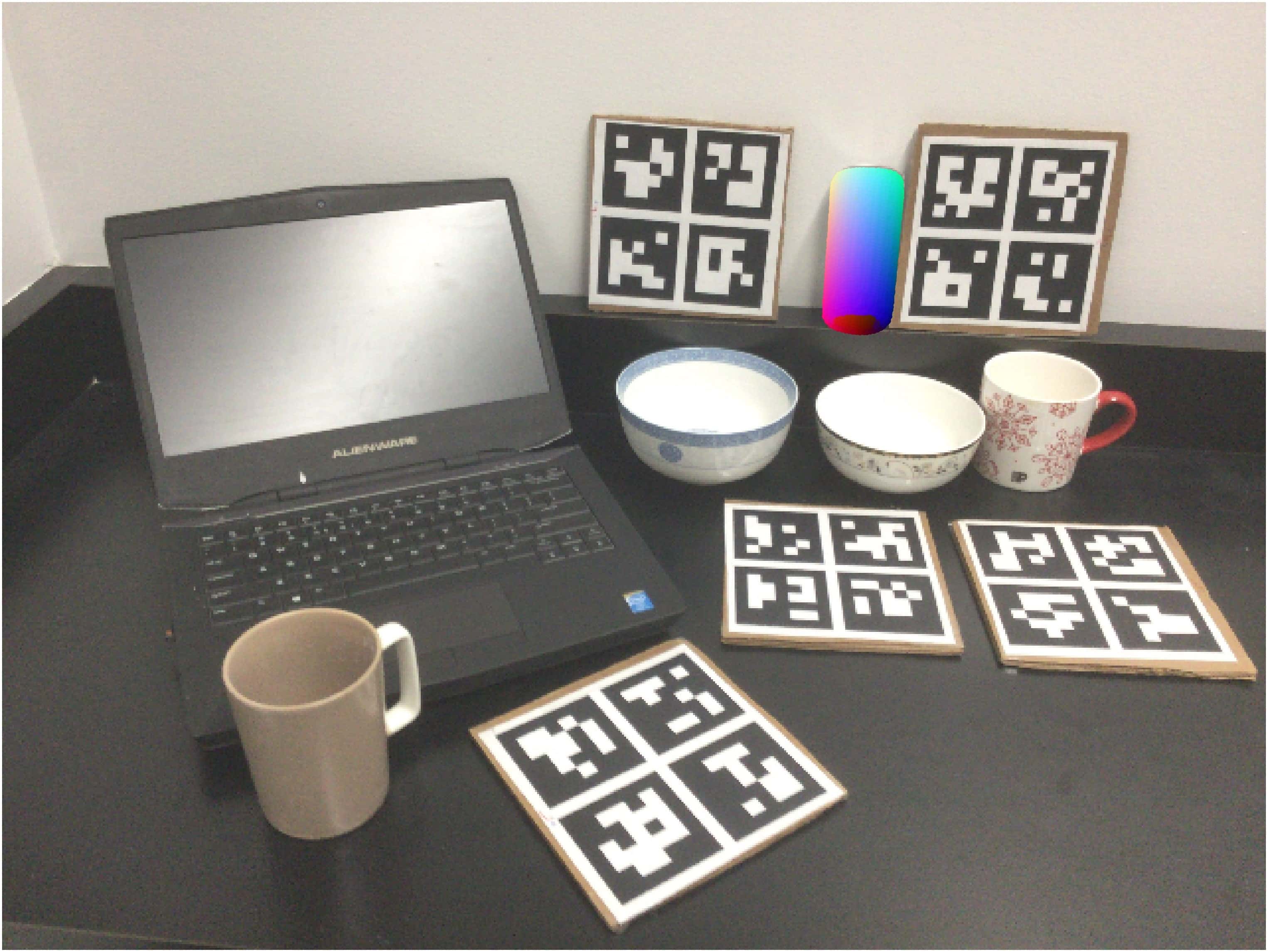}
    \end{overpic}
\end{minipage}
\begin{minipage}{0.19\textwidth}
    \begin{overpic}[width=1\textwidth]{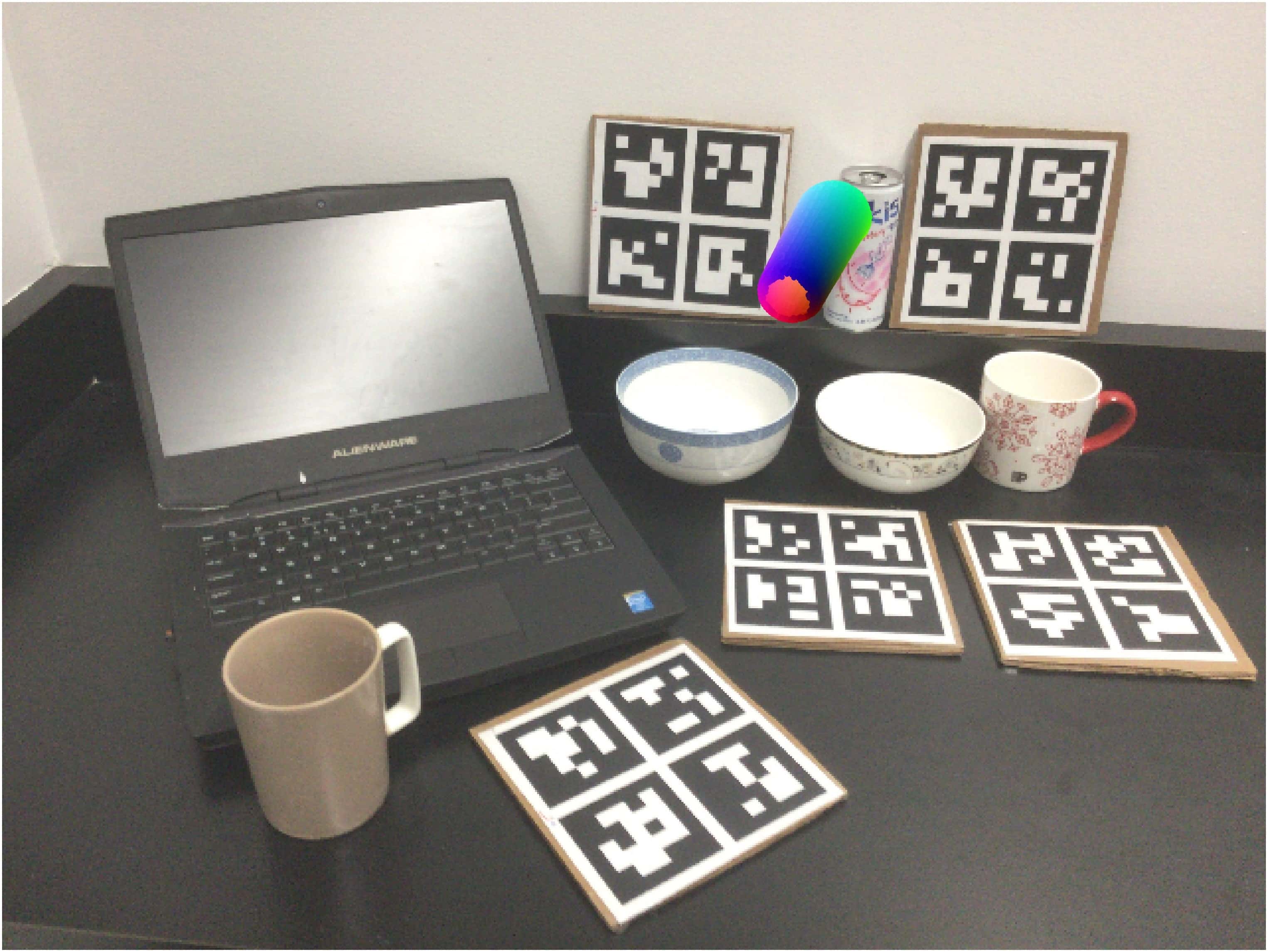}
    \end{overpic}
\end{minipage}
\begin{minipage}{0.19\textwidth}
    \begin{overpic}[width=1\textwidth]{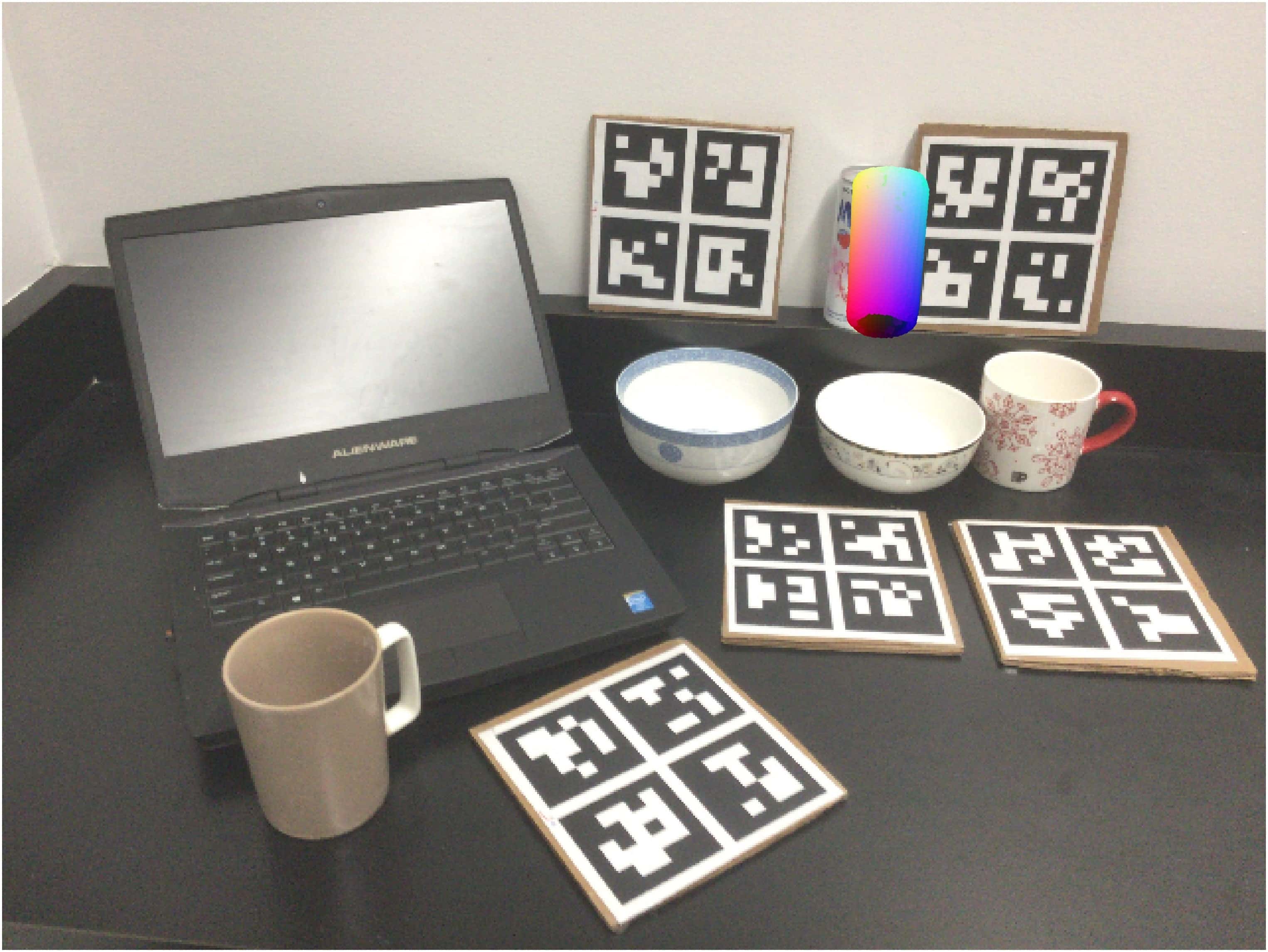}
    \end{overpic}
\end{minipage}
\begin{minipage}{0.19\textwidth}
    \begin{overpic}[width=1\textwidth]{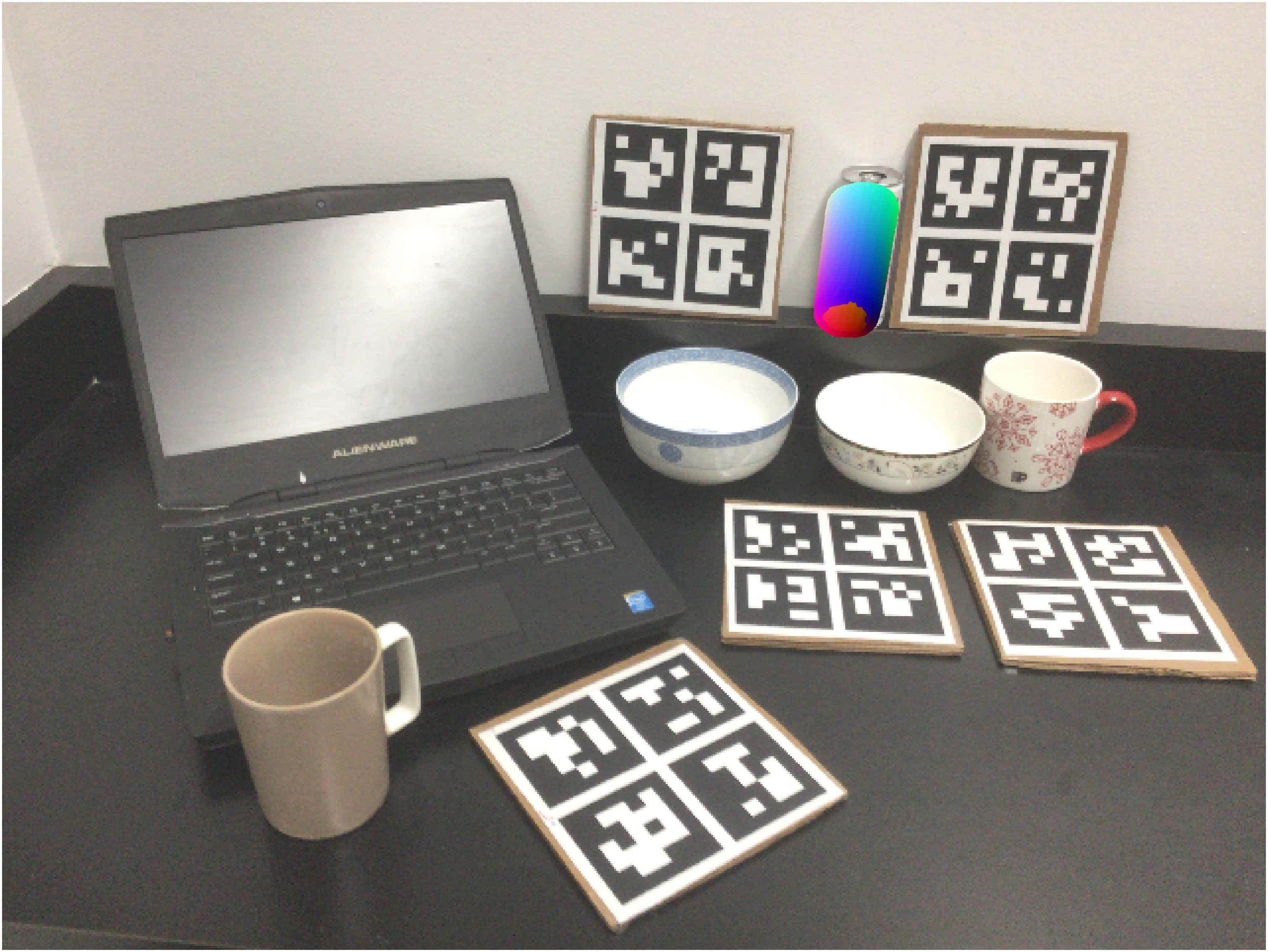}
    \end{overpic}
\end{minipage}
\begin{minipage}{\textwidth}
    \centering{Prompt: \texttt{White tall  can}}
\end{minipage}

%% file: main/figures/qualitative/toyl/toyl.tex
\hspace*{0.00mm}
\begin{minipage}{0.19\textwidth}
    \centering{\small{Anchor}}
\end{minipage}
\begin{minipage}{0.19\textwidth}
    \centering{\small{Ground-truth}}
\end{minipage}
\begin{minipage}{0.19\textwidth}
    \centering{\small{ObjectMatch}~\cite{gumeli2023objectmatch}}
\end{minipage}
\begin{minipage}{0.19\textwidth}
    \centering{\small{SIFT~\cite{lowe1999sift}}}
\end{minipage}
\begin{minipage}{0.19\textwidth}
    \centering{\small{Ours}}
\end{minipage}

\hspace*{0.00mm}
\begin{minipage}{0.19\textwidth}
    \begin{overpic}[width=1\textwidth]{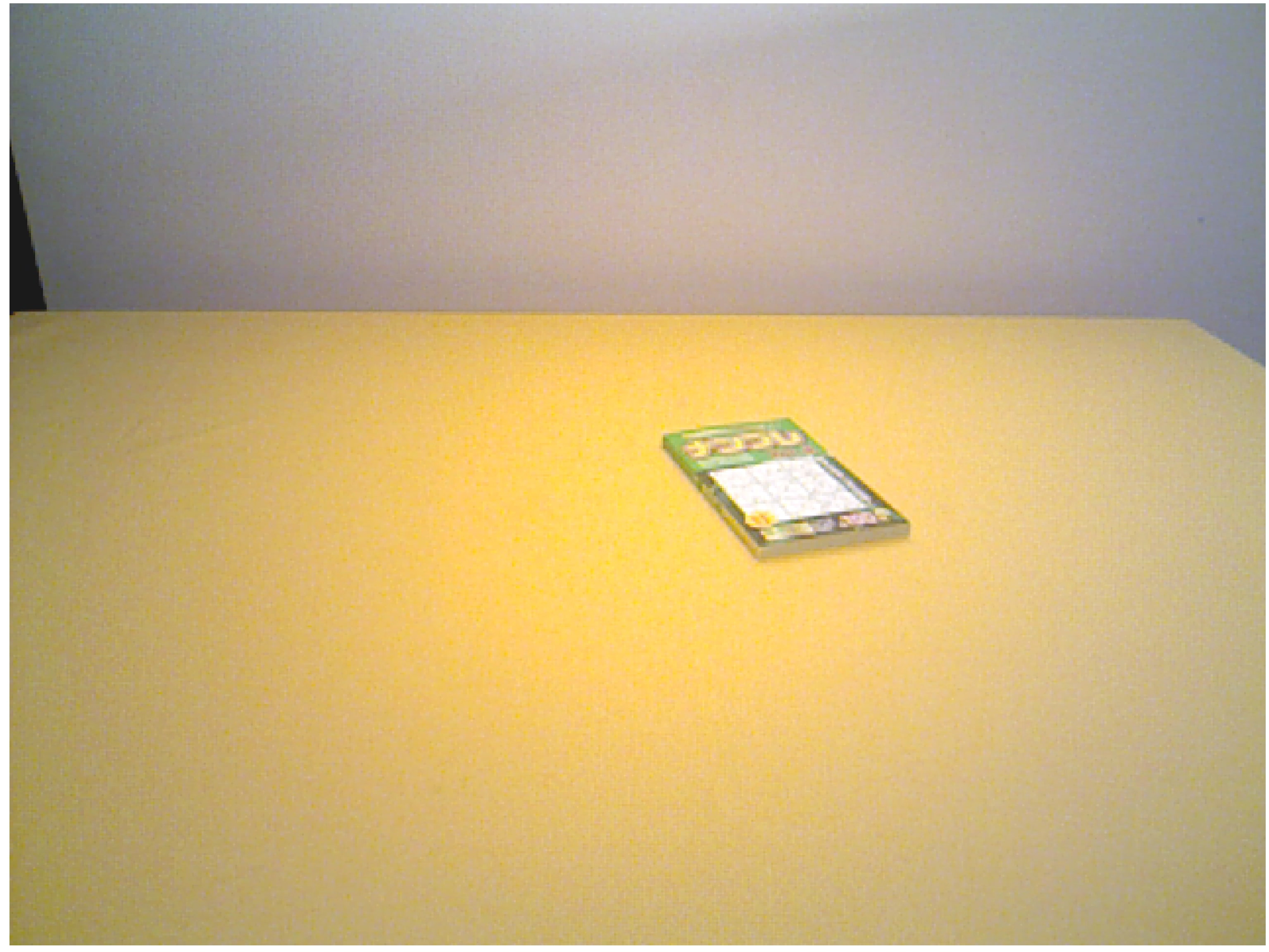}
    \end{overpic}
\end{minipage}
\begin{minipage}{0.19\textwidth}
    \begin{overpic}[width=1\textwidth]{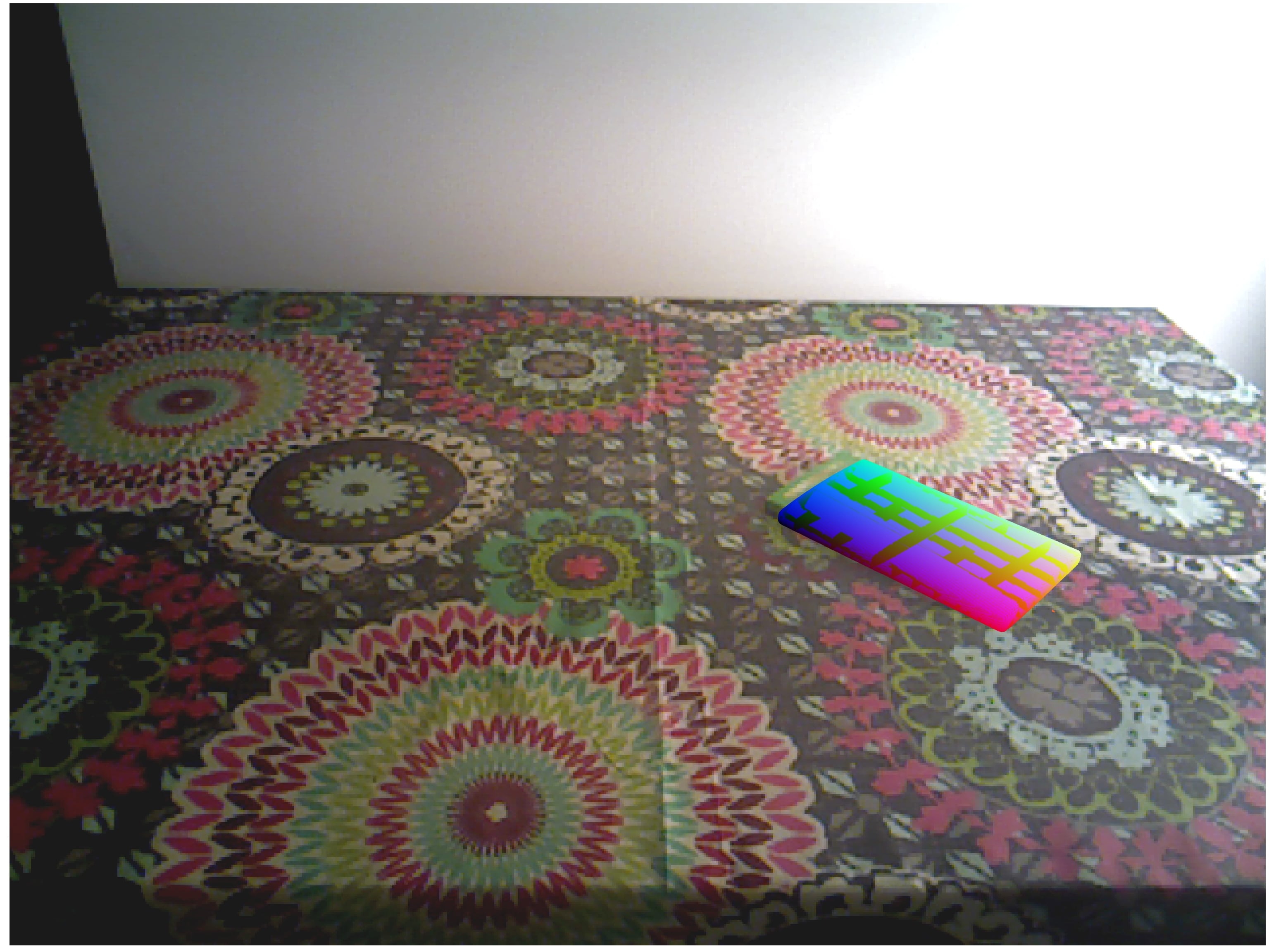}
    \end{overpic}
\end{minipage}
\begin{minipage}{0.19\textwidth}
    \begin{overpic}[width=1\textwidth]{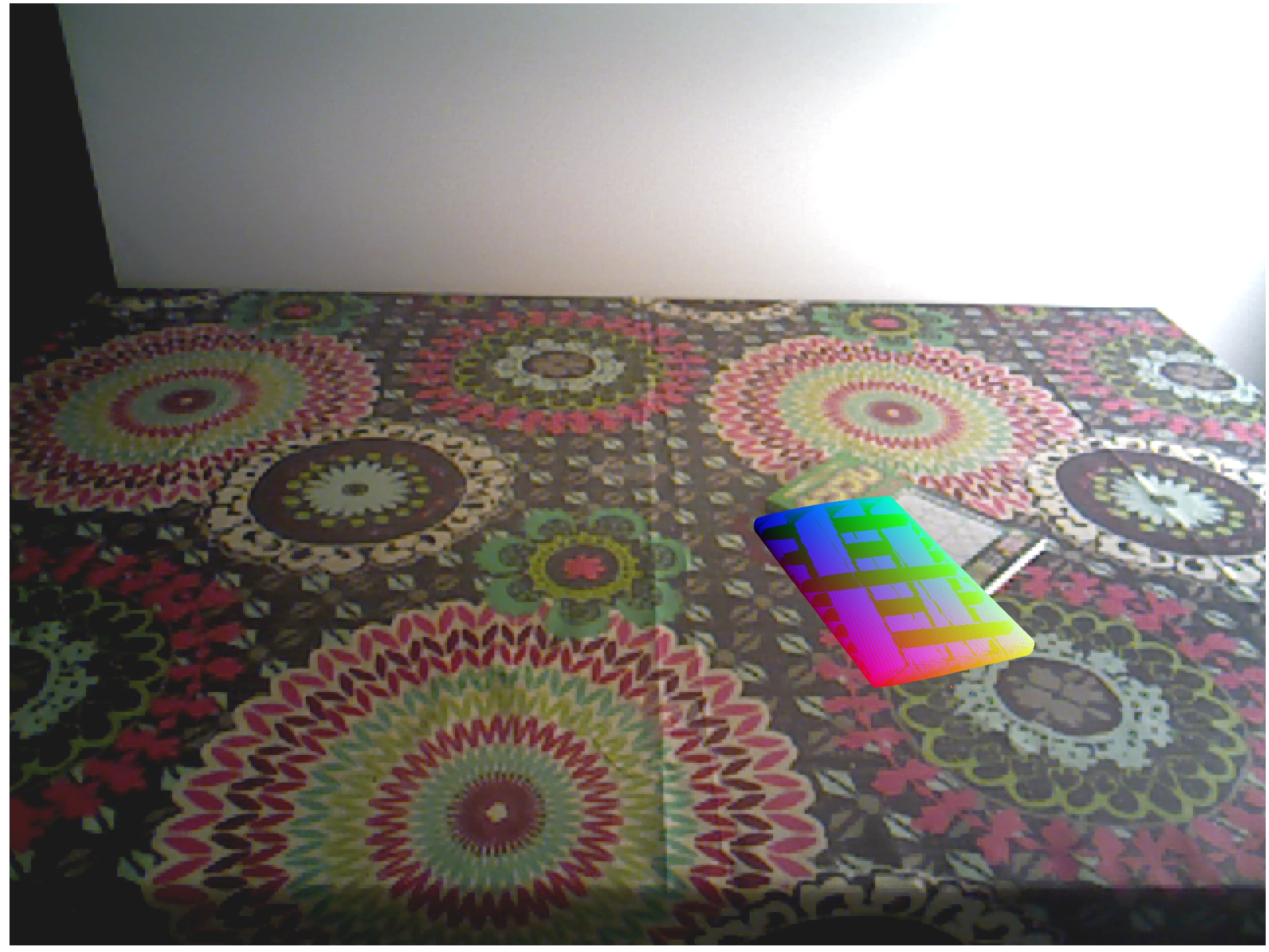}
    \end{overpic}
\end{minipage}
\begin{minipage}{0.19\textwidth}
    \begin{overpic}[width=1\textwidth]{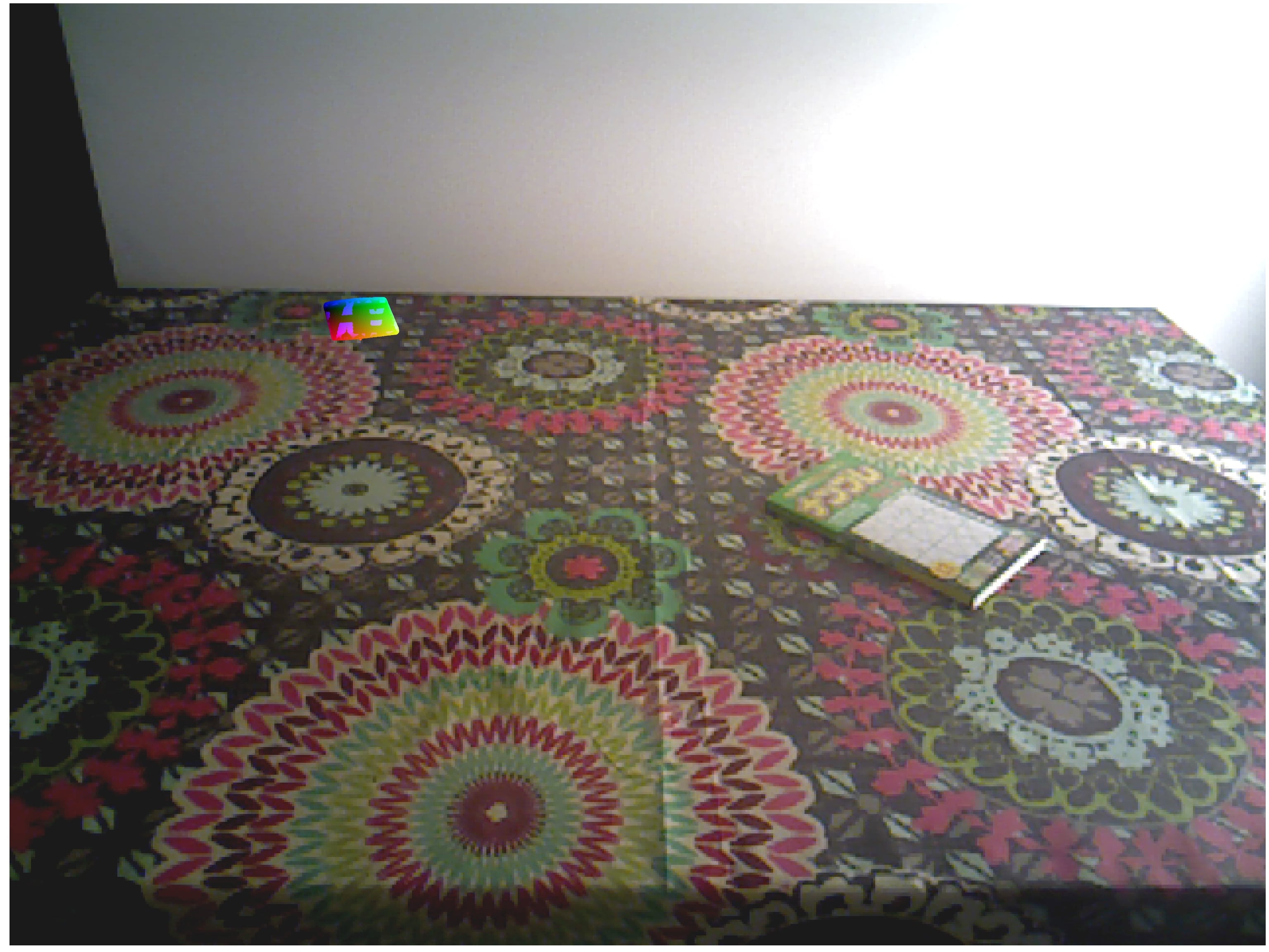}
    \end{overpic}
\end{minipage}
\begin{minipage}{0.19\textwidth}
    \begin{overpic}[width=1\textwidth]{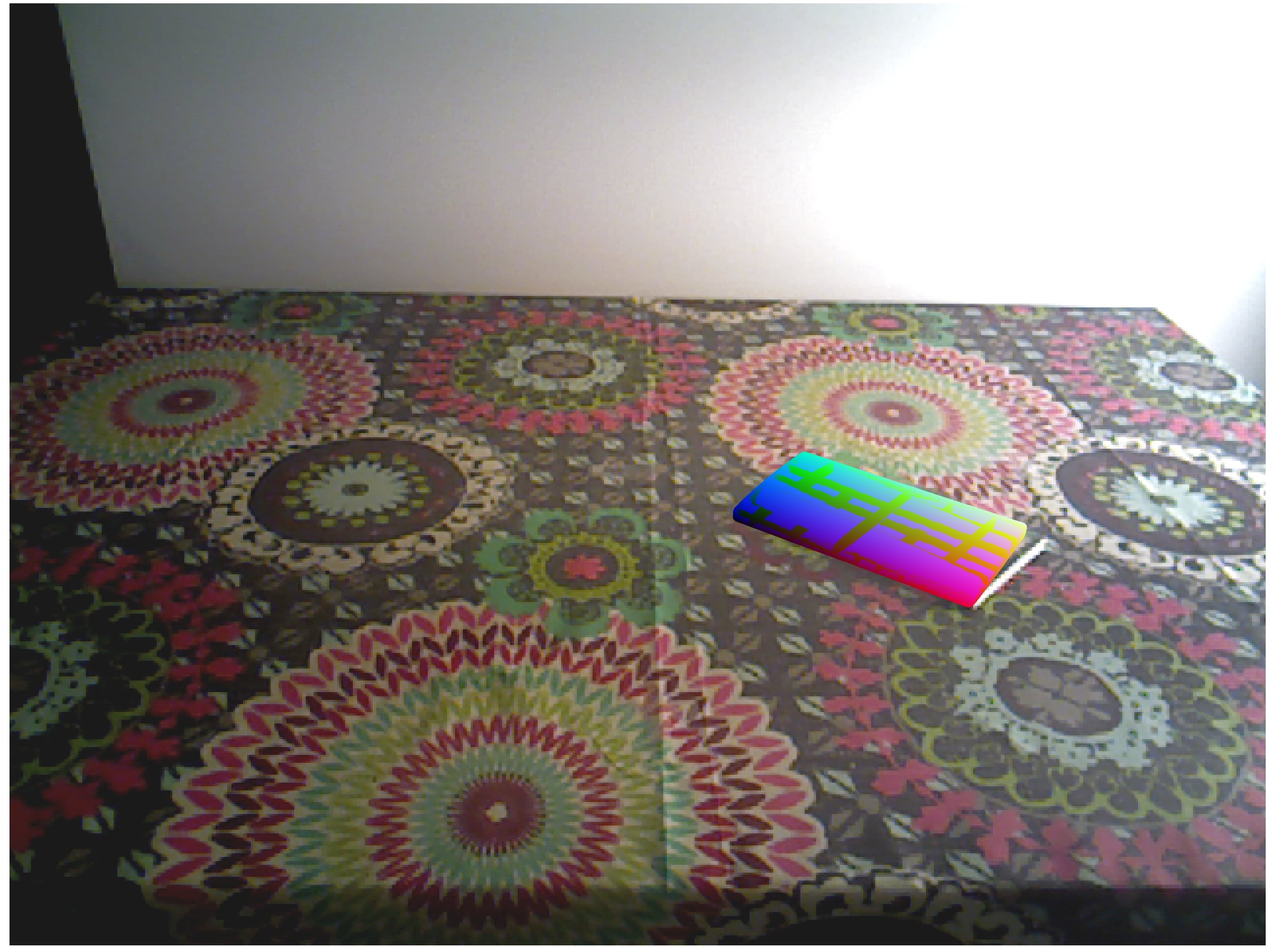}
    \end{overpic}
\end{minipage}
\vspace*{2mm}
\begin{minipage}{\textwidth}
    \centering{Prompt: \texttt{Green sudoku magazine}}
\end{minipage}
\hspace*{0.00mm}
\begin{minipage}{0.19\textwidth}
    \begin{overpic}[width=1\textwidth]{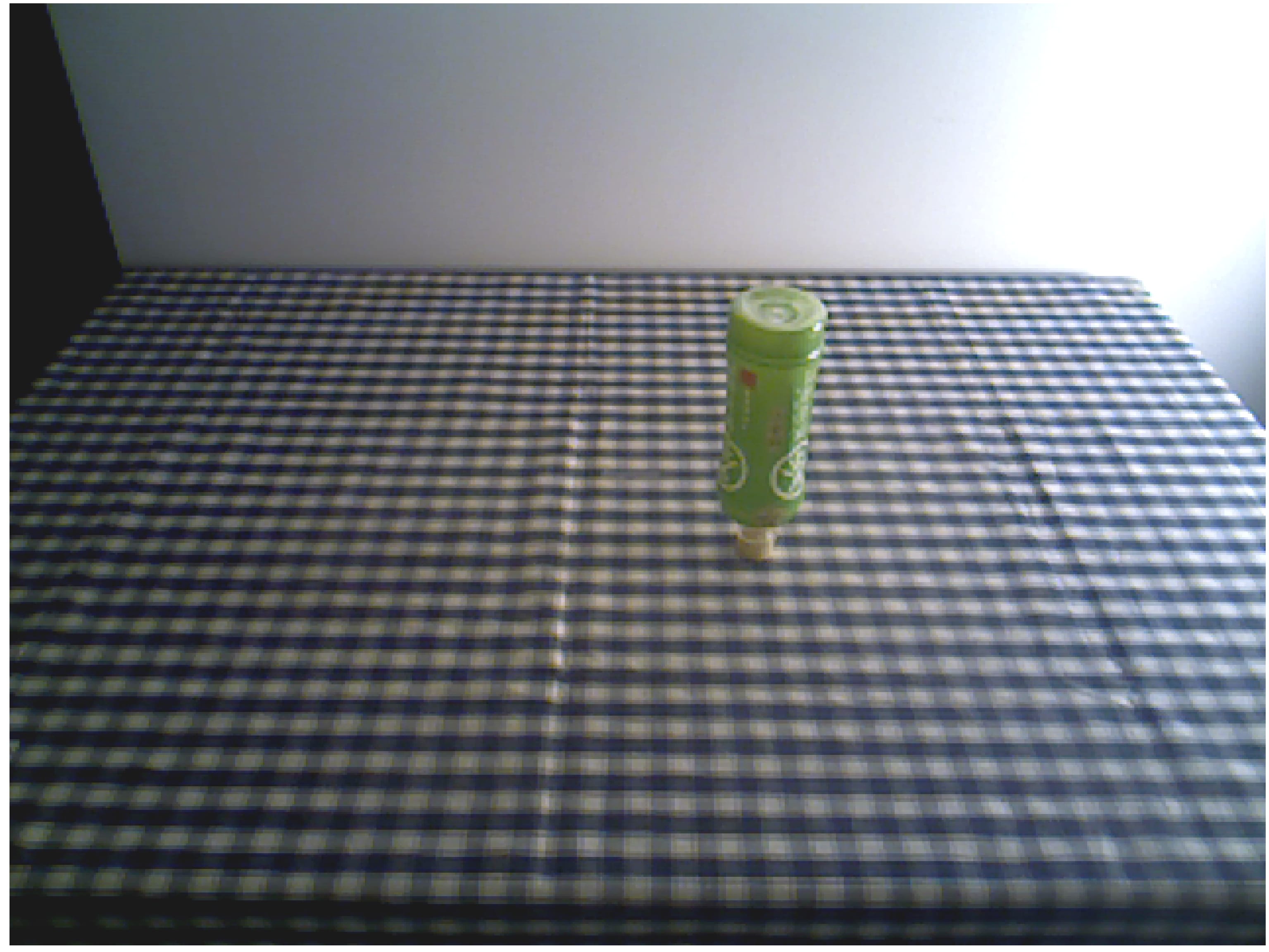}
    \end{overpic}
\end{minipage}
\begin{minipage}{0.19\textwidth}
    \begin{overpic}[width=1\textwidth]{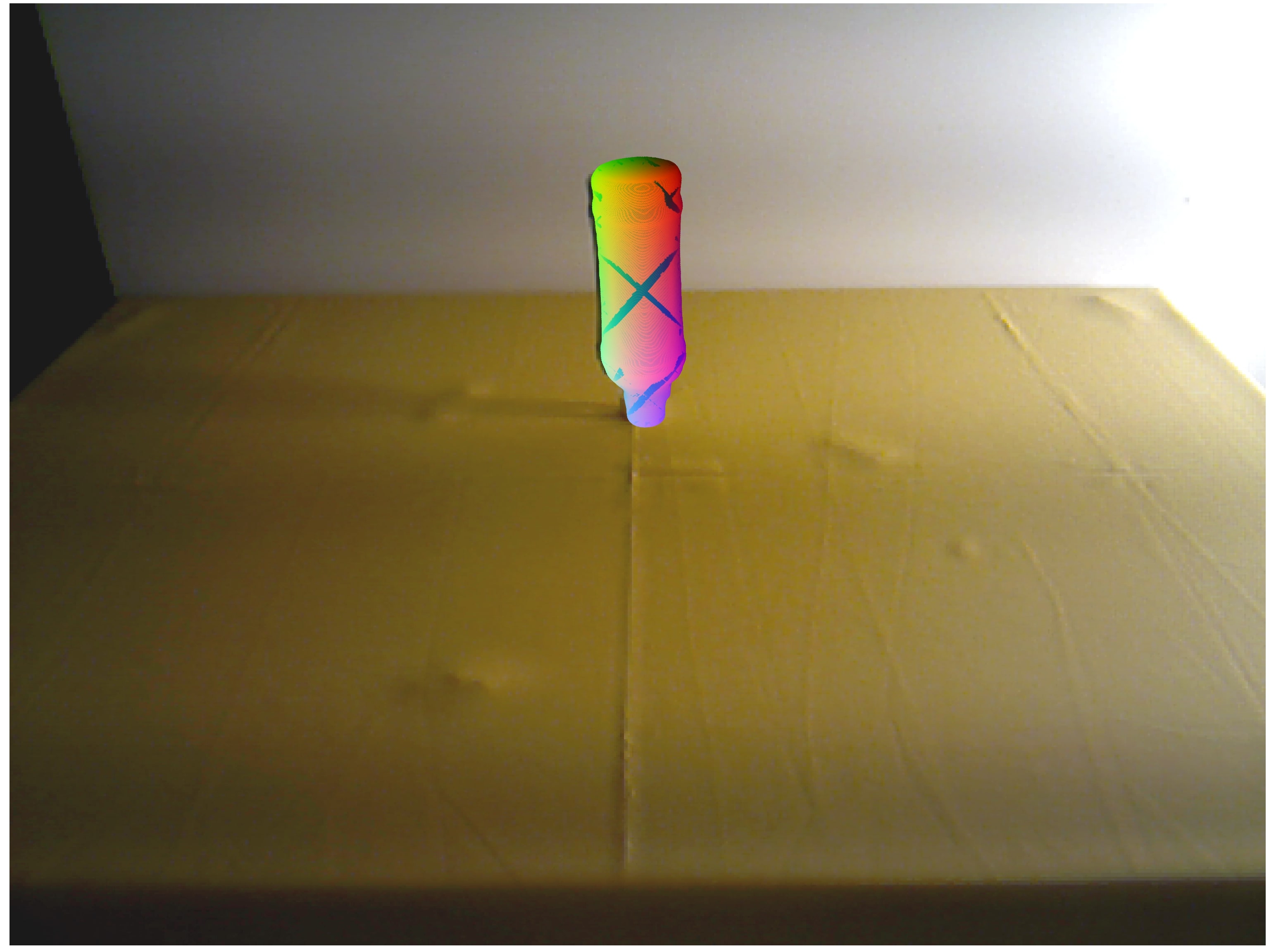}
    \end{overpic}
\end{minipage}
\begin{minipage}{0.19\textwidth}
    \begin{overpic}[width=1\textwidth]{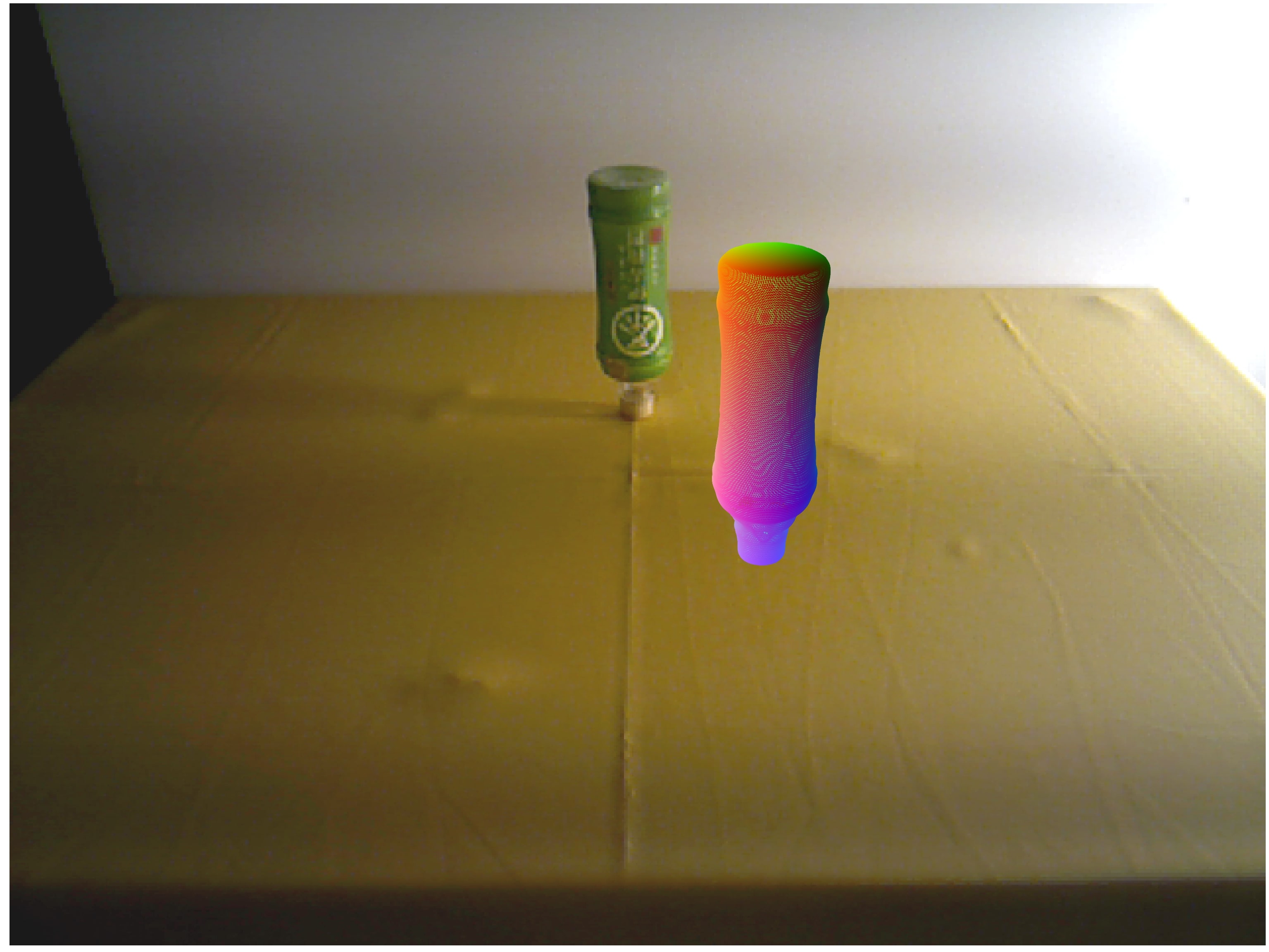}
    \end{overpic}
\end{minipage}
\begin{minipage}{0.19\textwidth}
    \begin{overpic}[width=1\textwidth]{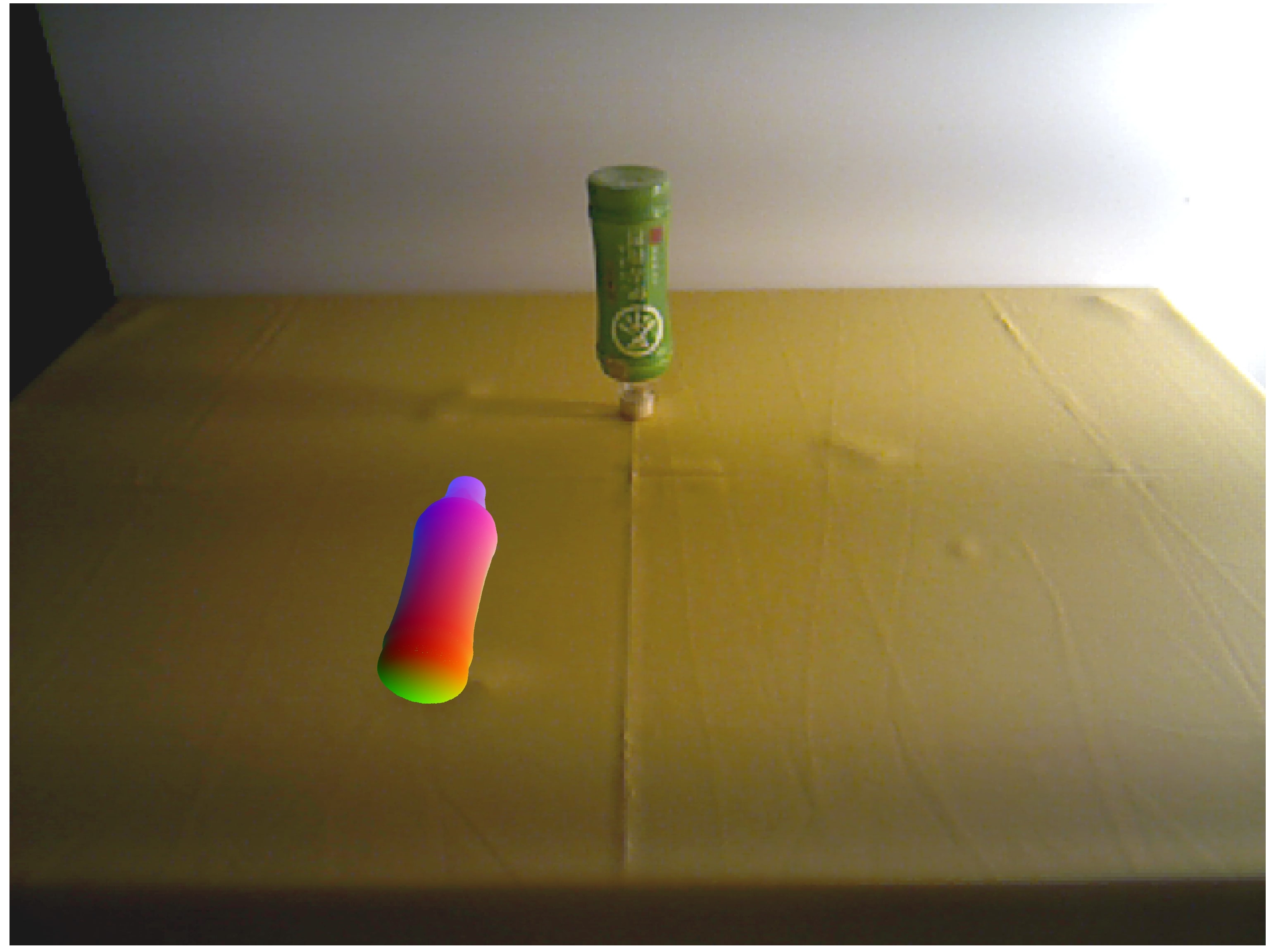}
    \end{overpic}
\end{minipage}
\begin{minipage}{0.19\textwidth}
    \begin{overpic}[width=1\textwidth]{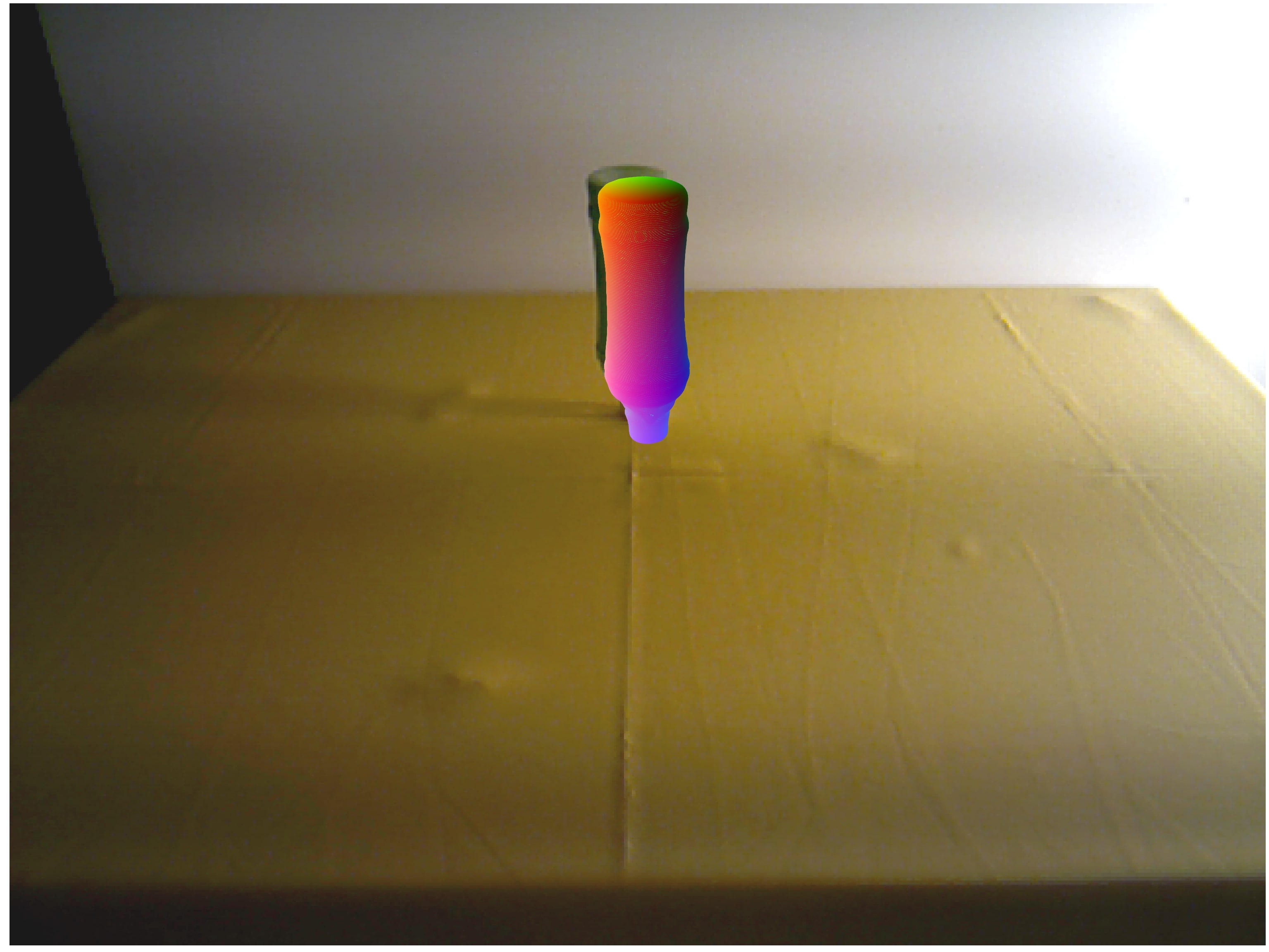}
    \end{overpic}
\end{minipage}
\begin{minipage}{\textwidth}
    \centering{Prompt: \texttt{Green plastic bottle}}
\end{minipage}

%% file: main/tables/ablation_arch.tex
\begin{table}[t!]
\centering
\tabcolsep 3pt
\caption{
Ablation on the architecture components on REAL275~\cite{nocs}, with our predicted masks.
Key: ADD: ADD(S)-0.1d, \textbf{Bold}: best method.
}
\vspace{-3mm}
\label{tab:ablation_arch}
\resizebox{\columnwidth}{!}{%
\begin{tabular}{clcccccc}
    \toprule
    & Method & AR\higherbetter & VSD\higherbetter & MSSD\higherbetter & MSPD\higherbetter & ADD\higherbetter & mIoU\higherbetter \\    \midrule
    {\color{gray} \small 1} & w/o cost agg. guidance & 24.8 & 17.2 & 28.8 & 28.4 & 17.9 &58.5 \\ 
    {\color{gray} \small 2} & w/o decoder guidance & 21.2 & 16.4 & 23.2 & 24.0 & 14.4 &53.6 \\
    {\color{gray} \small 3} & w/o extra upsampling & 25.1 & 16.4 & 27.6 & 31.1 & 18.5 & 65.0 \\ 
    {\color{gray} \small 4} & w RANSAC~\cite{ransac} & 15.0 & 4.7 & 16.0 & 24.4 & 5.2 & 66.5 \\ 
    \midrule
    \rowcolor{myazure} {\color{gray} \small 5} & \acronym & \textbf{32.2} & \textbf{23.6} & \textbf{36.6} & \textbf{36.4} & \textbf{24.3} & \textbf{66.5} \\
    \bottomrule
\end{tabular}
}
\end{table}

%% file: main/tables/ablation_segm.tex
\begin{table}[t!]
\centering
\tabcolsep 3pt
\caption{
Ablation on the two tasks performed by \acronym on REAL275~\cite{nocs}.
Key: ADD: ADD(S)-0.1d, \textbf{Bold}: best method.
}
\vspace{-3mm}
\label{tab:ablation_segm}
\resizebox{\columnwidth}{!}{%
\begin{tabular}{cllcccccc}
    \toprule
    & Method & Prior & AR\higherbetter & VSD\higherbetter & MSSD\higherbetter & MSPD\higherbetter & ADD\higherbetter & mIoU\higherbetter \\
    \midrule
    {\color{gray} \small 1} & Segm. only & Ours & 24.8 & 15.1 & 28.2 & 31.0 & 14.6 & 63.3 \\ 
    {\color{gray} \small 2} & Segm. only & Oracle & 33.2 & 18.4 & 35.8 & 45.4 & 18.1 & 100.0 \\ 
    {\color{gray} \small 3} & Matches only & Oracle & 40.3 & 27.3 & 42.8 & 50.7 & 26.6 & 100.0 \\ 
    \midrule
    {\color{gray} \small 4} & \acronym & Oracle & 46.5 & 32.1 & 50.9 & 56.7 & 34.9 & 100.0 \\
    {\color{gray} \small \cellcolor{myazure}5} & \cellcolor{myazure}\acronym & \cellcolor{myazure}Ours & \cellcolor{myazure}\textbf{32.2} & \cellcolor{myazure}\textbf{23.6} & \cellcolor{myazure}\textbf{36.6} & \cellcolor{myazure}\textbf{36.4} & \cellcolor{myazure}\textbf{24.3} &\cellcolor{myazure} \textbf{66.5} \\
    \bottomrule
\end{tabular}
}
\end{table}



%% file: main/tables/ablation_prompt.tex
\begin{table}[t!]
\centering
\tabcolsep 6pt
\caption{
Ablation on the prompt on REAL275~\cite{nocs}.
Key: ADD: ADD(S)-0.1d, \textbf{Bold}: best method.
}
\vspace{-3mm}
\label{tab:ablation_prompt}
\resizebox{\columnwidth}{!}{%
\begin{tabular}{cllcccccc}
    \toprule
    & Prompt type & Prior & AR\higherbetter & VSD\higherbetter & MSSD\higherbetter & MSPD\higherbetter & ADD\higherbetter & mIoU\higherbetter \\
    \midrule
    {\color{gray} \small 1} & \multirow{2}{*}{\rotatebox{21}{No name}} & \oracle{Oracle} & \oracle{38.6} & \oracle{22.2} & \oracle{42.7} & \oracle{50.8} & \oracle{21.5} & \oracle{100.0} \\ 
    {\color{gray} \small 2} & & Ours & 3.0 & 1.4 & 4.1 & 3.6 & 0.4 & 3.0 \\ 
    \midrule
    {\color{gray} \small 3} & \multirow{2}{*}{\rotatebox{21}{Misleading}} & \oracle{Oracle} & \oracle{39.4} & \oracle{22.1} & \oracle{43.5} & \oracle{52.8} & \oracle{24.9}  & \oracle{100.0} \\
    {\color{gray} \small 4} & & Ours &25.4 & 19.1 & 29.0 & 28.0 & 14.5 &56.4 \\
    \midrule
    {\color{gray} \small 5} & \multirow{2}{*}{\rotatebox{21}{Generic}} & \oracle{Oracle} & \oracle{39.0} & \oracle{22.7} & \oracle{43.4} & \oracle{50.7} & \oracle{26.1} & \oracle{100.0} \\ 
    {\color{gray} \small 6} & & Ours & 30.0 & 21.9 & 34.0 & 34.1 & 19.0 &63.4  \\ 
    \midrule
    {\color{gray} \small 7} & \multirow{2}{*}{\rotatebox{21}{Oryon}} & \oracle{Oracle} & \oracle{46.5} & \oracle{32.1} & \oracle{50.9} & \oracle{56.7} & \oracle{34.9}& \oracle{100.0} \\ 
    {\color{gray} \small 8} & & \cellcolor{myazure}Ours & \cellcolor{myazure}\textbf{32.2} & \cellcolor{myazure}\textbf{23.6} & \cellcolor{myazure}\textbf{36.6} & \cellcolor{myazure}\textbf{36.4} & \cellcolor{myazure}\textbf{24.3} & \cellcolor{myazure}\textbf{66.5} \\ 
    \bottomrule
\end{tabular}
}
\end{table}

%% file: main/figures/errors.tex
\begin{figure}[t]
    \centering
    \begin{overpic}[width=\columnwidth, trim=20 0 60 0]{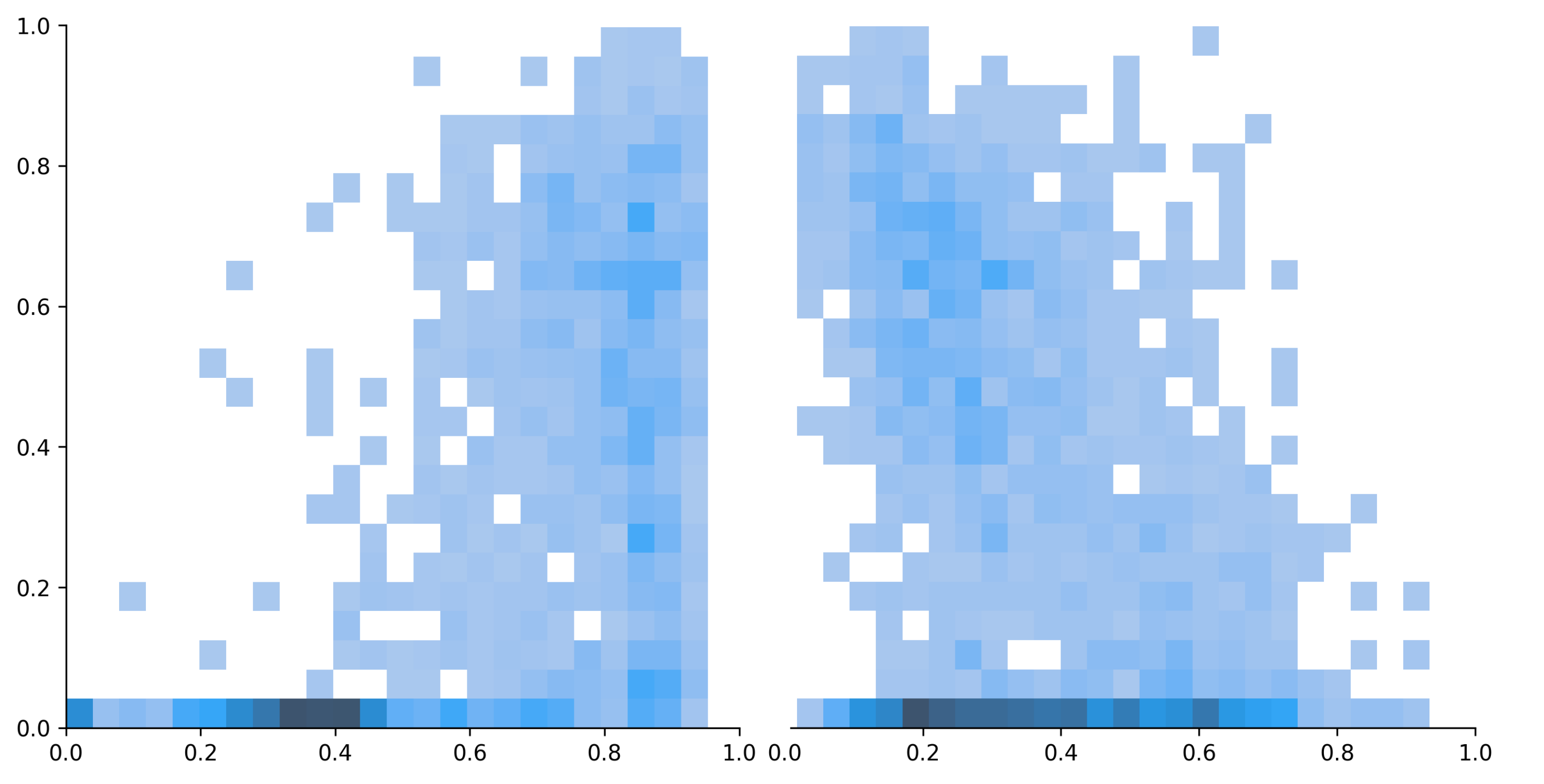}
        \put(-3,25){\rotatebox{90}{\small{AR}}}
        \put(18,-3){\small{(a) mIoU}}
        \put(63,-3){\small{(b) Camera distance}}
        \end{overpic}
    \vspace{-2mm}
    \caption{
    Distribution of AR with mIoU (a) and camera distance between image pairs (b) of our best experiment on REAL275~\cite{nocs}.
    }
    \label{fig:errors}
\end{figure}

%% file: main/sections/5_conclusion.tex
\section{Conclusions}\label{sec:conclusion}
We present \acronym, an approach to tackle generalizable pose estimation from a new perspective. 
Instead of relying on a visual representation of the novel object and on complex onboarding procedures, we use a textual description of the object of interest, which can be provided by a non-expert user.
We show that our approach is successful in the challenging scenario in which the image pair shows different scenes.
\acronym accomplishes this by jointly addressing image matching and segmentation in an open-vocabulary approach, and by leveraging on a text-visual fusion module.
We provide extensive ablation studies to assess the importance of each component and of the prompt composition.

\noindent\textbf{Limitations.} \acronym requires depth maps in order to run the registration algorithm and retrieve the relative pose. 
This limits the applicability of our method to contexts in which the depth information is available.
Another limitation is in providing a description for certain objects (e.g., mechanical components) in order to express the prompt.

\noindent\textbf{Future work.} \acronym could be adapted to work with RGB images only by considering depth prediction.

\noindent\textbf{Acknowledgements.} This work was supported by the European Union’s Horizon Europe research and innovation programme under grant agreement No 101058589 (AI-PRISM), and it made use of time on the Tier 2 HPC facility JADE2, funded by EPSRC (EP/T022205/1).

%% file: supp/commands.tex
\newcommand{\fabiocomment}[1]{\todo[color=purple!20, inline, author=Fabio]{#1}}
\newcommand{\davidecomment}[1]{\todo[color=blue!20, inline, author=Davide]{#1}}

\newcommand{\fabio}[1]{\textbf{\textcolor{purple!75}{#1}}}
\newcommand{\davide}[1]{\textbf{\textcolor{blue!75}{#1}}}
\newcommand{\changjae}[1]{\textbf{\textcolor{olive!75}{#1}}}
\newcommand{\andrea}[1]{\textbf{\textcolor{orange!75}{#1}}}
\newcommand{\warning}[1]{\textbf{\textcolor{red!75}{#1}}}

\newcommand{\imgencoder}[0]{$\phi_V$\xspace}
\newcommand{\textencoder}[0]{$\phi_T$\xspace}
\newcommand{\guidance}[0]{$\phi_G$\xspace}
\newcommand{\decoder}[0]{$\phi_D$\xspace}
\newcommand{\fusion}[0]{$\phi_{TV}$\xspace}

\newcommand{\learnedprompts}[0]{$\mathbf{D}$\xspace}
\newcommand{\learnedembs}[0]{$\mathbf{E}^D$\xspace}

\newcommand{\clipfeats}[1]{$\mathbf{E}^{#1}$\xspace}
\newcommand{\costvolume}[1]{$\mathbf{V}^{#1}$\xspace}
\newcommand{\costemb}[1]{$\mathbf{C}^{#1}$\xspace}
\newcommand{\finalfeats}[1]{$\mathbf{F}^{#1}$\xspace}
\newcommand{\promptemb}[0]{$\mathbf{e}^T$\xspace}

\newcommand{\onedim}[1]{$\in \mathbb{R}^{#1}$\xspace}
\newcommand{\twodim}[2]{$\in \mathbb{R}^{#1\times #2}$\xspace}
\newcommand{\threedim}[3]{$\in \mathbb{R}^{#1\times #2\times #3}$\xspace}

\newcommand{\rgb}[1]{RGB$^{#1}$\xspace}
\newcommand{\depth}[1]{D$^{#1}$\xspace}
\newcommand{\pcd}[1]{P$^{#1}$\xspace}
\newcommand{\predmask}[1]{$\mathbf{M}^{#1}$\xspace}
\newcommand{\query}[0]{$Q$\xspace}
\newcommand{\anchor}[0]{$A$\xspace}
\newcommand{\object}[0]{$O$\xspace}
\newcommand{\prompt}[0]{$T$\xspace}

\newcommand{\objectpcd}[0]{$\mathbf{O}$\xspace}
\newcommand{\predpose}[0]{$\mathbf{T}$\xspace}
\newcommand{\gtpose}[0]{$\mathbf{\hat{T}}$\xspace}

\newcommand{\lowerbetter}[0]{{\color{black!50}{$\,\downarrow$}}}
\newcommand{\higherbetter}[0]{{\color{black!50}{$\,\uparrow$}}}
\newcommand{\oracle}[1]{\textcolor{gray}{#1}}

\definecolor{visual}{HTML}{3399FF}
\definecolor{text}{HTML}{97D077}

\definecolor{nocsbottle}{RGB}{31,119,180}
\definecolor{nocsbowl}{RGB}{255,127,14}
\definecolor{nocscamera}{RGB}{44,160,44}
\definecolor{nocscan}{RGB}{214,39,40}
\definecolor{nocslaptop}{RGB}{148,103,189}
\definecolor{nocsmug}{RGB}{140,86,75}
\newcommand{\impp}[1]{{\textcolor{Green}{+#1}}}
\newcommand{\impn}[1]{{\textcolor{BrickRed}{-#1}}}

\newcommand{\correct}[1]{{\textcolor{Green}{#1}}}
\newcommand{\wrong}[1]{{\textcolor{BrickRed}{#1}}}

\newcommand{\acronym}{Oryon\xspace}
\definecolor{myazure}{rgb}{0.8509,0.8980,0.9412}
\newcommand{\cmark}{\ding{51}}%
\newcommand{\xmark}{\ding{55}}%

\newcommand{\captionprompt}[3]{
    \centering \texttt{#1} \\
    \centering \texttt{\correct{#2}} \\
    \centering \texttt{\wrong{#3}}
}

%% file: supp/sections/0_introduction.tex
\section{Introduction}
\label{sec:supp_intro}

We provide additional material in support of our main paper. This document is organized as follows:
\begin{itemize}
    \item In Sec.~\ref{sec:supp_setting}, we compare the assumptions and requirements of \acronym{} with those of state-of-the-art methods that address pose estimation in a generalizable setting.
    \item In Sec.~\ref{sec:supp_implementation}, we provide additional implementation details and display the architecture of the fusion module (\fusion{}) in Sec.~\ref{sec:supp_fusion} and the decoder (\decoder{}) in Sec.~\ref{sec:supp_decoder}.
    \item In Sec.~\ref{sec:supp_dataset}, we outline the process we followed to generate our training and test datasets.
    \item In Sec.~\ref{sec:metrics} we extend the discussion on the pose metrics.
    \item In Sec.~\ref{sec:supp_prompts}, we list the textual prompts we used for evaluation on REAL275~\cite{nocs} and Toyota-Light~\cite{toyl} (TOYL for short), including the alternative prompts used in the ablation studies. We also present some examples of synonym sets from ShapeNet6D~\cite{he2022fs6d} (SN6D for short).
    \item In Sec.~\ref{sec:supp_qualitative}, we show additional qualitative results on pose estimation (Sec.~\ref{sec:supp_pose}) and segmentation (Sec.~\ref{sec:supp_segm}) on TOYL and REAL275. Moreover, we provide examples of feature distance visualization for \finalfeats{A} and \finalfeats{Q}, demonstrating how the features change when different prompts are used (Sec.~\ref{sec:supp_feats}).
\end{itemize}

%% file: supp/sections/1_comparison.tex
\section{Setting comparison}
\label{sec:supp_setting}

In Tab.~\ref{tab:supp_comparison}, we present a comparison of the data requirements of state-of-the-art methods for generalizable pose estimation. For each method, we report the input data, the reference used to estimate the pose, and whether the method can estimate the full pose or only the rotation component. 
We also summarize the object onboarding process, which has to be carried out at evaluation time for each novel object. 
Finally, we report the localization prior used, i.e., an external module that is used to crop or segment the object of interest. 
Note that some methods include the localization module within their pipeline (\cite{osop,gen6d}, and ours), instead of employing an external module.

We identify three main groups in the current literature. 
The first is model-based methods, which require a 3D model of the object at test time~\cite{ove6d,labbe2022megapose,osop}. 
When present, the object onboarding phase requires generating a set of templates (i.e., a set of views of the novel object) from the object 3D model~\cite{ove6d,osop}.

The second is model-free methods, which assume a video sequence of the object to be available~\cite{gen6d,onepose,oneposepp}. 
In these methods, the object onboarding phase is particularly cumbersome. 
Gen6D~\cite{gen6d} requires the user to perform SfM to reconstruct the object point cloud, which is then manually cropped and oriented. 
The resulting point cloud is used as an object model. 
OnePose~\cite{onepose} and OnePose++~\cite{oneposepp} also perform SfM on the video sequence, but they only use it to recover the camera poses, so the manual cropping of the point cloud is not needed. 
Therefore, the onboarding process of model-free methods requires human intervention, as opposed to that of model-based methods. 
OnePose and OnePose++ both adopt a prior detector (YOLOv5~\cite{yolov5}), while in Gen6D the localization is included in the pipeline.

The third group is composed of methods for relative pose estimation~\cite{nguyen2023nope,zhang2022relpose, lin2023relpose++,wang2023posediffusion}. 
These methods do not require any type of object model but instead rely on one or more support views of the novel object. 
In particular, NOPE~\cite{nguyen2023nope} adopts a single support view, while RelPose~\cite{zhang2022relpose} can work with two or more support views. 
Both methods rely on RGB data, and no geometric information is present. 
Because of this, they only estimate the relative rotation.
On the other hand, PoseDiffusion~\cite{wang2023posediffusion} and RelPose++~\cite{lin2023relpose++} assume that the scenes are object-centric, and enstablish a constraint on the distance of the first view from the origin of the coordinate system (e.g., the unit distance for RelPose++).
This allows both methods to estimate a relative translation, up to a global scale transformation of the scene.
Due to the object-centric assumption, both methods rely on an object detector to crop the object of interest.
A different approach is used in LatentFusion~\cite{park2020latentfusion}, in which a neural renderer is used to predict the depth and estimate the pose with respect to a set of reference views.
LatentFusion works with as less as a single reference view, and can estimate a complete pose in an absolute scale as also the depth information is provided.
Note that the original method requires the ground truth segmentation mask at both train and test time.

\acronym{} is fundamentally different from model-based methods as the object 3D model is not required. 
With respect to model-free methods instead, \acronym{} does not rely on complex object onboarding procedures, and it does not assume that the novel object is physically available so that a video sequence can be captured. 
Instead, to test on a novel object, only a textual description of it is required. 
Finally, compared to most relative pose estimation methods, \acronym{} is able to estimate the complete pose due to the requirement of depth information, including the absolute value of the translation component.
Moreover, \acronym{} does not rely on an external module for localization, as this step is carried out within the pipeline in an open-vocabulary fashion.

\input{supp/tables/comparison}

%% file: supp/tables/comparison.tex
\begin{table*}[t!]
\centering
\tabcolsep 3pt
\caption{
Comparison of the data requirements of \acronym with examples of state-of-the-art methods for generalizable pose estimation.
We classify the methods based on:
\textbf{Input}: the type of input data, typically RGB or RGBD; \textbf{Reference}: additional data used to identify the novel object at test time; \textbf{Full pose}: whether the method is capable of estimating the 6D pose or is limited to the rotation component; \textbf{Object onboarding}: eventual process required at test time to acquire data about the object to pose; \textbf{Localization prior}: eventual external modules used to localize the object, typically a segmentation mask or a detector.
}
\vspace{-3mm}
\label{tab:supp_comparison}
\resizebox{\textwidth}{!}{%
    \begin{tabular}{|l|l|l|c|l|l|}
        \toprule
        Method & Input & Reference & Full pose & Object onboarding & Localization prior \\
        \hline
        OVE6D~\cite{ove6d} & D & 3D model & \cmark & Generates and encodes 4K object templates & Segm. mask \\
        MegaPose~\cite{labbe2022megapose} & RGBD & 3D model & \cmark & - & Detector \\
        OSOP~\cite{osop} & RGB & 3D model & \cmark & Generates 90 object templates & - \\
        \hline
        Gen6D~\cite{gen6d} & RGB & Video sequence & \cmark & SfM and manual cropping of point cloud & - \\
        OnePose~\cite{onepose} & RGB & Video sequence & \cmark & SfM to retrieve camera viewpoints & Detector\\
        OnePose++~\cite{oneposepp} & RGB & Video sequence & \cmark & SfM to retrieve camera viewpoints & Detector\\
        \hline
        NOPE~\cite{nguyen2023nope} & RGB & Single supp. view & & - & - \\
        RelPose~\cite{zhang2022relpose} & RGB & Two or more supp. views & & - & - \\
        RelPose++~\cite{lin2023relpose++} & RGB & Two or more supp. views & \cmark & - & Detector \\
        PoseDiffusion~\cite{wang2023posediffusion} & RGB & Two or more supp. views & \cmark & - & Detector \\
        LatentFusion~\cite{park2020latentfusion} & RGBD & One or more supp. views & \cmark & - & Segm. mask \\
        \hline
        \acronym (ours) & RGBD & Single supp. view & \cmark & Expression of textual prompt & - \\
        \bottomrule
    \end{tabular}
}
\end{table*}

%% file: supp/sections/2_details.tex
\section{Implementation details}
\label{sec:supp_implementation}

\subsection{Text-visual fusion architecture}
\label{sec:supp_fusion}
In Fig.~\ref{fig:supp_diagram_fusion}, we show the details of a single layer of the fusion module \fusion{}. Note that the input visual feature map \clipfeats{} and the prompt features \promptemb have different feature sizes, as \clipfeats{} \threedim{1024}{24}{24}, while \promptemb{ \twodim{80}{768}, where 80 is the number of prompt templates. 
To compute the cost volume, we require the same feature dimension; therefore we apply a 2D convolution to \clipfeats{} to obtain a matrix \threedim{768}{24}{24}. 
The guidance backbone \guidance{} is used to output a feature map $\mathbf{G}$ \threedim{512}{24}{24}, which is projected to a feature map \threedim{128}{24}{24} before concatenating with the keys and queries of the two attention layers. 
We also project the prompt features \promptemb{} to a lower-dimensional space to obtain a feature list \twodim{80}{128}, which is concatenated to the visual feature map in the text guidance head. 
The text guidance head is used to inject prompt information into a lower-resolution feature map to enhance the semantic content of the features.

Note that the fusion module \fusion{} is composed of two of the layers depicted in Fig.~\ref{fig:supp_diagram_fusion}. 
Experimentally, we found beneficial to use the weights of CATSeg~\cite{cho2023catseg} for the window attention and shifted window attention parts of the module. 
In CATSeg, these two modules are followed by a Linear Transformer layer, which is used to model the relationship among the different classes in the textual prompt. 
Instead, in our setting, the prompt \prompt{} only describes a single class (i.e., object); therefore we replace the Linear Transformer layer with the text guidance head, which is trained from scratch.

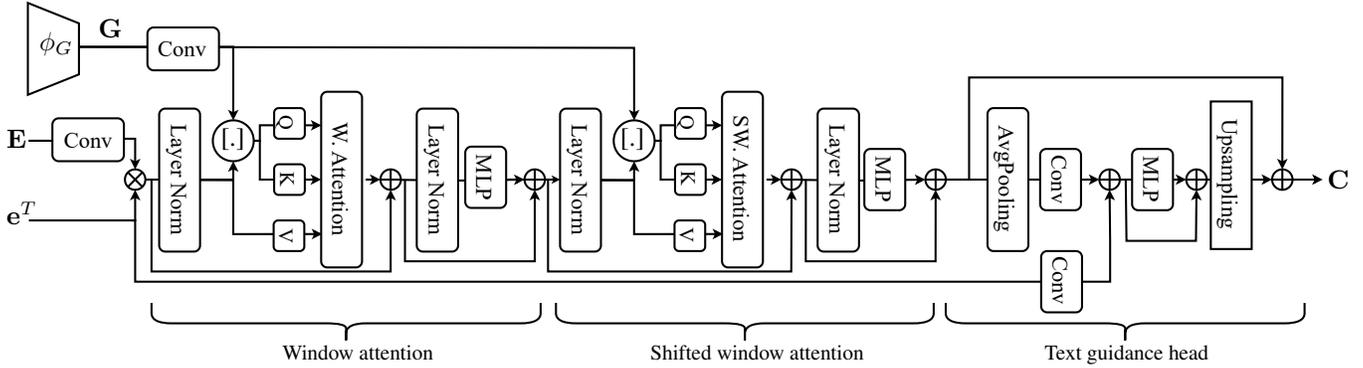
\begin{figure*}[]
    \centering
    \input{supp/figures/fusion}
    \caption{
        Architecture detail of a layer of the fusion module \fusion.
        \textbf{W. Attention}: attention layer that performs self-attention within windows, as in~\cite{liu2021Swin}; \textbf{SW. Attention}: attention layer that performs self-attention among shifted windows, as in~\cite{liu2021Swin}; \textbf{Upsampling}: upsampling by interpolation; $\mathbf{\oplus}$ denotes feature sum; $\mathbf{\otimes}$ denotes cost computation by cosine similarity; [.] denotes feature concatenation.
    }
    \label{fig:supp_diagram_fusion}
\end{figure*}

\subsection{Decoder architecture}
\label{sec:supp_decoder}

In Fig.~\ref{fig:supp_diagram_decoder}, we show the details of the decoder \decoder{}. 
A single decoder layer is composed by a transposed 2D convolution that doubles the spatial resolution of the input feature map. 
The result is concatenated with a projection of the feature map derived from the guidance backbone \guidance{}. 
We use $\mathbf{G}_1$ for the first layer and $\mathbf{G}_2$ for the second, while the third layer does not use any guidance. 
After the concatenation, we apply two blocks of operations, each composed of a 2D convolution followed by group normalization and ReLU activation. 
For better visualization, this group of operations is depicted as ConvGroup in Fig.~\ref{fig:supp_diagram_decoder}.

As for the fusion module \fusion{}, for the first two upsampling layers and the guidance projections, we finetune the weights provided by CATSeg~\cite{cho2023catseg}. 
We find this approach to be beneficial, as the upsampling layers are already trained for segmentation and therefore provide a good initialization to learn fine-grained features necessary for matching. 
The third upsampling layer is instead trained from scratch.

\begin{figure}[]
    \centering
    \input{supp/figures/decoder}
    \caption{
        Architecture detail of the decoder \decoder.
        \textbf{TransConv}: transposed 2D convolution; \textbf{ConvGroup}: group composed by two blocks, each contains a 2D convolution, a group normalization and a ReLU; [.] denotes feature concatenation.
    }
    \label{fig:supp_diagram_decoder}
\end{figure}
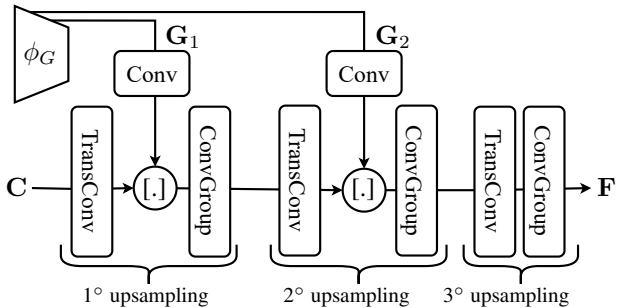

\subsection{Hardware and implementation}
\label{sec:supp_hardware}
We implement \acronym{} with PyTorch Lightning~\cite{lightning}. 
We train each model on four Nvidia V100 GPUs and set the batch size to 32. 
Training a single model for 20 epochs takes about 12 hours in our standard setting. 
We test on a single GPU of the same model; a complete evaluation on a dataset takes about one hour, including I/O operations and metric computations.

%% file: supp/figures/fusion.tex
\hspace*{0.00mm}
\begin{minipage}{\textwidth}
    \begin{overpic}[width=1\textwidth]{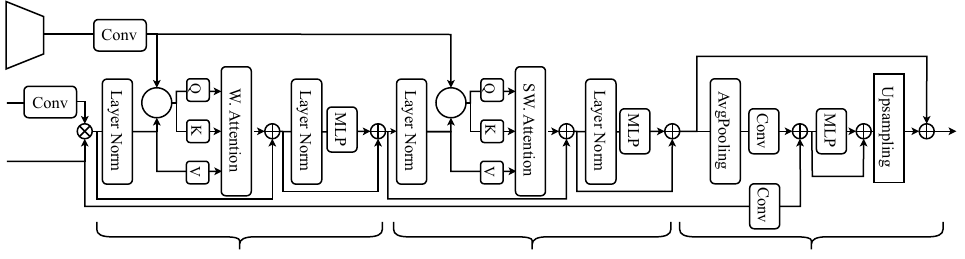}
        \put(6,24){$\mathbf{G}$}
        \put(99.5,12.5){\costemb{}}
        \put(1.5,23){\guidance}
        \put(-1,15.6){\clipfeats{}}
        \put(-1,9.8){\promptemb}
        \put(15.3,15.8){[.]}
        \put(45.7,15.8){[.]}
        \put(20,-0.5){\footnotesize Window attention}
        \put(48,-0.5){\footnotesize Shifted window attention}
        \put(78,-0.5){\footnotesize Text guidance head}
    \end{overpic}
\end{minipage}

%% file: supp/figures/decoder.tex
\hspace*{0.00mm}
\begin{minipage}{0.45\textwidth}
    \begin{overpic}[width=1\textwidth]{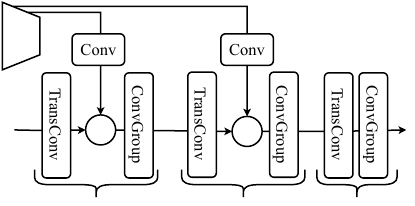}
        \put(2,40){\guidance}
        \put(26,42.8){$\mathbf{G}_1$}
        \put(61.5,42.8){$\mathbf{G}_2$}
        \put(-1,17){\costemb{}}
        \put(22.1,17){[.]}
        \put(57.4,17){[.]}
        \put(12,-1){\footnotesize 1$^{\circ}$ upsampling}
        \put(46,-1){\footnotesize 2$^{\circ}$ upsampling}
        \put(73,-1){\footnotesize 3$^{\circ}$ upsampling}
        \put(99,17){\finalfeats{}}
    \end{overpic}
\end{minipage}

%% file: supp/sections/3_dataset.tex
\section{Dataset generation}
\label{sec:supp_dataset}
In this section, we describe the process to generate the scene pairs for our training and test datasets. 
Given an object \object{} in two scenes \anchor{} and \query{}, let \pcd{A}, \pcd{Q} be the point clouds resulting from the unprojection of the two depth maps \depth{A}, \depth{Q}, and let \rgb{A}, \rgb{Q} be the respective RGB images. 
The point clouds \pcd{A}, \pcd{Q} are then filtered by using the ground-truth mask of \anchor{} and \query{} of the object \object{}, thus retaining only the points that belong to the object of interest in the two scenes. 
The objective is to find the transformation $T_{A\rightarrow Q}$ that aligns \pcd{A} to \pcd{Q}. 
Let $T_A$, $T_Q$ be the ground-truth pose of \object{} in \anchor{} and \query{} respectively.

In order to generate the poses required for the training process, the following strategy is adopted:
\begin{enumerate}
    \item Sample a random object \object{}, and a random scene \anchor{} in which \object{} is present.
    \item Consider all the images that contain \object{} that belong to a different scene from \anchor{}. Sample a random \query{} from this set.
    \item Given $T_A$ and $T_Q$, compute the relative pose $T_{A\rightarrow Q}$ = $T_Q (T_A)^{-1}$.
    \item Apply $T_{A\rightarrow Q}$ to $P_A$, thus aligning it to $P_Q$. Use the Nearest Neighbor algorithm to compute a set of 3D matches between $P_A$ and $P_Q$. To be considered a match, the Euclidean distance between two points must be $\leq 2$ mm.
    \item Use the intrinsic camera parameters to project the set of matches in 2D, thus obtaining the pixel-level correspondences $\textbf{x}^{A}$, $\textbf{x}^{Q}$ between \rgb{A} and \rgb{Q}.
\end{enumerate}

\noindent Note that the relative pose $T_{A\rightarrow Q}$ is only used for evaluation, as the training objectives are the ground-truth matches $\textbf{x}^{A}$, $\textbf{x}^{Q}$. 
The above process is repeated until the desired number of image pairs is obtained (i.e., 2K for REAL275 and TOYL, 20K for SN6D). 
Additionally, when generating the training set from SN6D, we reject all scene pairs with fewer than 100 ground-truth matches.

%% file: supp/sections/4_metrics.tex
\section{Pose metrics details}
\label{sec:metrics}

We adopt the Average Recall (AR)~\cite{bop-challenge} as main pose metric, which is defined as the average of three different metrics (VSD, MSSD, MSPD).
Additionally, we report the ADD(S)-0.1d~\cite{hodavn2016evaluation} and its definition.
In the following definitions, let \objectpcd \twodim{N}{3} be the point cloud representing the object model, and \gtpose, \predpose \twodim{3}{4} be the ground-truth and predicted 6D pose respectively.
Let also $d$ be the diameter of \objectpcd.

\noindent \textbf{Visible Surface Discrepancy (VSD)} compares two distance maps, obtained by rendering \objectpcd respectively with \gtpose and \predpose on the original depth of the image. 
The error $e_{VSD}$ is thus obtained as the mean of the distance map error, and the recall $\theta_{VSD}$ is defined as the mean of a set of recalls, in which the thresholds varies from 5\% to 50\% of $d$.
Finally, VSD is defined as the mean of a set of recalls on $\theta_{VSD}$, in which the threshold varies from 0.05 to 0.5.
Due to its definition, VSD is agnostic to object symmetry.

\noindent \textbf{Maximum Symmetry-Aware Surface Distance (MSSD)}. Consider the maximum distance (i.e.,  between any pair of points) between \objectpcd transformed with the ground-truth pose \gtpose and \objectpcd transformed with the predicted pose \predpose.
This error $e_{MSSD}$ is used to compute a set of recalls, in which the thresholds vary from 5\% to 50\% of $d$.
The average of the recall set is the final metric MSSD.
In order to take into account rotation to symmetry, $e_{MSSD}$ is computed for each symmetry axis of \objectpcd, and the smallest error is selected~\cite{bop-challenge}.

\noindent \textbf{Maximum Symmetry-Aware Projection Distance (MSPD)} is defined similarly to MSSD. Instead of computing the error in the 3D space, the transformed point clouds are projected in 2D before computing the maximum distance.
As for VSD and MSSD, the resulting error $e_{MSPD}$ is used to compute a set of recalls, whose thresholds vary from 5\% to 50\% of $w / 640$, where $w$ is the width of the image. 
As in the MSSD, $e_{MSPD}$ is computed with all the symmetry sets of \objectpcd, and the smallest distance is used to compute the recalls.

\noindent \textbf{ADD(S)-0.1d}~\cite{hodavn2016evaluation} (ADD for short) is quite different from the metrics defined in the BOP challenge. The error $e_{ADD}$ is defined as the mean distance between \objectpcd transformed with the ground-truth pose \gtpose and \objectpcd transformed with the predicted pose \predpose.
Instead of computing a mean of recalls, ADD is defined as a single recall of $e_{ADD}$, where the threshold is defined as 10\% of $d$.

The error measured by ADD is similar to the one of MSSD, but the first is more effective at measuring the success rate at a quite restrictive threshold (i.e., the cases in which the pose is very accurate).
On the other hand, MSSD is less sensible to noise as it is defined as a set of recalls, and is less dependent on object model shape as it measure the maximum distance instead of the mean~\cite{hodavn2016evaluation}.
MSPD shares similar characteristics of MSSD, while VSD is effective at measuring the depth distance, regardless of the object symmetry.
For this reasons, we adopt Average Recall (AR) as main metric, and we also report ADD(S)-0.1d due to its wide use in the pose estimation community~\cite{zebrapose,geometricaware6d}.

\noindent 

%% file: supp/sections/5_prompts.tex
\section{Prompt details}
\label{sec:supp_prompts}

In this section, we present the textual prompts used for all our tests with REAL275~\cite{nocs} and TOYL~\cite{toyl}, along with some examples of textual metadata from the training set, SN6D~\cite{he2022fs6d}. 
Note that we omit the prompt template used to augment the object descriptions. 
We adopt the template list used by CLIP~\cite{clip}, which includes 80 templates.

\subsection{REAL275 prompts}
\label{sec:supp_nocs}

\begin{figure*}[]
    \centering
    \input{supp/figures/prompts/nocs/prompts}
    \caption{
        Prompts used in the REAL275~\cite{nocs} dataset.
        For each object, we show the object name, the \correct{detailed description} (i.e., the default one), and the \wrong{misleading description} we adopt in the ablation study.
        All the objects are cropped for better visualization.
    }
    \label{fig:supp_nocs_prompts}
\end{figure*}
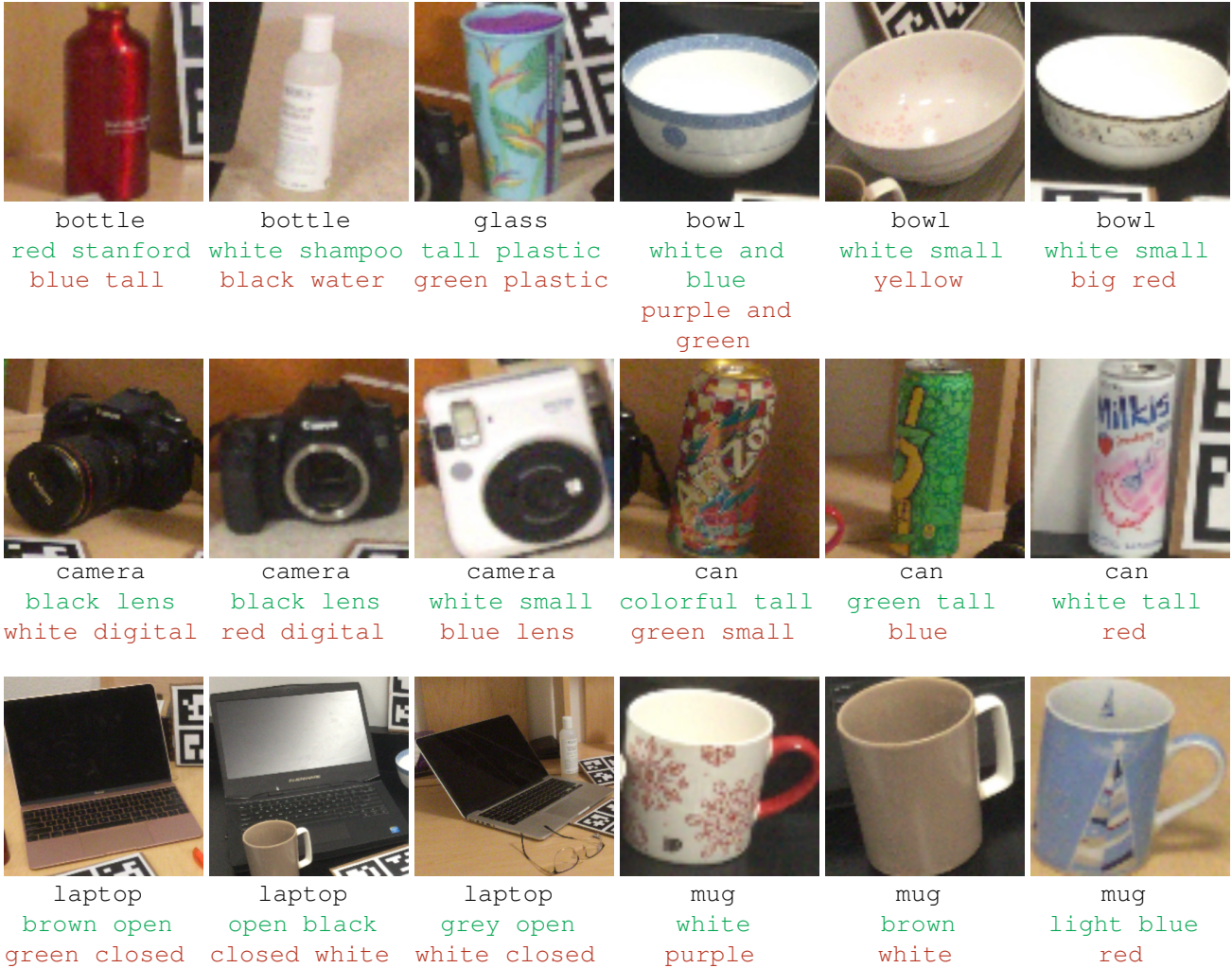

In Fig.~\ref{fig:supp_nocs_prompts}, we provide the exhaustive list of prompts used for each object model in REAL275. 
For each object, we show an example crop obtained from our test set and the textual information used to compose the prompt. 
We report in black the object name, in green the description used in our default setting, and in red the misleading description used in the ablation study.

\subsection{Toyota-Light prompts}
\label{sec:supp_toyl}

\begin{figure*}[]
    \centering
    \input{supp/figures/prompts/toyl/prompts}
    \vspace{-6mm}
    \caption{
        Prompts used in the TOYL~\cite{toyl} dataset.
        For each object, we show the object name and the \correct{detailed description} (i.e., the default one).
        All the objects are cropped for better visualization.
    }
    \label{fig:supp_toyl_prompts}
\end{figure*}
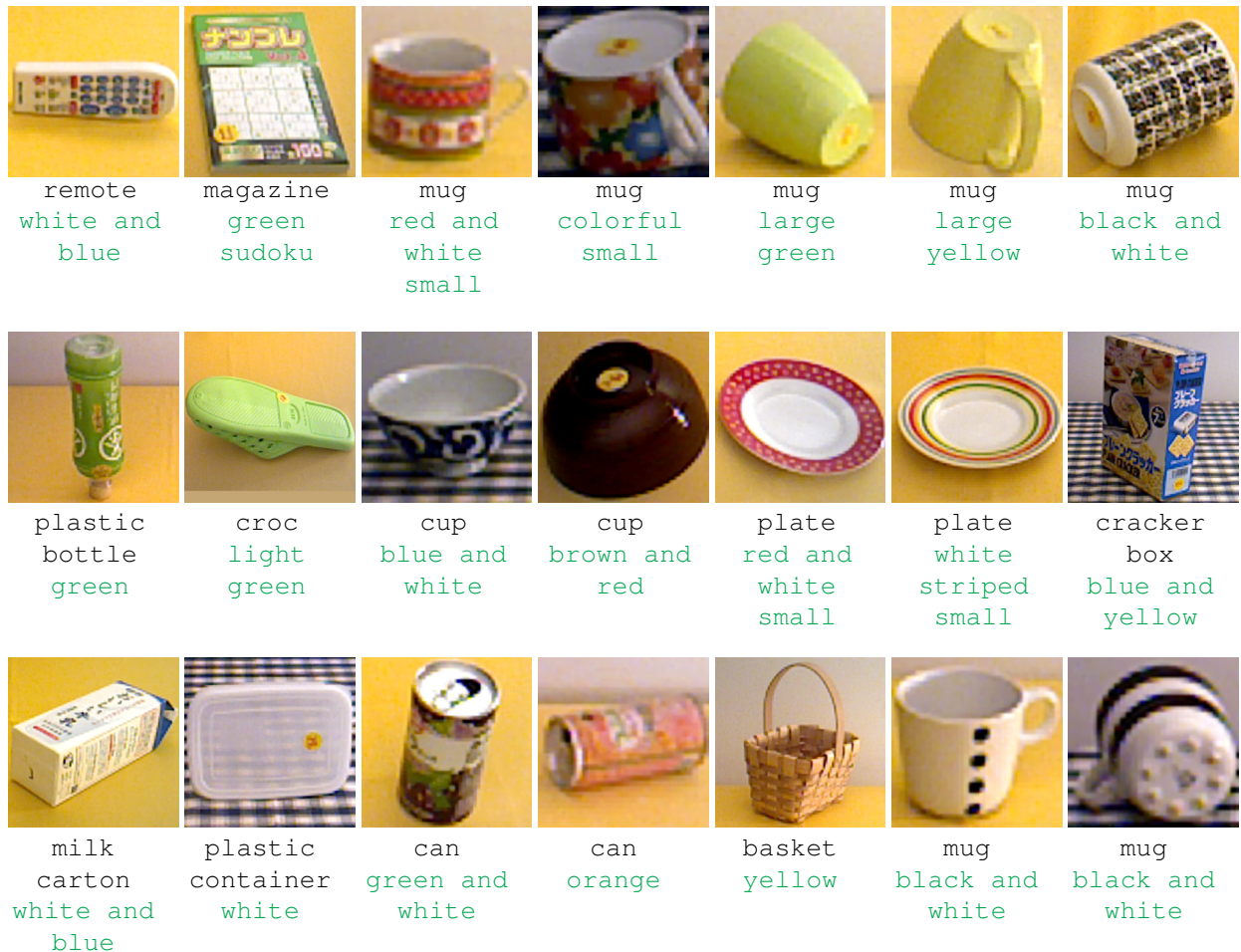

In Fig.~\ref{fig:supp_toyl_prompts}, we present the exhaustive list of prompts used for each object model in the TOYL. 
For each object, we show an example crop obtained from our test set and the textual information used to compose the prompt. 
We report in black the object name and in green the description used in our default setting. 
Note how, with respect to REAL275, the poses of the objects show more variety (e.g., objects tilted on one side or upside-down), as well as more object types.

\subsection{ShapeNet6D synsets}
\label{sec:supp_sn6d}

\begin{figure*}[]
    \centering
    \input{supp/figures/prompts/sn6d/prompts}
    \caption{
        Sample objects from the SN6D~\cite{he2022fs6d} dataset.
        For each object, we show the object default name, followed by its synonyms.
        Note that not all objects have an associated synonym set (e.g., the rifle, skateboard, and table objects).
        All the objects are cropped for better visualization, and padding is added when necessary.
    }
    \label{fig:supp_sn6d_prompts}
\end{figure*}
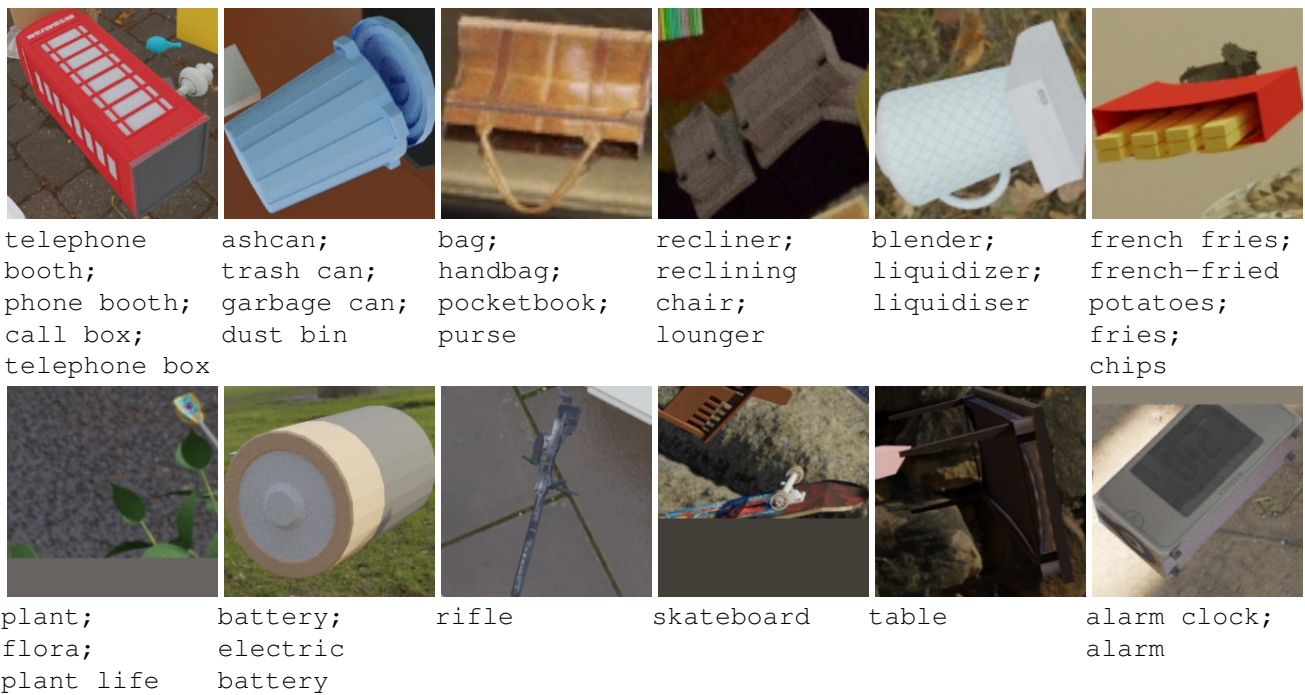

In Fig.~\ref{fig:supp_sn6d_prompts}, we showcase some examples of objects in the SN6D~\cite{he2022fs6d} dataset, along with the object names and the synsets (i.e., synonym sets) we adopt to build the prompt. 
For each object, we show an example crop obtained from the training set and the list of synonyms obtained from ShapeNetSem~\cite{shapenetsem}. 
The first name in the list is the default one, and the subsequent names (if present) are the synonyms. 
This dataset is completely synthetic and does not depict a realistic scenario, as all the objects are rendered in random poses and on a random background. 
Nonetheless, SN6D shows a wide variety of poses and objects, along with rich textual information useful for constructing the textual prompts.

%% file: supp/figures/prompts/nocs/prompts.tex
\hspace*{0.00mm}
\begin{minipage}{0.16\textwidth}
    \begin{overpic}[width=1\textwidth]{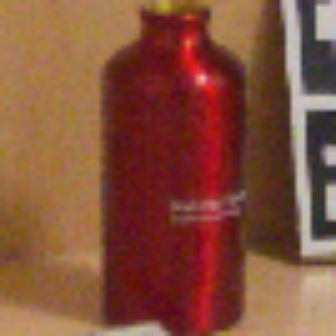}
    \end{overpic}
\end{minipage}
\begin{minipage}{0.16\textwidth}
    \begin{overpic}[width=1\textwidth]{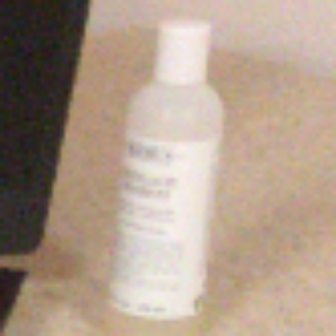}
    \end{overpic}
\end{minipage}
\begin{minipage}{0.16\textwidth}
    \begin{overpic}[width=1\textwidth]{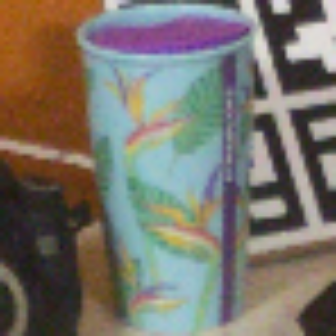}
    \end{overpic}
\end{minipage}
\begin{minipage}{0.16\textwidth}
    \begin{overpic}[width=1\textwidth]{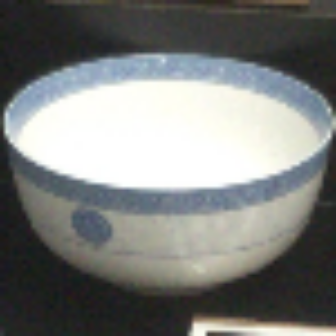}
    \end{overpic}
\end{minipage}
\begin{minipage}{0.16\textwidth}
    \begin{overpic}[width=1\textwidth]{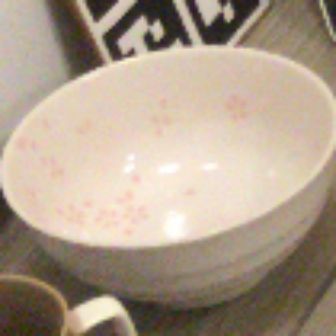}
    \end{overpic} 
\end{minipage}
\begin{minipage}{0.16\textwidth}
    \begin{overpic}[width=1\textwidth]{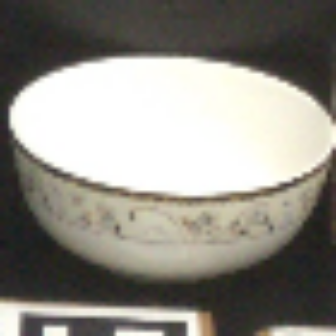}
    \end{overpic} 
\end{minipage}

\smallskip
\begin{minipage}[t]{0.16\textwidth}
    \captionprompt{bottle}{red stanford}{blue tall}
\end{minipage}
\begin{minipage}[t]{0.16\textwidth}
    \captionprompt{bottle}{white shampoo}{black water}
\end{minipage}
\begin{minipage}[t]{0.16\textwidth}
    \captionprompt{glass}{tall plastic}{green plastic}
\end{minipage}
\begin{minipage}[t]{0.16\textwidth}
    \captionprompt{bowl}{white and blue}{purple and green}
\end{minipage}
\begin{minipage}[t]{0.16\textwidth}
    \captionprompt{bowl}{white small}{yellow}
\end{minipage}
\begin{minipage}[t]{0.16\textwidth}
    \captionprompt{bowl}{white small}{big red}
\end{minipage}

\smallskip
\hspace*{0.00mm}
\begin{minipage}{0.16\textwidth}
    \begin{overpic}[width=1\textwidth]{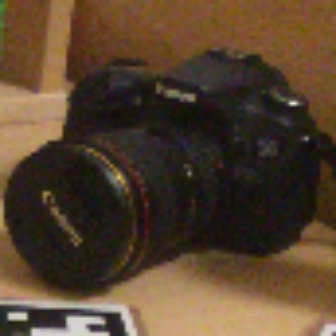}
    \end{overpic}
\end{minipage}
\begin{minipage}{0.16\textwidth}
    \begin{overpic}[width=1\textwidth]{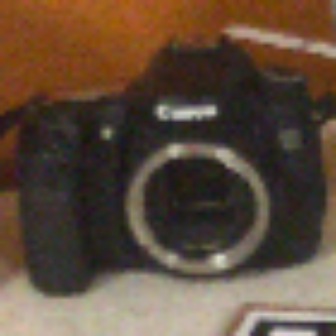}
    \end{overpic}
\end{minipage}
\begin{minipage}{0.16\textwidth}
    \begin{overpic}[width=1\textwidth]{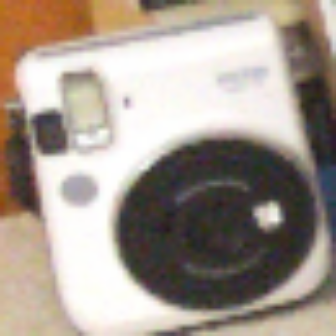}
    \end{overpic}
\end{minipage}
\begin{minipage}{0.16\textwidth}
    \begin{overpic}[width=1\textwidth]{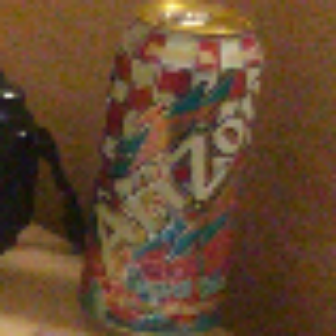}
    \end{overpic}
\end{minipage}
\begin{minipage}{0.16\textwidth}
    \begin{overpic}[width=1\textwidth]{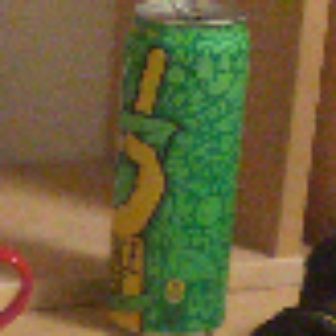}
    \end{overpic}
\end{minipage}
\begin{minipage}{0.16\textwidth}
    \begin{overpic}[width=1\textwidth]{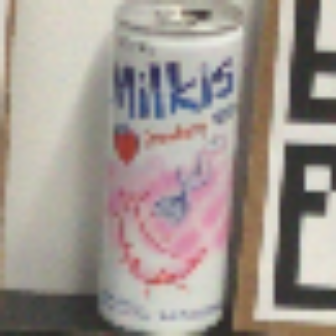}
    \end{overpic}
\end{minipage}

\smallskip
\begin{minipage}[t]{0.16\textwidth}
    \captionprompt{camera}{black lens}{white digital}
\end{minipage}
\begin{minipage}[t]{0.16\textwidth}
    \captionprompt{camera}{black lens}{red digital}
\end{minipage}
\begin{minipage}[t]{0.16\textwidth}
    \captionprompt{camera}{white small}{blue lens}
\end{minipage}
\begin{minipage}[t]{0.16\textwidth}
    \captionprompt{can}{colorful tall}{green small}
\end{minipage}
\begin{minipage}[t]{0.16\textwidth}
    \captionprompt{can}{green tall}{blue} 
\end{minipage}
\begin{minipage}[t]{0.16\textwidth}
    \captionprompt{can}{white tall}{red}
\end{minipage}

\smallskip
\hspace*{0.00mm}
\begin{minipage}{0.16\textwidth}
    \begin{overpic}[width=1\textwidth]{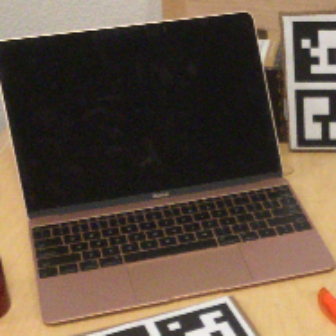}
    \end{overpic}
\end{minipage}
\begin{minipage}{0.16\textwidth}
    \begin{overpic}[width=1\textwidth]{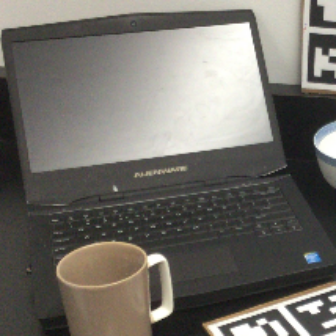}
    \end{overpic}
\end{minipage}
\begin{minipage}{0.16\textwidth}
    \begin{overpic}[width=1\textwidth]{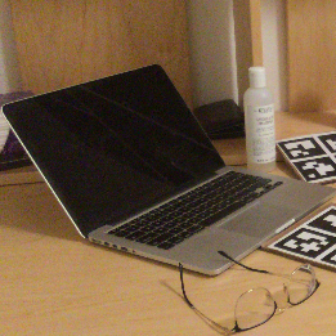}
    \end{overpic}
\end{minipage}
\begin{minipage}{0.16\textwidth}
    \begin{overpic}[width=1\textwidth]{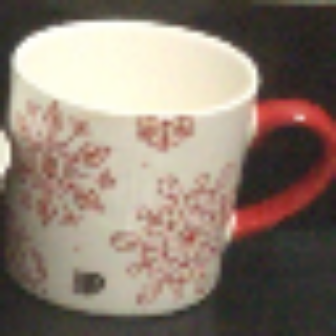}
    \end{overpic}
\end{minipage}
\begin{minipage}{0.16\textwidth}
    \begin{overpic}[width=1\textwidth]{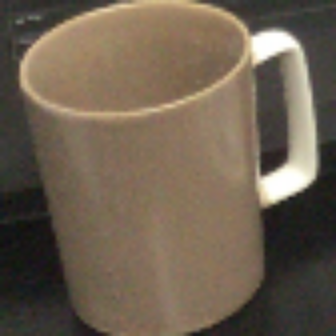}
    \end{overpic}
\end{minipage}
\begin{minipage}{0.16\textwidth}
    \begin{overpic}[width=1\textwidth]{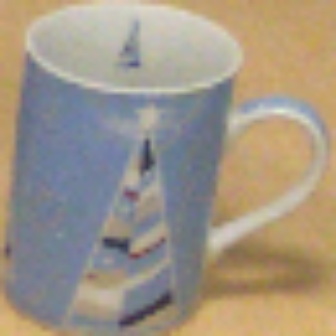}
    \end{overpic}
\end{minipage}

\smallskip
\begin{minipage}[t]{0.16\textwidth}
    \captionprompt{laptop}{brown open}{green closed}
\end{minipage}
\begin{minipage}[t]{0.16\textwidth}
    \captionprompt{laptop}{open black}{closed white}
\end{minipage}
\begin{minipage}[t]{0.16\textwidth}
    \captionprompt{laptop}{grey open}{white closed}
\end{minipage}
\begin{minipage}[t]{0.16\textwidth}
    \captionprompt{mug}{white}{purple}
\end{minipage}
\begin{minipage}[t]{0.16\textwidth}
    \captionprompt{mug}{brown}{white}
\end{minipage}
\begin{minipage}[t]{0.16\textwidth}
    \captionprompt{mug}{light blue}{red}
\end{minipage}

%% file: supp/figures/prompts/toyl/prompts.tex
\hspace*{0.00mm}
\begin{minipage}{0.13\textwidth}
    \begin{overpic}[width=1\textwidth]{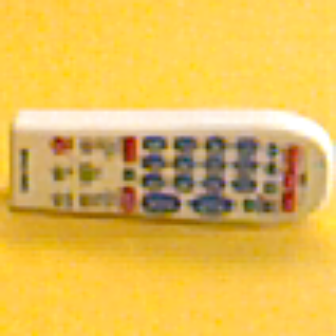}
    \end{overpic}
\end{minipage}
\begin{minipage}{0.13\textwidth}
    \begin{overpic}[width=1\textwidth]{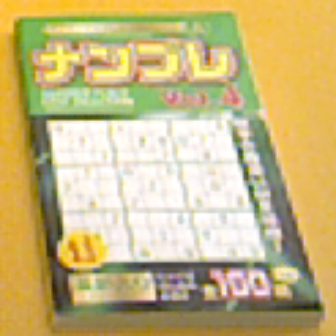}
    \end{overpic}
\end{minipage}
\begin{minipage}{0.13\textwidth}
    \begin{overpic}[width=1\textwidth]{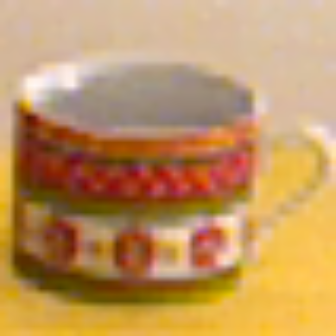}
    \end{overpic}
\end{minipage}
\begin{minipage}{0.13\textwidth}
    \begin{overpic}[width=1\textwidth]{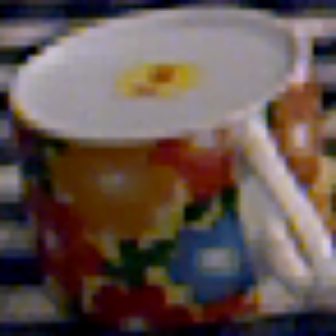}
    \end{overpic}
\end{minipage}
\begin{minipage}{0.13\textwidth}
    \begin{overpic}[width=1\textwidth]{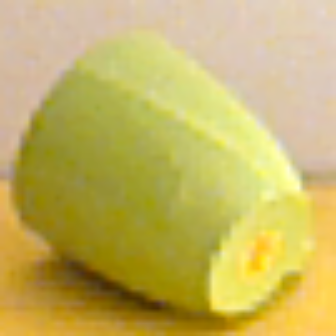}
    \end{overpic} 
\end{minipage}
\begin{minipage}{0.13\textwidth}
    \begin{overpic}[width=1\textwidth]{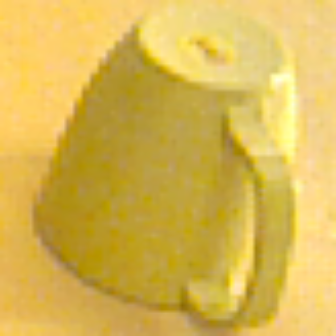}
    \end{overpic} 
\end{minipage}
\begin{minipage}{0.13\textwidth}
    \begin{overpic}[width=1\textwidth]{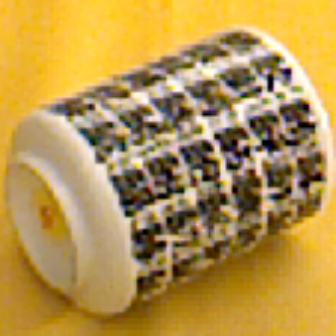}
    \end{overpic} 
\end{minipage}
\smallskip
\begin{minipage}[t]{0.13\textwidth}
  \captionprompt{remote}{white and blue}{}
\end{minipage}
\begin{minipage}[t]{0.13\textwidth}
  \captionprompt{magazine}{green sudoku}{}
\end{minipage}
\begin{minipage}[t]{0.13\textwidth}
  \captionprompt{mug}{red and white small}{}
\end{minipage}
\begin{minipage}[t]{0.13\textwidth}
  \captionprompt{mug}{colorful small}{}
\end{minipage}
\begin{minipage}[t]{0.13\textwidth}
  \captionprompt{mug}{large green}{}
\end{minipage}
\begin{minipage}[t]{0.13\textwidth}
  \captionprompt{mug}{large yellow}{}
\end{minipage}
\begin{minipage}[t]{0.13\textwidth}
  \captionprompt{mug}{black and white}{}
\end{minipage}

\hspace*{0.00mm}
\begin{minipage}{0.13\textwidth}
    \begin{overpic}[width=1\textwidth]{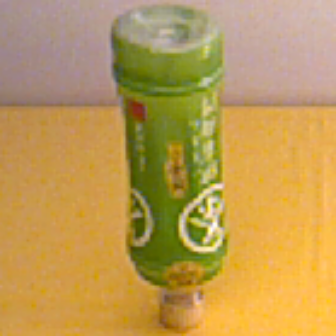}
    \end{overpic}
\end{minipage}
\begin{minipage}{0.13\textwidth}
    \begin{overpic}[width=1\textwidth]{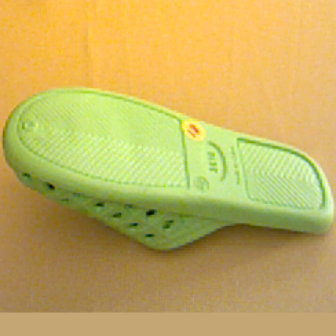}
    \end{overpic}
\end{minipage}
\begin{minipage}{0.13\textwidth}
    \begin{overpic}[width=1\textwidth]{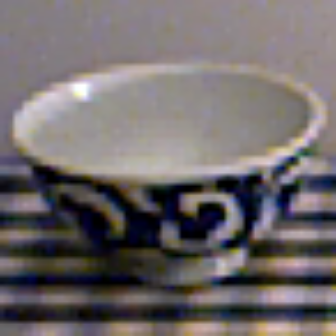}
    \end{overpic}
\end{minipage}
\begin{minipage}{0.13\textwidth}
    \begin{overpic}[width=1\textwidth]{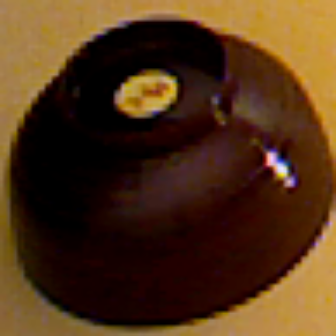}
    \end{overpic}
\end{minipage}
\begin{minipage}{0.13\textwidth}
    \begin{overpic}[width=1\textwidth]{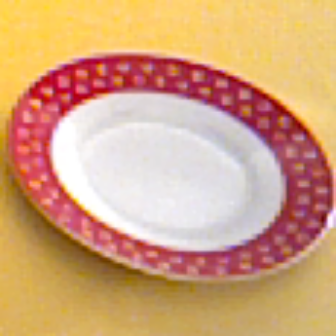}
    \end{overpic}
\end{minipage}
\begin{minipage}{0.13\textwidth}
    \begin{overpic}[width=1\textwidth]{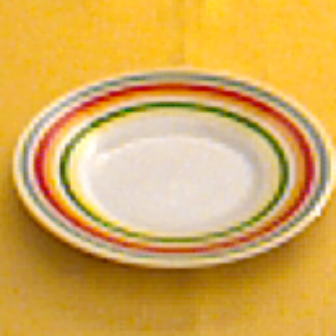}
    \end{overpic}
\end{minipage}
\begin{minipage}{0.13\textwidth}
    \begin{overpic}[width=1\textwidth]{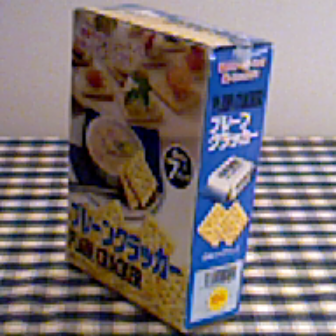}
    \end{overpic}
\end{minipage}

\smallskip
\begin{minipage}[t]{0.13\textwidth}
  \captionprompt{plastic bottle}{green}{}
\end{minipage}
\begin{minipage}[t]{0.13\textwidth}
  \captionprompt{croc}{light green}{}
\end{minipage}
\begin{minipage}[t]{0.13\textwidth}
  \captionprompt{cup}{blue and white}{}
\end{minipage}
\begin{minipage}[t]{0.13\textwidth}
  \captionprompt{cup}{brown and red}{}
\end{minipage}
\begin{minipage}[t]{0.13\textwidth}
  \captionprompt{plate}{red and white small}{}
\end{minipage}
\begin{minipage}[t]{0.13\textwidth}
  \captionprompt{plate}{white striped small}{}
\end{minipage}
\begin{minipage}[t]{0.13\textwidth}
  \captionprompt{cracker box}{blue and yellow}{}
\end{minipage}

\hspace*{0.00mm}
\begin{minipage}{0.13\textwidth}
    \begin{overpic}[width=1\textwidth]{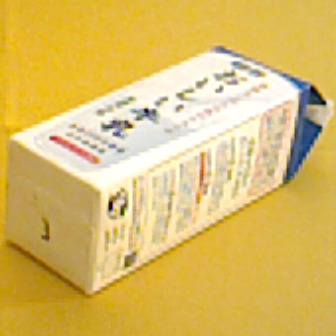}
    \end{overpic}
\end{minipage}
\begin{minipage}{0.13\textwidth}
    \begin{overpic}[width=1\textwidth]{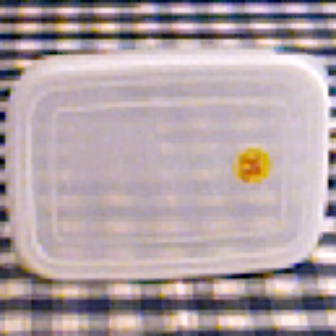}
    \end{overpic}
\end{minipage}
\begin{minipage}{0.13\textwidth}
    \begin{overpic}[width=1\textwidth]{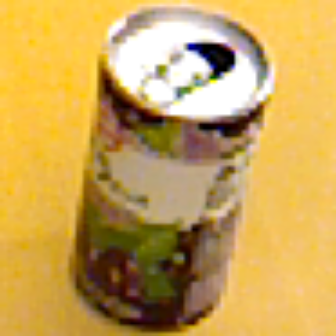}
    \end{overpic}
\end{minipage}
\begin{minipage}{0.13\textwidth}
    \begin{overpic}[width=1\textwidth]{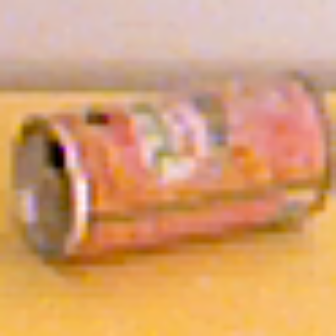}
    \end{overpic}
\end{minipage}
\begin{minipage}{0.13\textwidth}
    \begin{overpic}[width=1\textwidth]{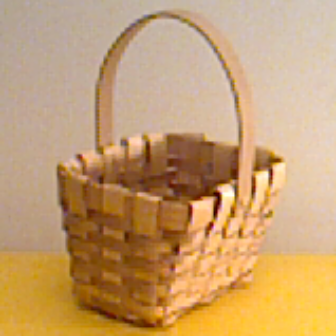}
    \end{overpic}
\end{minipage}
\begin{minipage}{0.13\textwidth}
    \begin{overpic}[width=1\textwidth]{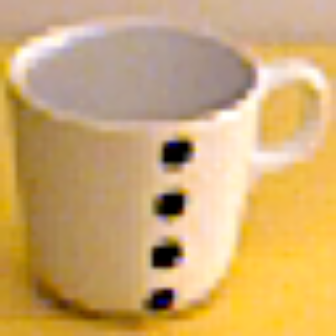}
    \end{overpic}
\end{minipage}
\begin{minipage}{0.13\textwidth}
    \begin{overpic}[width=1\textwidth]{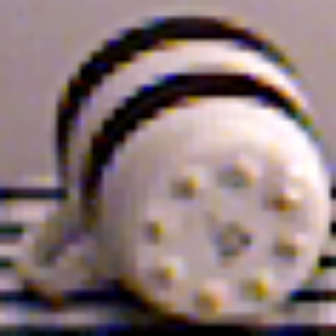}
    \end{overpic}
\end{minipage}

\smallskip
\begin{minipage}[t]{0.13\textwidth}
  \captionprompt{milk carton}{white and blue}{}
\end{minipage}
\begin{minipage}[t]{0.13\textwidth}
  \captionprompt{plastic container}{white}{}
\end{minipage}
\begin{minipage}[t]{0.13\textwidth}
  \captionprompt{can}{green and white}{}
\end{minipage}
\begin{minipage}[t]{0.13\textwidth}
  \captionprompt{can}{orange}{}
\end{minipage}
\begin{minipage}[t]{0.13\textwidth}
  \captionprompt{basket}{yellow}{}
\end{minipage}
\begin{minipage}[t]{0.13\textwidth}
  \captionprompt{mug}{black and white}{}
\end{minipage}
\begin{minipage}[t]{0.13\textwidth}
  \captionprompt{mug}{black and white}{}
\end{minipage}

%% file: supp/figures/prompts/sn6d/prompts.tex
\hspace*{0.00mm}
\begin{minipage}{0.16\textwidth}
    \begin{overpic}[width=1\textwidth]{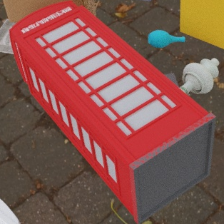}
    \end{overpic}
\end{minipage}
\begin{minipage}{0.16\textwidth}
    \begin{overpic}[width=1\textwidth]{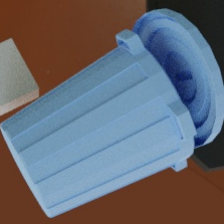}
    \end{overpic}
\end{minipage}
\begin{minipage}{0.16\textwidth}
    \begin{overpic}[width=1\textwidth]{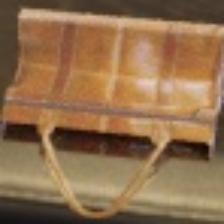}
    \end{overpic}
\end{minipage}
\begin{minipage}{0.16\textwidth}
    \begin{overpic}[width=1\textwidth]{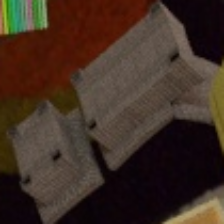}
    \end{overpic}
\end{minipage}
\begin{minipage}{0.16\textwidth}
    \begin{overpic}[width=1\textwidth]{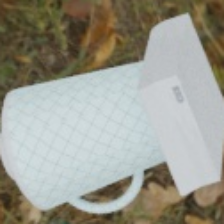}
    \end{overpic} 
\end{minipage}
\begin{minipage}{0.16\textwidth}
    \begin{overpic}[width=1\textwidth]{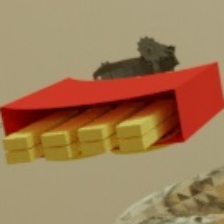}
    \end{overpic} 
\end{minipage}

\smallskip
\begin{minipage}[t]{0.16\textwidth}
    \texttt{telephone booth; \\ phone booth; \\ call box; \\ telephone box}
\end{minipage}
\begin{minipage}[t]{0.16\textwidth}
    \texttt{ashcan; \\ trash can; \\ garbage can; \\ dust bin}
\end{minipage}
\begin{minipage}[t]{0.16\textwidth}
    \texttt{bag; \\ handbag; \\ pocketbook; \\ purse}
\end{minipage}
\begin{minipage}[t]{0.16\textwidth}
    \texttt{recliner; \\ reclining chair; \\ lounger}
\end{minipage}
\begin{minipage}[t]{0.16\textwidth}
    \texttt{blender; \\ liquidizer; \\ liquidiser}
\end{minipage}
\begin{minipage}[t]{0.16\textwidth}
    \texttt{french fries; \\ french-fried potatoes; \\ fries; \\ chips}
\end{minipage}

\smallskip
\hspace*{0.00mm}
\begin{minipage}{0.16\textwidth}
    \begin{overpic}[width=1\textwidth]{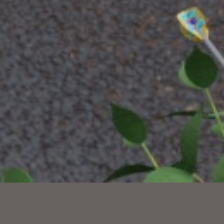}
    \end{overpic}
\end{minipage}
\begin{minipage}{0.16\textwidth}
    \begin{overpic}[width=1\textwidth]{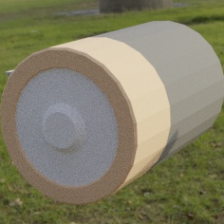}
    \end{overpic}
\end{minipage}
\begin{minipage}{0.16\textwidth}
    \begin{overpic}[width=1\textwidth]{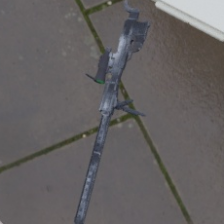}
    \end{overpic}
\end{minipage}
\begin{minipage}{0.16\textwidth}
    \begin{overpic}[width=1\textwidth]{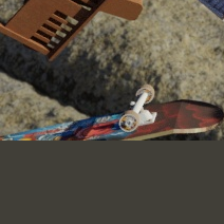}
    \end{overpic}
\end{minipage}
\begin{minipage}{0.16\textwidth}
    \begin{overpic}[width=1\textwidth]{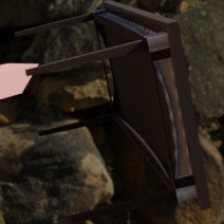}
    \end{overpic}
\end{minipage}
\begin{minipage}{0.16\textwidth}
    \begin{overpic}[width=1\textwidth]{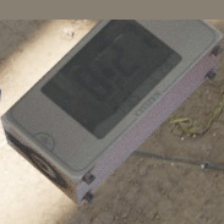}
    \end{overpic}
\end{minipage}

\smallskip
\begin{minipage}[t]{0.16\textwidth}
    \texttt{plant; \\ flora; \\ plant life}
\end{minipage}
\begin{minipage}[t]{0.16\textwidth}
    \texttt{battery; \\ electric battery}
\end{minipage}
\begin{minipage}[t]{0.16\textwidth}
    \texttt{rifle}
\end{minipage}
\begin{minipage}[t]{0.16\textwidth}
    \texttt{skateboard}
\end{minipage}
\begin{minipage}[t]{0.16\textwidth}
    \texttt{table} 
\end{minipage}
\begin{minipage}[t]{0.16\textwidth}
    \texttt{alarm clock; \\ alarm}
\end{minipage}

%% file: supp/sections/6_qualitative.tex
\section{Additional qualitative results}
\label{sec:supp_qualitative}

In Sec.~\ref{sec:supp_pose}, we report additional qualitative results on pose estimation on the test datasets. 
In Sec.~\ref{sec:supp_segm}, we show qualitative results of the segmentation mask obtained by \acronym in our best setting and compare them with the ground-truth masks and the ones predicted by OVSeg~\cite{liang2023ovseg}.

\subsection{Pose results}
\label{sec:supp_pose}

In Fig.~\ref{fig:supp_nocs_pose}, we show qualitative results of the object poses on REAL275~\cite{nocs}. 
\acronym is generally effective at localizing the object of interest, while the main source of errors in the pose is related to the rotation components (see Fig.~\ref{fig:supp_nocs_pose}(a,c,f)). 
In all these cases, the object appears small, and therefore the retrieved matches are coarse-grained. 
On the other hand, ObjectMatch~\cite{gumeli2023objectmatch} and SIFT~\cite{lowe1999sift} show larger errors, often related to the translation components (Fig.~\ref{fig:supp_nocs_pose}(a,b,d) for ObjectMatch, and Fig.~\ref{fig:supp_nocs_pose}(e) for SIFT). 
Note that all methods in this visualization use our segmentation mask; therefore, translation errors are due to spurious matches rather than an incorrect mask.

\begin{figure*}[]
    \centering
    \input{supp/figures/qualitative/pose_nocs/pose}
    \vspace{-4mm}
    \caption{
    Examples of qualitative pose results from the REAL275~\cite{nocs} dataset.
    All the results use the segmentation mask predicted by \acronym.
    We show the object model colored by mapping its 3D coordinates to the RGB space.
    Query images are darkened to highlight the object poses.
    }
    \label{fig:supp_nocs_pose}
\end{figure*}

In Fig.~\ref{fig:supp_toyl_pose}, we show qualitative results of the object poses on TOYL~\cite{toyl}. 
As in REAL275, \acronym localization on TOYL is quite accurate. 
Rotation errors are still present and are related to small objects (see Fig.~\ref{fig:supp_toyl_pose}(a, d, f)), which in TOYL are more common due to higher variation in poses compared to REAL275. 
We observe that ObjectMatch's effectiveness in this dataset is limited: in some cases, large errors cause the projected object to fall outside the image plane (Fig.~\ref{fig:supp_toyl_pose}(a, b)), while in other cases, significant translation errors are present (Fig.~\ref{fig:supp_toyl_pose}(d, e, f)). 
As observed in the main results, the high variation in light conditions of TOYL is not tolerated by ObjectMatch, which fails to find accurate matches. On the other hand, SIFT performances are closer to ours. 
Localization is more effective than ObjectMatch (Fig.~\ref{fig:supp_toyl_pose}(a, b, c, d, f)), but rotation errors are still present (Fig.~\ref{fig:supp_toyl_pose}(a, b, f)). 
The scale invariance properties of SIFT make it more effective for this scenario.

\begin{figure*}[]
    \centering
    \input{supp/figures/qualitative/pose_toyl/pose}
    \vspace{-4mm}
    \caption{
    Examples of qualitative pose results from the TOYL~\cite{toyl} dataset.
    All the results use the segmentation mask predicted by \acronym.
    We show the object model colored by mapping its 3D coordinates to the RGB space.
    Query images are darkened to highlight the object poses.
    }
    \label{fig:supp_toyl_pose}
\end{figure*}

\subsection{Segmentation results}
\label{sec:supp_segm}

In Fig.~\ref{fig:supp_nocs_segm}, we show some qualitative results of the segmentation masks on REAL275~\cite{nocs}. 
On this dataset, \acronym outperforms OVSeg~\cite{liang2023ovseg} by 10.1 points in mean Intersection-over-Union (mIoU). 
In Fig.~\ref{fig:supp_nocs_segm}, we can observe that the masks predicted by \acronym are generally coarser than the ones predicted by OVSeg. This is caused by the lower resolution we adopt for the segmentation mask (192$\times$192), while OVSeg generates the masks using the original image resolution (480$\times$640). 
There are cases in which this results in a clearly better performance for OVSeg (Fig.~\ref{fig:supp_nocs_segm}(a, d)). 
However, compared to \acronym, false positives are much more frequent in OVSeg: the object is completely missed in Fig.~\ref{fig:supp_nocs_segm}(c, e, f), while in Fig.~\ref{fig:supp_nocs_segm}(b) the whole table is selected in the anchor scene, and the wrong object is segmented in the query one.

On the other hand, \acronym is less prone to false negatives: most errors are instead due to the inclusion of background in the mask (Fig.~\ref{fig:supp_nocs_segm}(b)) or segmenting only a part of the object (Fig.~\ref{fig:supp_nocs_segm}(d)). 
Both these errors can still provide a sufficient number of valid matches to perform registration, while a false positive results in an automatic failure in the pose estimation task. 
Note the particular case of Fig.~\ref{fig:supp_nocs_segm}(f), in which two mugs are present in the anchor image. 
This causes an ambiguity, and results in a separate mask for each mug. 
Similarly, in the query image of Fig.~\ref{fig:supp_nocs_segm}(c), the wrong mug is segmented.

\begin{figure*}[]
    \centering
    \input{supp/figures/qualitative/segm_nocs/segm}
    \vspace{-3mm}
    \caption{
    Examples of qualitative segmentation results from the REAL275~\cite{nocs} dataset.
    Images are darkened to highlight the masks.
    }
    \label{fig:supp_nocs_segm}
\end{figure*}

In Fig.~\ref{fig:supp_toyl_segm}, we show some qualitative results of the segmentation masks on TOYL~\cite{toyl}. 
On this dataset, \acronym is outperformed by OVSeg~\cite{liang2023ovseg} by 7.4 mIoU. 
Compared to REAL275, TOYL does not show additional objects other than the object of interest; therefore it is an easier dataset for the segmentation task. 
We observe that both \acronym and OVSeg can correctly locate the objects of interest, and all the errors in the examples are due to imprecision in the segmentation mask, instead of false positives or false negatives. 
Similarly to REAL275, OVSeg masks are generally more accurate than the ones predicted by \acronym. 
The higher resolution of OVSeg is an important advantage in this dataset, as the object can appear very small due to perspective (see Fig.~\ref{fig:supp_toyl_segm}(a, b, c)). 
Most of the errors in \acronym are due to partial object segmentation (Fig.~\ref{fig:supp_toyl_segm}(b, d, f)) and background inclusion in the mask (see Fig.~\ref{fig:supp_toyl_segm}(a, e)).

\begin{figure*}[]
    \centering
    \input{supp/figures/qualitative/segm_toyl/segm}
    \caption{
    Examples of qualitative segmentation results from the TOYL~\cite{toyl} dataset.
    Images are darkened to highlight the masks.
    }
    \label{fig:supp_toyl_segm}
\end{figure*}

\subsection{Feature visualization}
\label{sec:supp_feats}

\begin{figure*}[]
    \centering
    \input{supp/figures/qualitative/feat_nocs/feats}
    \caption{
    Visualization of the feature distance in \finalfeats{A}, \finalfeats{Q} with respect to a reference point on the object (denoted as {\color{black!10!green}$\bullet$} in the anchor image) on REAL275~\cite{nocs}.
    For each sample, we test three different prompts we adopted in the ablation study:
    \textbf{Standard prompt}: the standard prompt we adopt, composed of a description and the object name;
    \textbf{Name only}: the standard prompt without the description;
    \textbf{Misleading desc.}: the standard description is replaced by a misleading one (see Fig.~\ref{fig:supp_nocs_prompts}).
    The RGB images are modified to highlight the object of interest.
    }
    \label{fig:supp_feats}
\end{figure*}
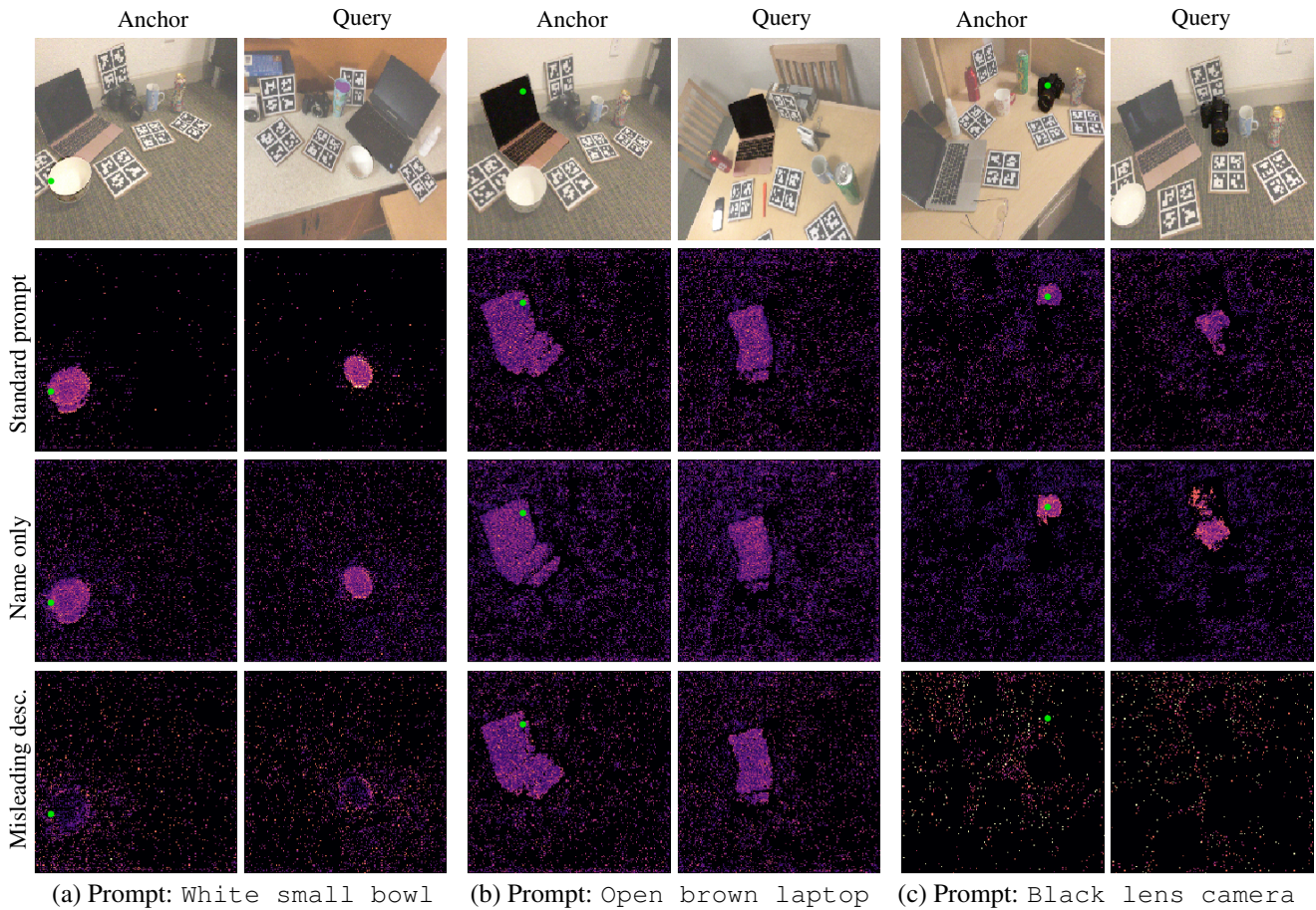

In Fig.~\ref{fig:supp_feats}, we present some examples of the visualization of the feature distance in the feature maps \finalfeats{A}, \finalfeats{Q} obtained by \acronym. 
For each image pair, we first sample a random reference point from the ground-truth segmentation mask on the anchor image (shown in green). 
We then compute the cosine similarity between the feature at the reference point and all the other features, both on the anchor and on the query image. 
In this way, we obtain a distance measure for each pixel, which is normalized across the pair and mapped to the RGB space. 
For each example, we show three versions of the feature maps obtained with the prompts we adopted in the ablation study in the main paper (i.e., the standard prompt, the prompt with only the object name, and the prompt with a misleading description).

When the standard prompt is adopted (first row), we can observe a sharp difference between the features of the background and the ones on the objects. Note also how in Fig.~\ref{fig:supp_feats}(a), the most similar points to the reference one on the query image are on the bowl border, as the reference point itself. 
When the object description is removed (second row), we can observe noticeable changes in Fig.~\ref{fig:supp_feats}(a), in which the background appears more noisy, and in Fig.~\ref{fig:supp_feats}(c), where a set of similar features appears on top of the camera in the query image. 
Instead, in Fig.~\ref{fig:supp_feats}(b), the difference with the distances of the original prompt is less relevant. 
As we reported in the ablation study, this alternative prompt causes a small drop of 2.2 in AR and of 3.1 in mIoU. 
Finally, in the last row, a misleading description is used in the prompt. 
This causes a clear drop in feature quality in Fig.~\ref{fig:supp_feats}(a, c): in the first example, the object outline is still visible due to similarity in the borders of the bowl, and in the second example only spurious similar features are present. 
Instead, the laptop object in Fig.~\ref{fig:supp_feats}(b) is less affected. 
Note that in the ablation study, the misleading prompt causes a drop of 6.8 and 10.1 in AR and mIoU, respectively.
These results suggest that the use of a misleading or more generic prompt can impact \acronym performances differently depending on the object of interest.

%% file: supp/figures/qualitative/pose_nocs/pose.tex
\hspace*{0.00mm}
\begin{minipage}{0.195\textwidth}
    \centering{\small{Anchor}}
\end{minipage}
\begin{minipage}{0.195\textwidth}
    \centering{\small{Ground truth}}
\end{minipage}
\begin{minipage}{0.195\textwidth}
    \centering{\small{ObjectMatch~\cite{gumeli2023objectmatch}}}
\end{minipage}
\begin{minipage}{0.195\textwidth}
    \centering{\small{SIFT~\cite{lowe1999sift}}}
\end{minipage}
\begin{minipage}{0.195\textwidth}
    \centering{\small{Ours}}
\end{minipage}

\hspace*{0.00mm}
\begin{minipage}{0.195\textwidth}
    \begin{overpic}[width=1\textwidth]{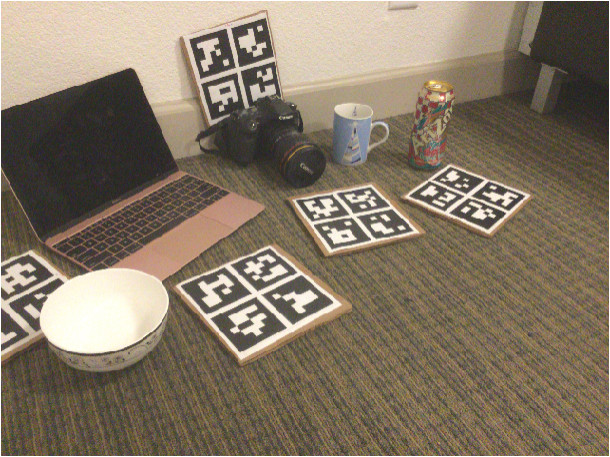}
    \end{overpic}
\end{minipage}
\begin{minipage}{0.195\textwidth}
    \begin{overpic}[width=1\textwidth]{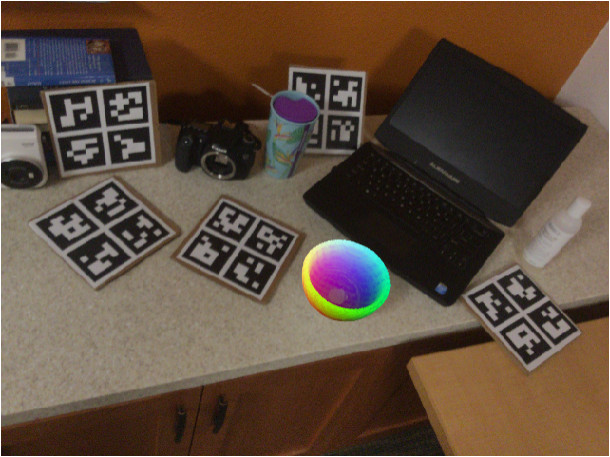}
    \end{overpic}
\end{minipage}
\begin{minipage}{0.195\textwidth}
    \begin{overpic}[width=1\textwidth]{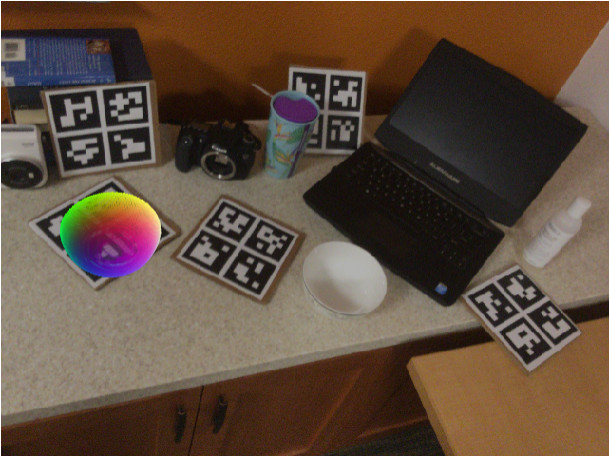}
    \end{overpic}
\end{minipage}
\begin{minipage}{0.195\textwidth}
    \begin{overpic}[width=1\textwidth]{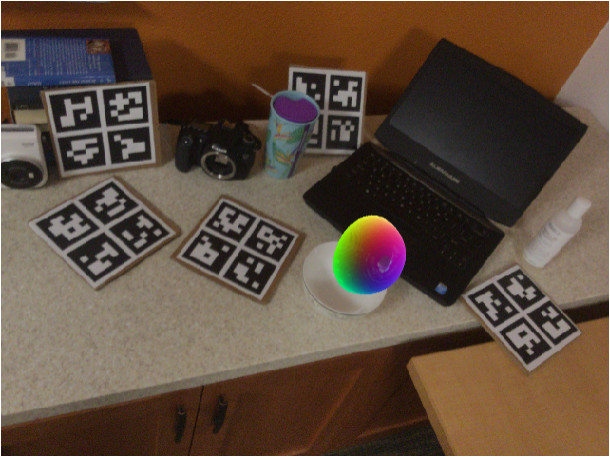}
    \end{overpic}
\end{minipage}
\begin{minipage}{0.195\textwidth}
    \begin{overpic}[width=1\textwidth]{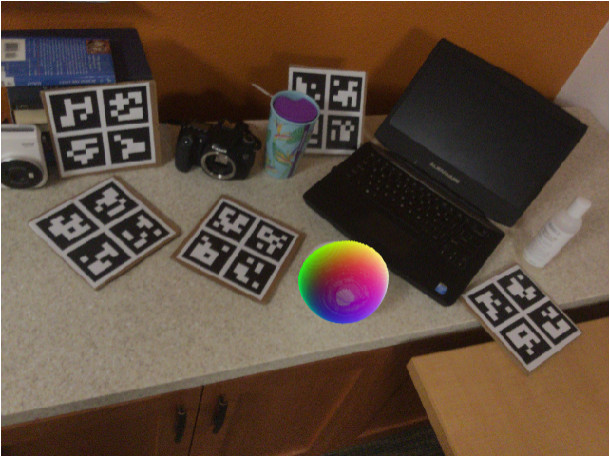}
    \end{overpic}
\end{minipage}
\vspace*{2mm}
\begin{minipage}{\textwidth}
    \centering{(a) Prompt: \texttt{White and blue bowl}}
\end{minipage}
\hspace*{0.00mm}
\begin{minipage}{0.195\textwidth}
    \begin{overpic}[width=1\textwidth]{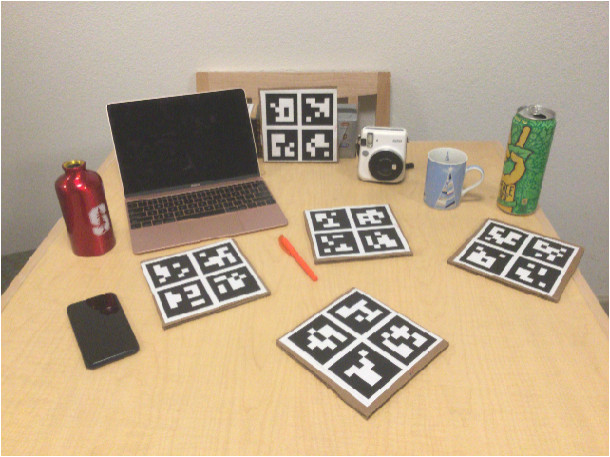}
    \end{overpic}
\end{minipage}
\begin{minipage}{0.195\textwidth}
    \begin{overpic}[width=1\textwidth]{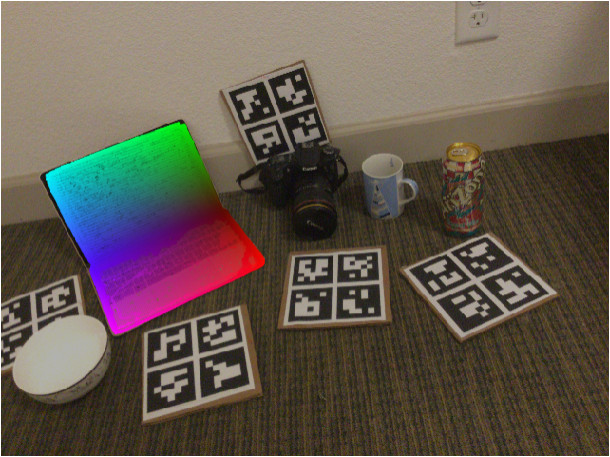}
    \end{overpic}
\end{minipage}
\begin{minipage}{0.195\textwidth}
    \begin{overpic}[width=1\textwidth]{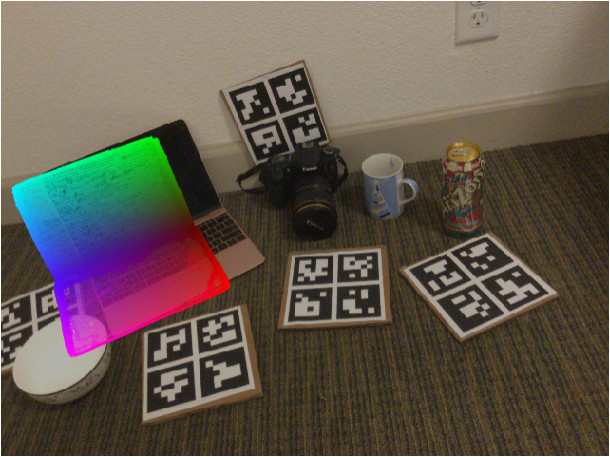}
    \end{overpic}
\end{minipage}
\begin{minipage}{0.195\textwidth}
    \begin{overpic}[width=1\textwidth]{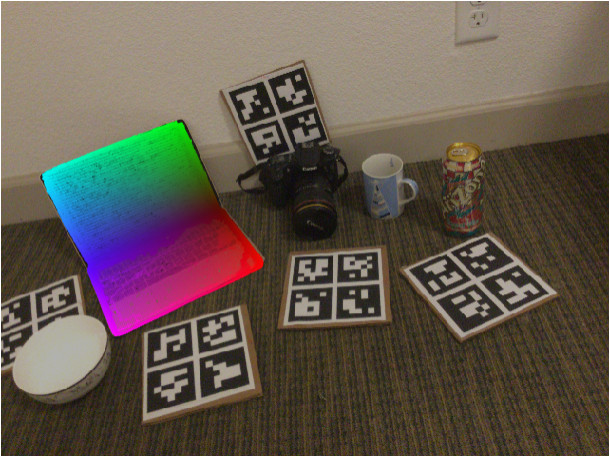}
    \end{overpic}
\end{minipage}
\begin{minipage}{0.195\textwidth}
    \begin{overpic}[width=1\textwidth]{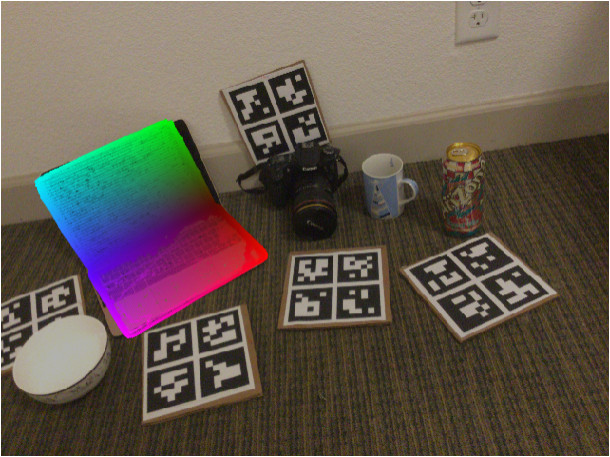}
    \end{overpic}
\end{minipage}
\vspace*{2mm}
\begin{minipage}{\textwidth}
    \centering{(b) Prompt: \texttt{Brown open laptop}}
\end{minipage}
\hspace*{0.00mm}
\begin{minipage}{0.195\textwidth}
    \begin{overpic}[width=1\textwidth]{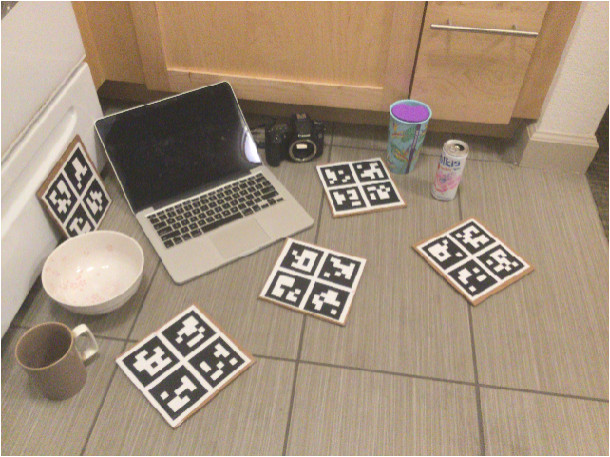}
    \end{overpic}
\end{minipage}
\begin{minipage}{0.195\textwidth}
    \begin{overpic}[width=1\textwidth]{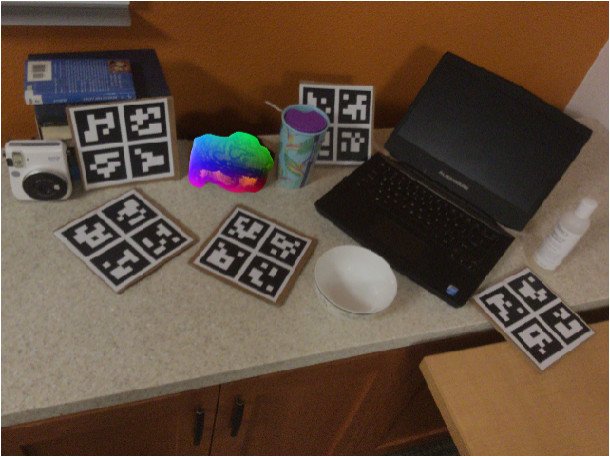}
    \end{overpic}
\end{minipage}
\begin{minipage}{0.195\textwidth}
    \begin{overpic}[width=1\textwidth]{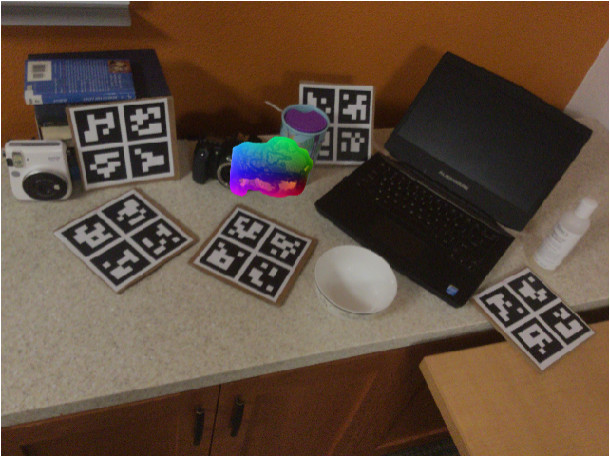}
    \end{overpic}
\end{minipage}
\begin{minipage}{0.195\textwidth}
    \begin{overpic}[width=1\textwidth]{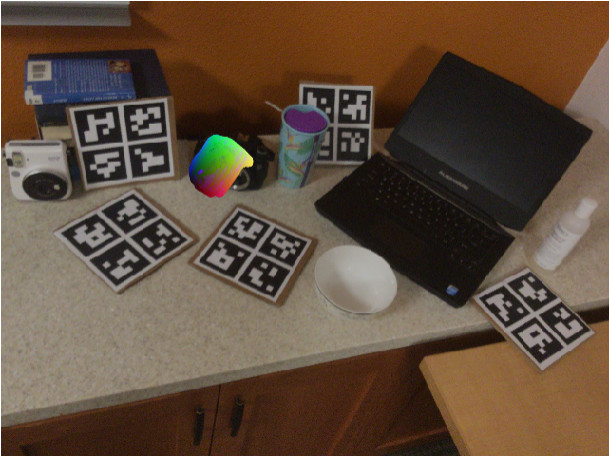}
    \end{overpic}
\end{minipage}
\begin{minipage}{0.195\textwidth}
    \begin{overpic}[width=1\textwidth]{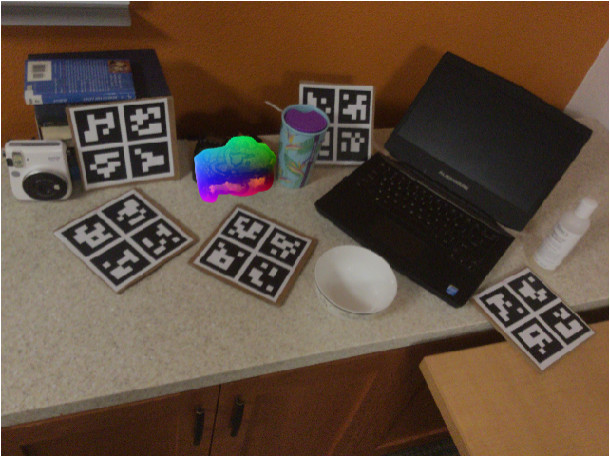}
    \end{overpic}
\end{minipage}
\vspace*{2mm}
\begin{minipage}{\textwidth}
    \centering{(c) Prompt: \texttt{Black lens camera}}
\end{minipage}
\hspace*{0.00mm}
\begin{minipage}{0.195\textwidth}
    \begin{overpic}[width=1\textwidth]{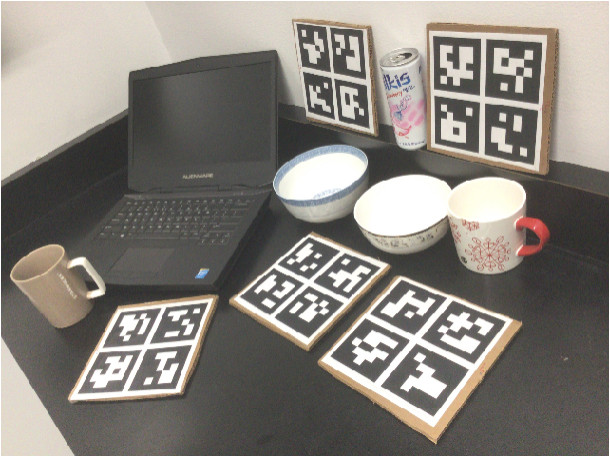}
    \end{overpic}
\end{minipage}
\begin{minipage}{0.195\textwidth}
    \begin{overpic}[width=1\textwidth]{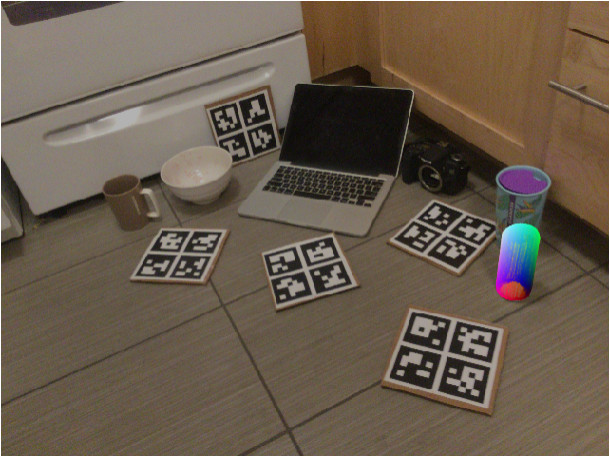}
    \end{overpic}
\end{minipage}
\begin{minipage}{0.195\textwidth}
    \begin{overpic}[width=1\textwidth]{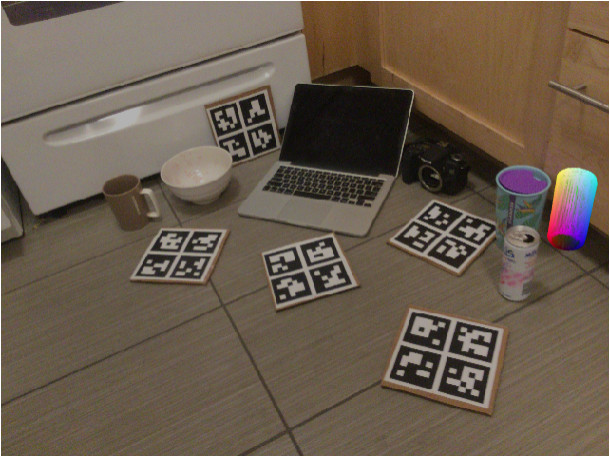}
    \end{overpic}
\end{minipage}
\begin{minipage}{0.195\textwidth}
    \begin{overpic}[width=1\textwidth]{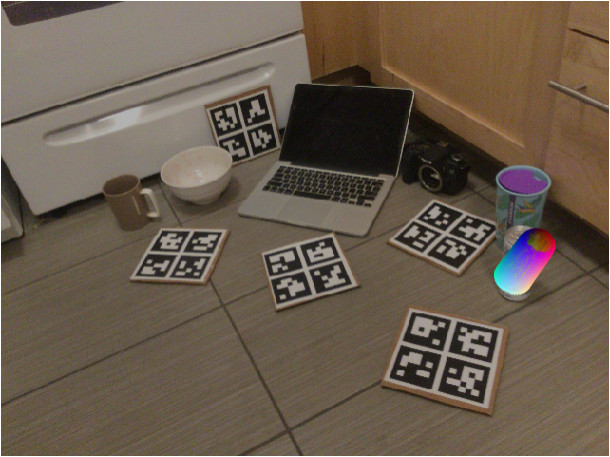}
    \end{overpic}
\end{minipage}
\begin{minipage}{0.195\textwidth}
    \begin{overpic}[width=1\textwidth]{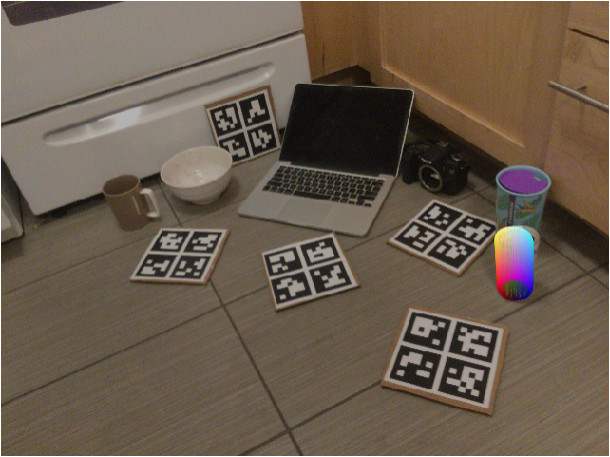}
    \end{overpic}
\end{minipage}
\vspace*{2mm}
\begin{minipage}{\textwidth}
    \centering{(d) Prompt: \texttt{White tall can}}
\end{minipage}

\hspace*{0.00mm}
\begin{minipage}{0.195\textwidth}
    \begin{overpic}[width=1\textwidth]{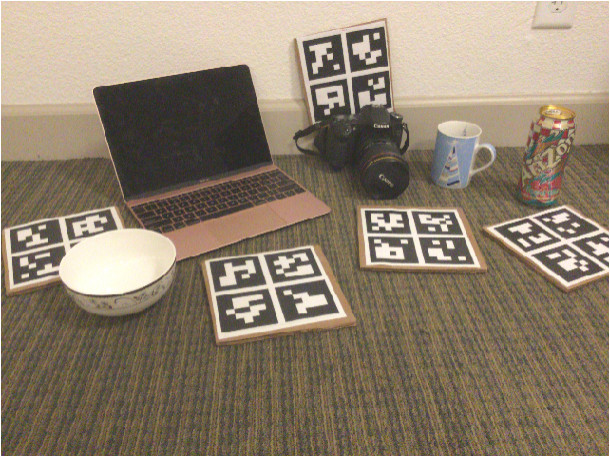}
    \end{overpic}
\end{minipage}
\begin{minipage}{0.195\textwidth}
    \begin{overpic}[width=1\textwidth]{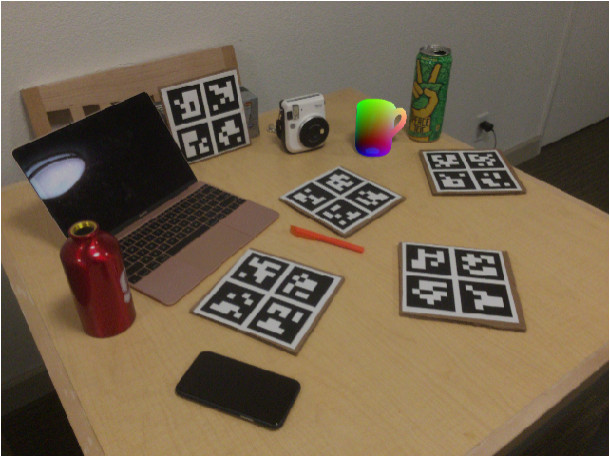}
    \end{overpic}
\end{minipage}
\begin{minipage}{0.195\textwidth}
    \begin{overpic}[width=1\textwidth]{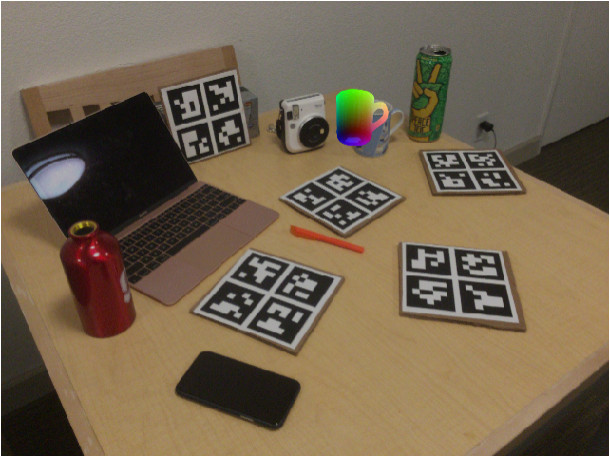}
    \end{overpic}
\end{minipage}
\begin{minipage}{0.195\textwidth}
    \begin{overpic}[width=1\textwidth]{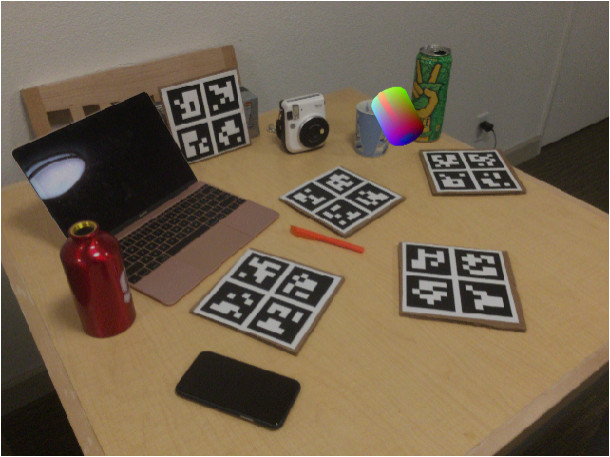}
    \end{overpic}
\end{minipage}
\begin{minipage}{0.195\textwidth}
    \begin{overpic}[width=1\textwidth]{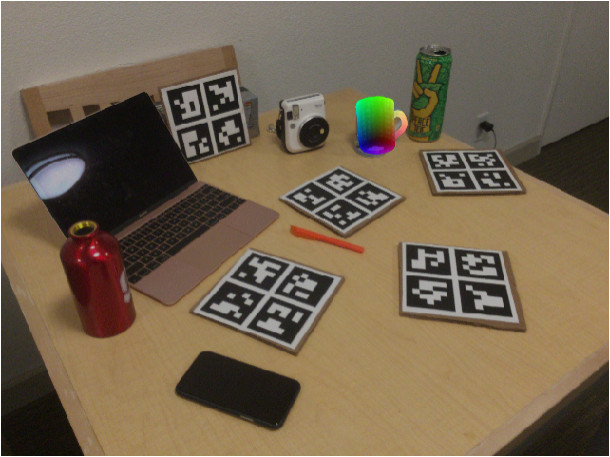}
    \end{overpic}
\end{minipage}
\vspace*{2mm}
\begin{minipage}{\textwidth}
    \centering{(e) Prompt: \texttt{Light blue mug}}
\end{minipage}

\hspace*{0.00mm}
\begin{minipage}{0.195\textwidth}
    \begin{overpic}[width=1\textwidth]{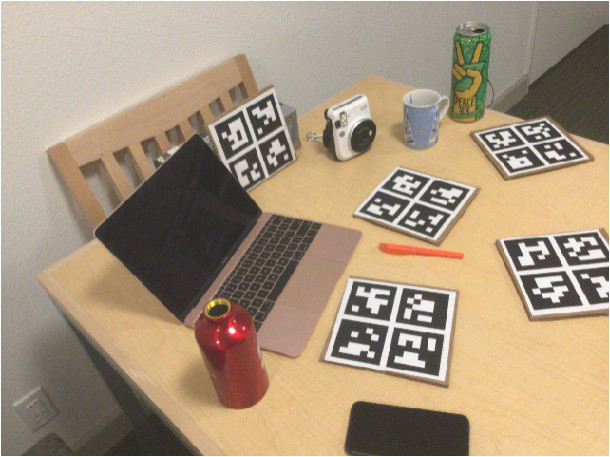}
    \end{overpic}
\end{minipage}
\begin{minipage}{0.195\textwidth}
    \begin{overpic}[width=1\textwidth]{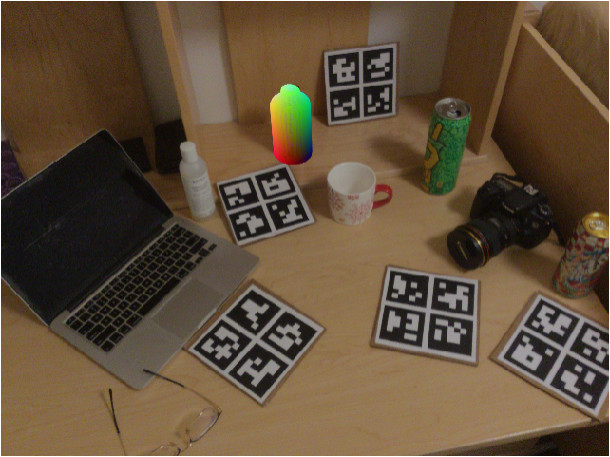}
    \end{overpic}
\end{minipage}
\begin{minipage}{0.195\textwidth}
    \begin{overpic}[width=1\textwidth]{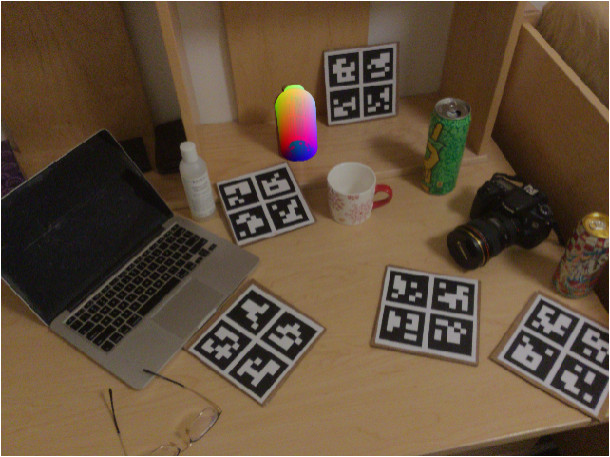}
    \end{overpic}
\end{minipage}
\begin{minipage}{0.195\textwidth}
    \begin{overpic}[width=1\textwidth]{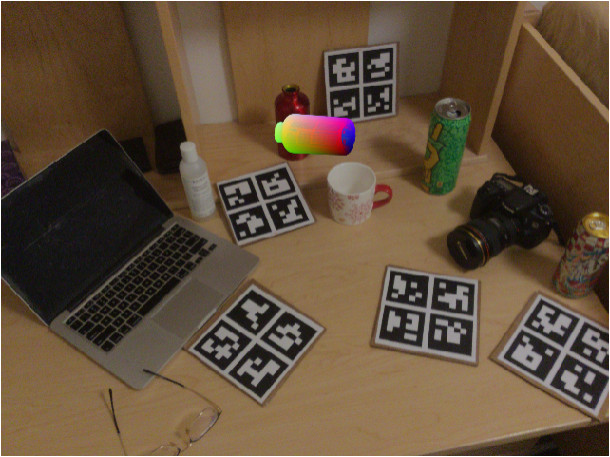}
    \end{overpic}
\end{minipage}
\begin{minipage}{0.195\textwidth}
    \begin{overpic}[width=1\textwidth]{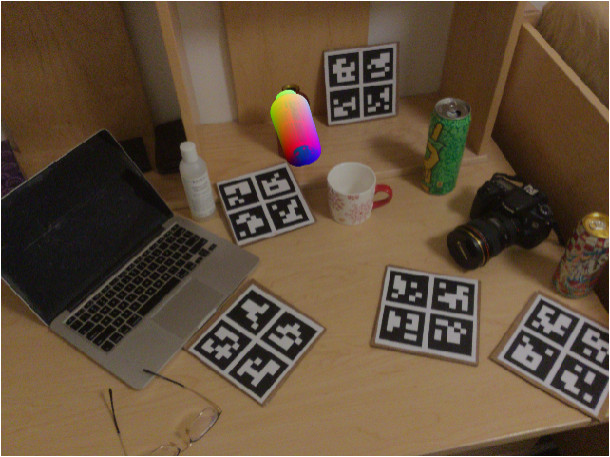}
    \end{overpic}
\end{minipage}
\vspace*{2mm}
\begin{minipage}{\textwidth}
    \centering{(f) Prompt: \texttt{Red Stanford bottle}}
\end{minipage}

%% file: supp/figures/qualitative/pose_toyl/pose.tex
\hspace*{0.00mm}
\begin{minipage}{0.195\textwidth}
    \centering{\small{Anchor}}
\end{minipage}
\begin{minipage}{0.195\textwidth}
    \centering{\small{Ground truth}}
\end{minipage}
\begin{minipage}{0.195\textwidth}
    \centering{\small{ObjectMatch~\cite{gumeli2023objectmatch}}}
\end{minipage}
\begin{minipage}{0.195\textwidth}
    \centering{\small{SIFT~\cite{lowe1999sift}}}
\end{minipage}
\begin{minipage}{0.195\textwidth}
    \centering{\small{Ours}}
\end{minipage}

\hspace*{0.00mm}
\begin{minipage}{0.195\textwidth}
    \begin{overpic}[width=1\textwidth]{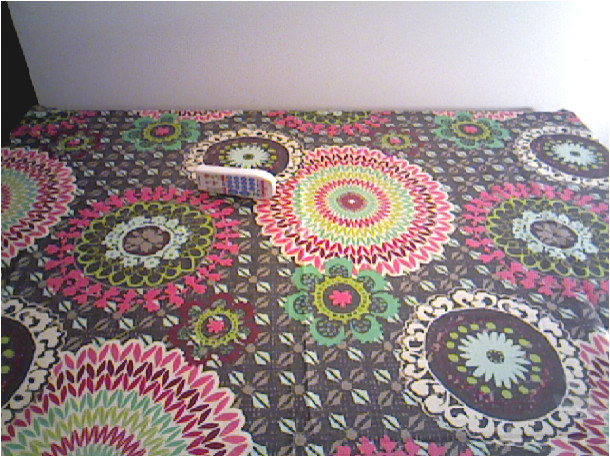}
    \end{overpic}
\end{minipage}
\begin{minipage}{0.195\textwidth}
    \begin{overpic}[width=1\textwidth]{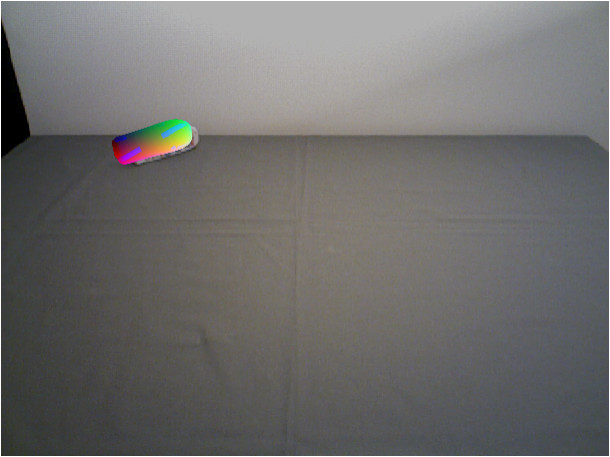}
    \end{overpic}
\end{minipage}
\begin{minipage}{0.195\textwidth}
    \begin{overpic}[width=1\textwidth]{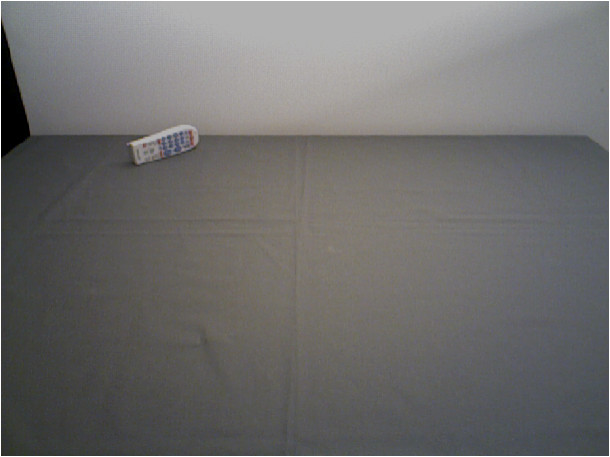}
    \end{overpic}
\end{minipage}
\begin{minipage}{0.195\textwidth}
    \begin{overpic}[width=1\textwidth]{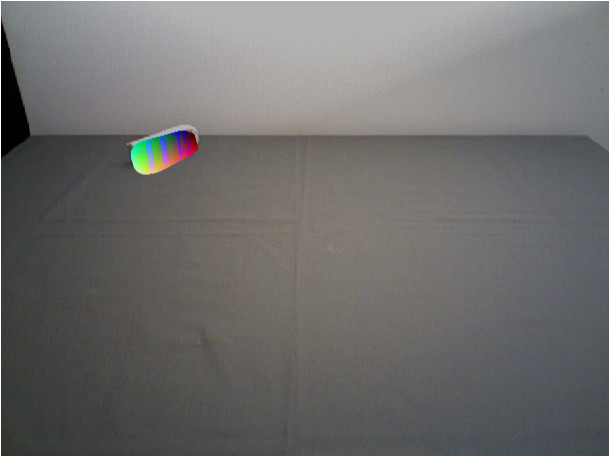}
    \end{overpic}
\end{minipage}
\begin{minipage}{0.195\textwidth}
    \begin{overpic}[width=1\textwidth]{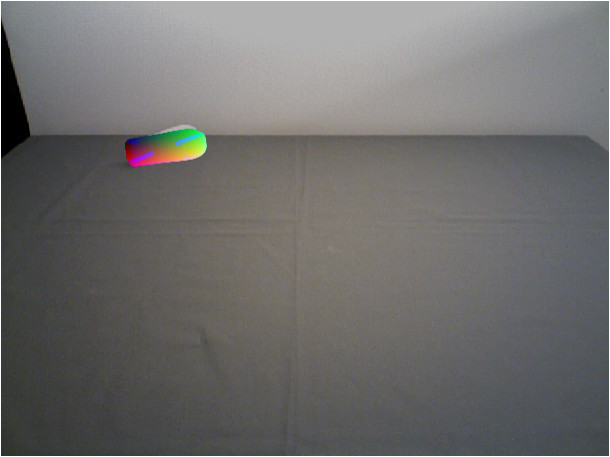}
    \end{overpic} 
\end{minipage}
\vspace*{2mm}
\begin{minipage}{\textwidth}
    \centering{(a) Prompt: \texttt{White and blue remote}}
\end{minipage}
\hspace*{0.00mm}
\begin{minipage}{0.195\textwidth}
    \begin{overpic}[width=1\textwidth]{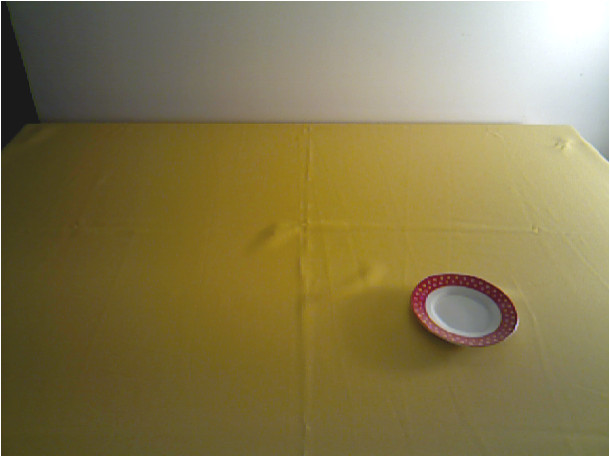}
    \end{overpic}
\end{minipage}
\begin{minipage}{0.195\textwidth}
    \begin{overpic}[width=1\textwidth]{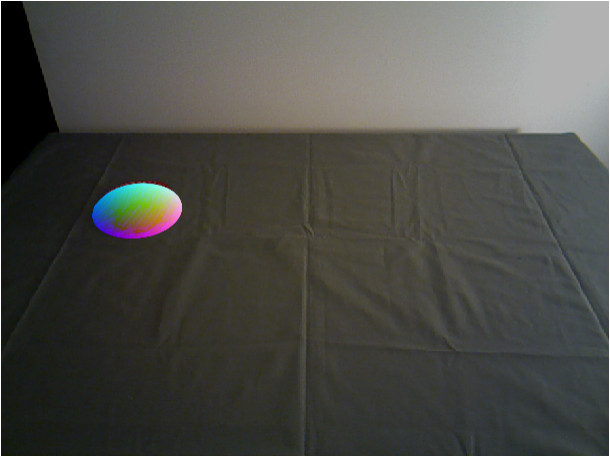}
    \end{overpic}
\end{minipage}
\begin{minipage}{0.195\textwidth}
    \begin{overpic}[width=1\textwidth]{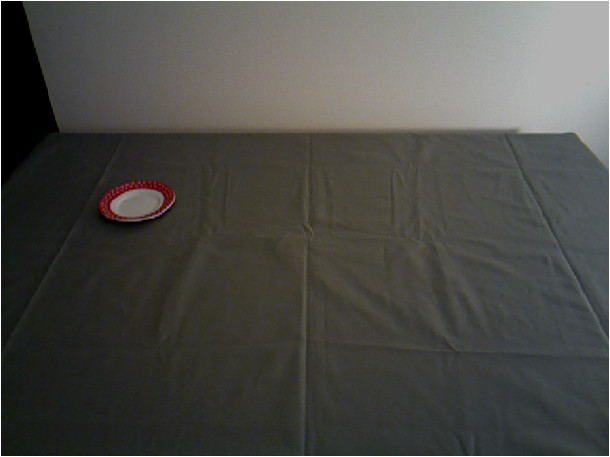}
    \end{overpic}
\end{minipage}
\begin{minipage}{0.195\textwidth}
    \begin{overpic}[width=1\textwidth]{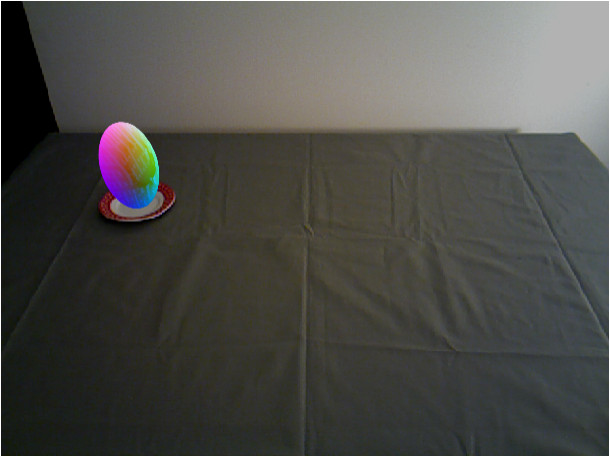}
    \end{overpic}
\end{minipage}
\begin{minipage}{0.195\textwidth}
    \begin{overpic}[width=1\textwidth]{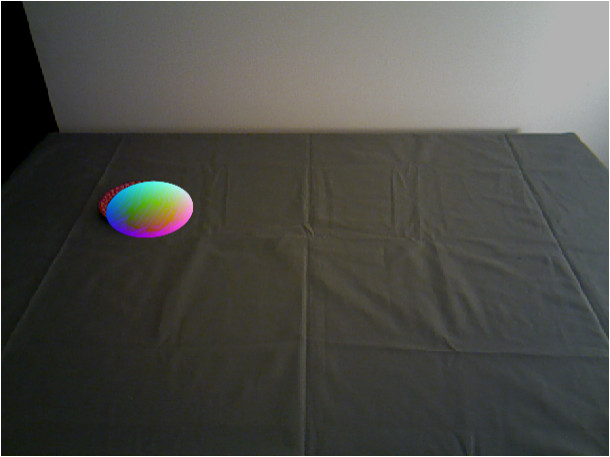}
    \end{overpic}
\end{minipage}
\vspace*{2mm}
\begin{minipage}{\textwidth}
    \centering{(b) Prompt: \texttt{Red and white small plate}}
\end{minipage}
\hspace*{0.00mm}
\begin{minipage}{0.195\textwidth}
    \begin{overpic}[width=1\textwidth]{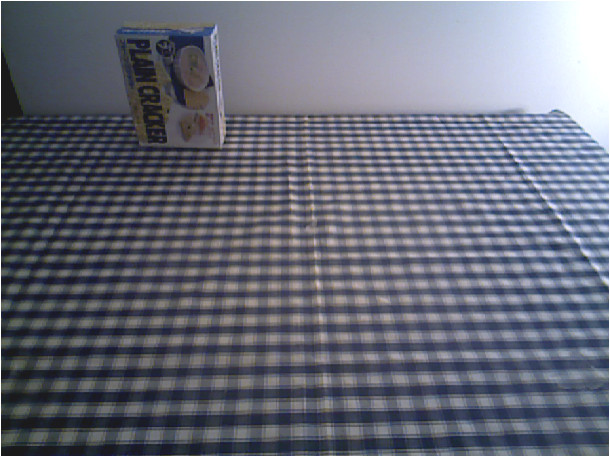}
    \end{overpic}
\end{minipage}
\begin{minipage}{0.195\textwidth}
    \begin{overpic}[width=1\textwidth]{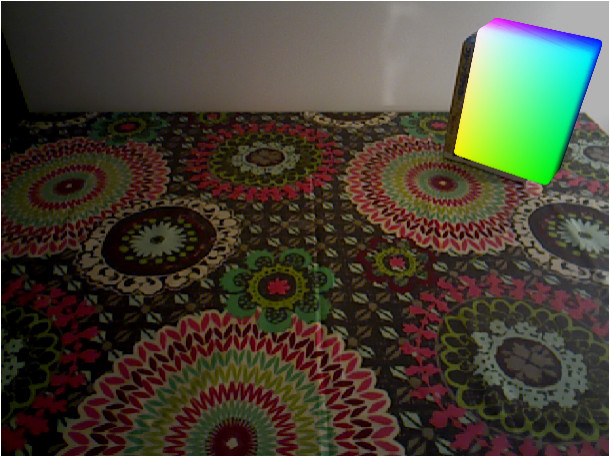}
    \end{overpic}
\end{minipage}
\begin{minipage}{0.195\textwidth}
    \begin{overpic}[width=1\textwidth]{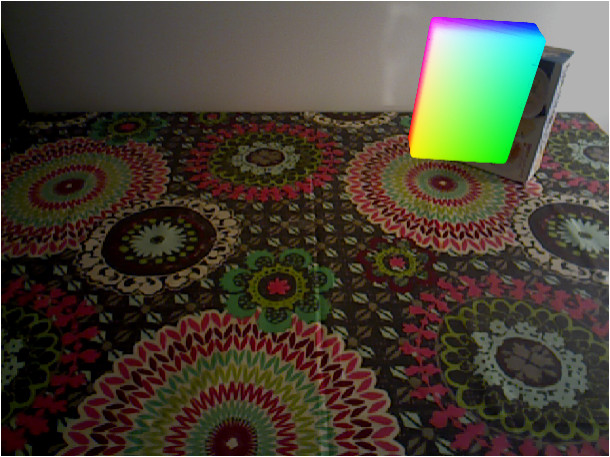}
    \end{overpic}
\end{minipage}
\begin{minipage}{0.195\textwidth}
    \begin{overpic}[width=1\textwidth]{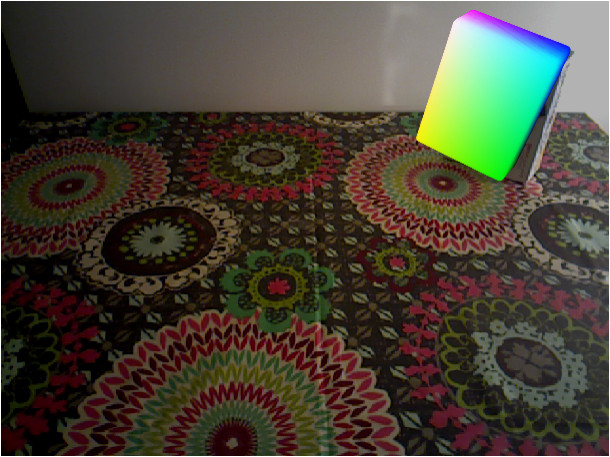}
    \end{overpic}
\end{minipage}
\begin{minipage}{0.195\textwidth}
    \begin{overpic}[width=1\textwidth]{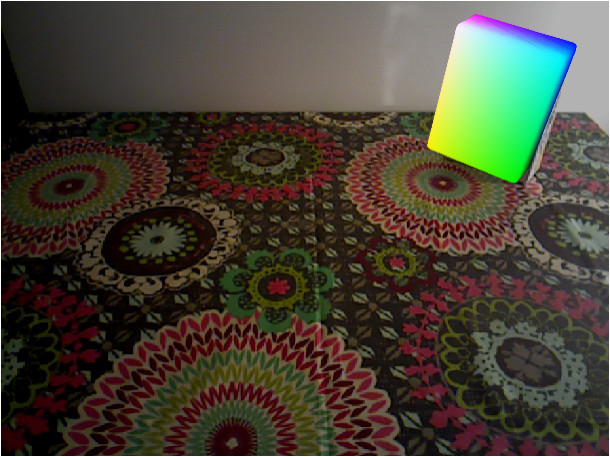}
    \end{overpic}
\end{minipage}
\vspace*{2mm}
\begin{minipage}{\textwidth}
    \centering{(c) Prompt: \texttt{Blue and yellow cracker box}}
\end{minipage}
\hspace*{0.00mm}
\begin{minipage}{0.195\textwidth}
    \begin{overpic}[width=1\textwidth]{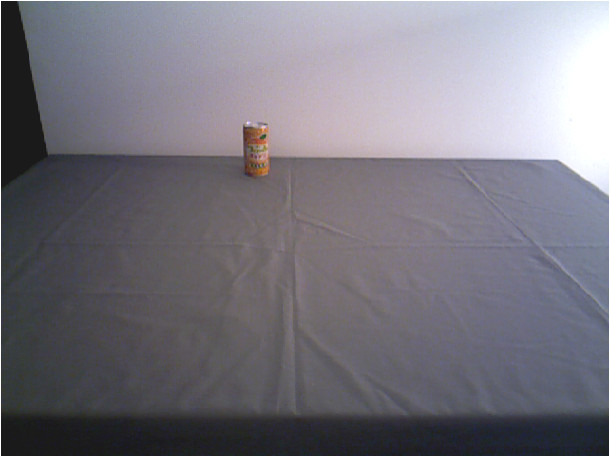}
    \end{overpic}
\end{minipage}
\begin{minipage}{0.195\textwidth}
    \begin{overpic}[width=1\textwidth]{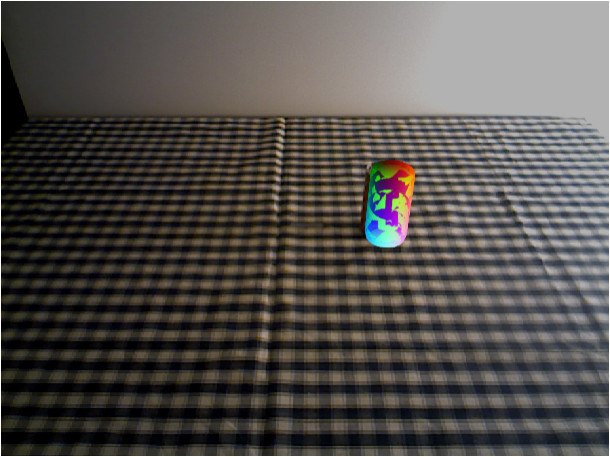}
    \end{overpic}
\end{minipage}
\begin{minipage}{0.195\textwidth}
    \begin{overpic}[width=1\textwidth]{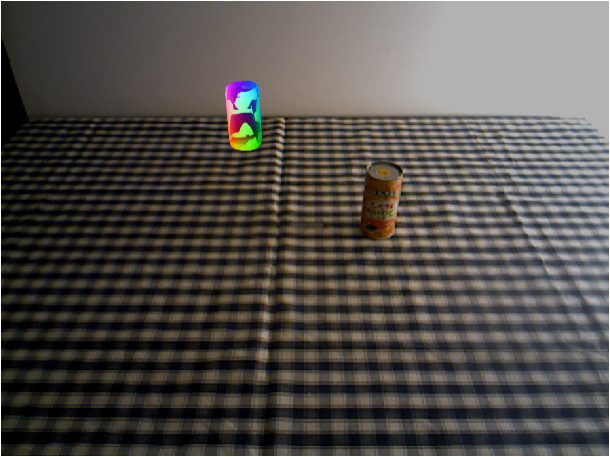}
    \end{overpic}
\end{minipage}
\begin{minipage}{0.195\textwidth}
    \begin{overpic}[width=1\textwidth]{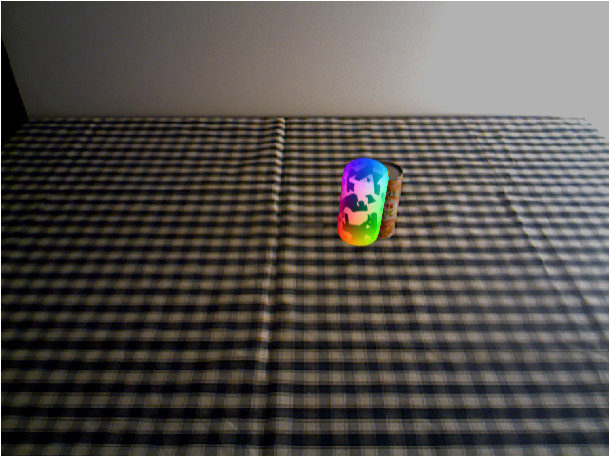}
    \end{overpic}
\end{minipage}
\begin{minipage}{0.195\textwidth}
    \begin{overpic}[width=1\textwidth]{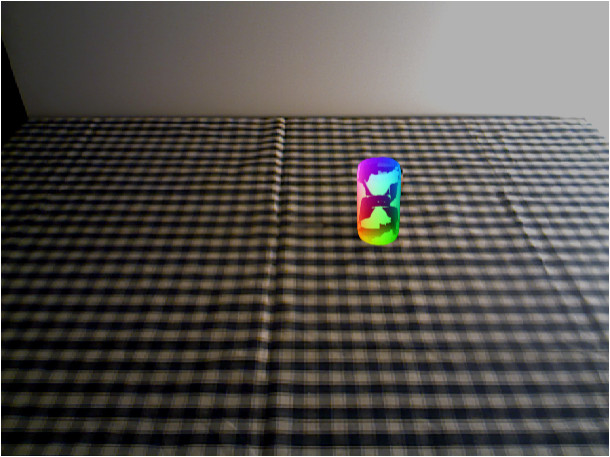}
    \end{overpic}
\end{minipage}
\vspace*{2mm}
\begin{minipage}{\textwidth}
    \centering{(d) Prompt: \texttt{Orange can}}
\end{minipage}

\hspace*{0.00mm}
\begin{minipage}{0.195\textwidth}
    \begin{overpic}[width=1\textwidth]{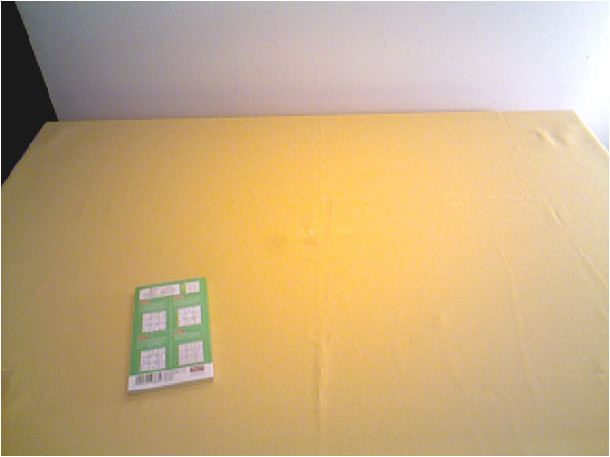}
    \end{overpic}
\end{minipage}
\begin{minipage}{0.195\textwidth}
    \begin{overpic}[width=1\textwidth]{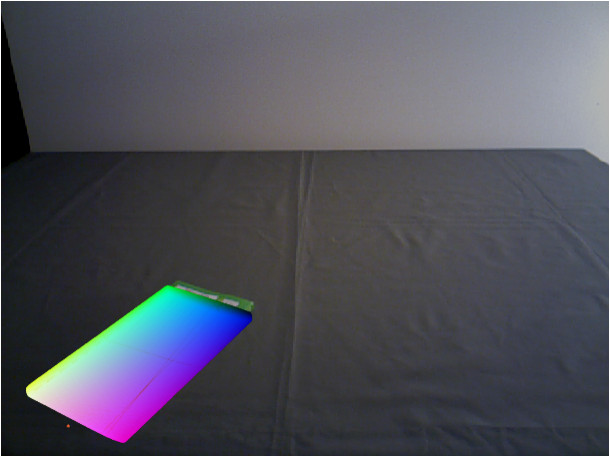}
    \end{overpic}
\end{minipage}
\begin{minipage}{0.195\textwidth}
    \begin{overpic}[width=1\textwidth]{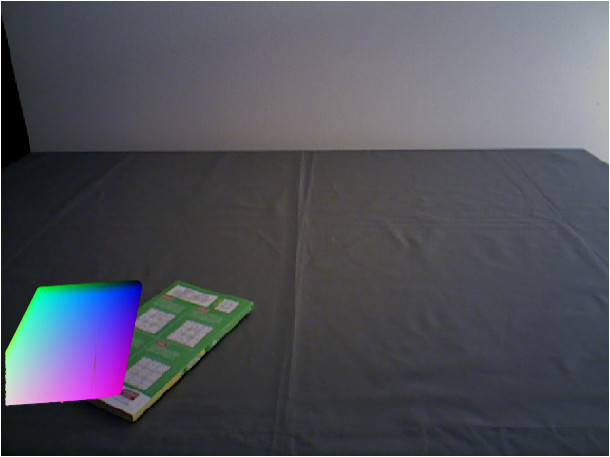}
    \end{overpic}
\end{minipage}
\begin{minipage}{0.195\textwidth}
    \begin{overpic}[width=1\textwidth]{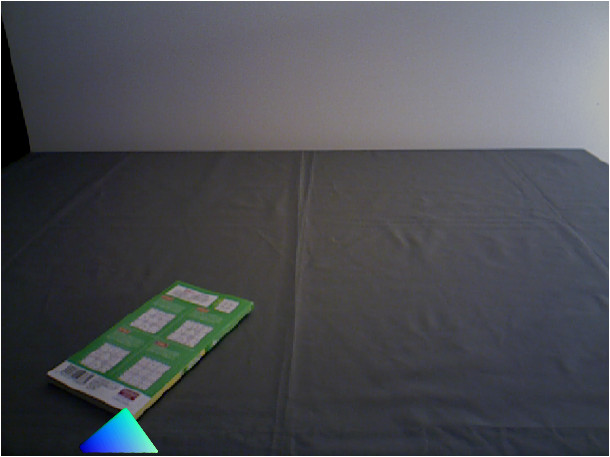}
    \end{overpic}
\end{minipage}
\begin{minipage}{0.195\textwidth}
    \begin{overpic}[width=1\textwidth]{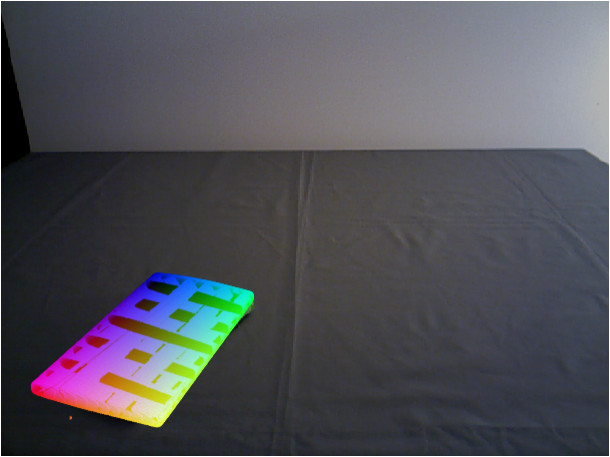}
    \end{overpic}
\end{minipage}
\vspace*{2mm}
\begin{minipage}{\textwidth}
    \centering{(e) Prompt: \texttt{Green sudoku magazine}}
\end{minipage}
\hspace*{0.00mm}
\begin{minipage}{0.195\textwidth}
    \begin{overpic}[width=1\textwidth]{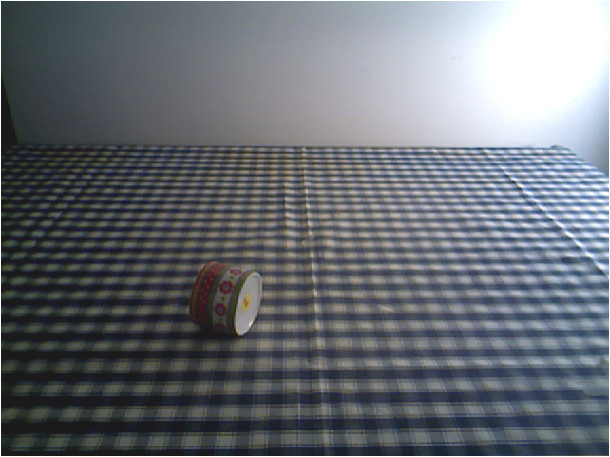}
    \end{overpic}
\end{minipage}
\begin{minipage}{0.195\textwidth}
    \begin{overpic}[width=1\textwidth]{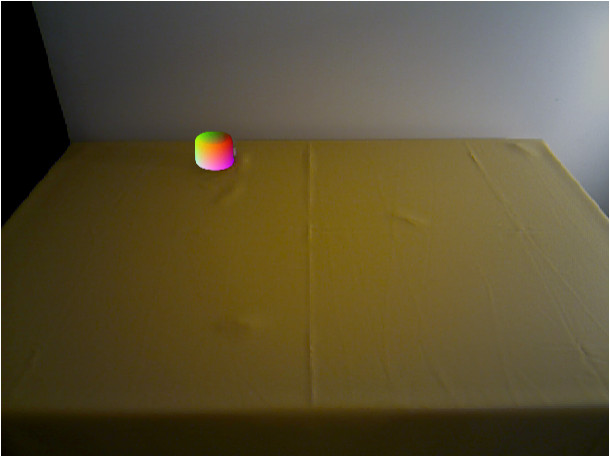}
    \end{overpic}
\end{minipage}
\begin{minipage}{0.195\textwidth}
    \begin{overpic}[width=1\textwidth]{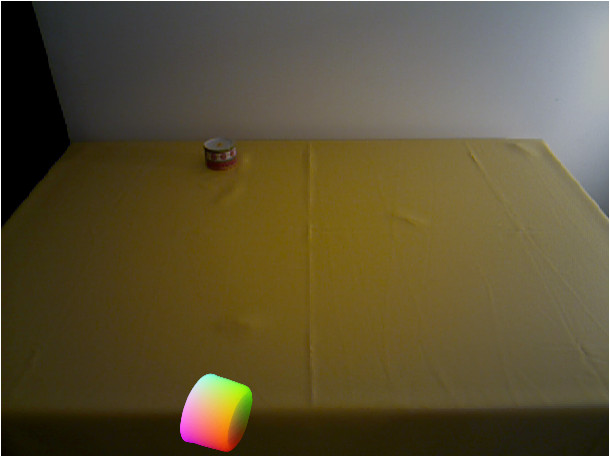}
    \end{overpic}
\end{minipage}
\begin{minipage}{0.195\textwidth}
    \begin{overpic}[width=1\textwidth]{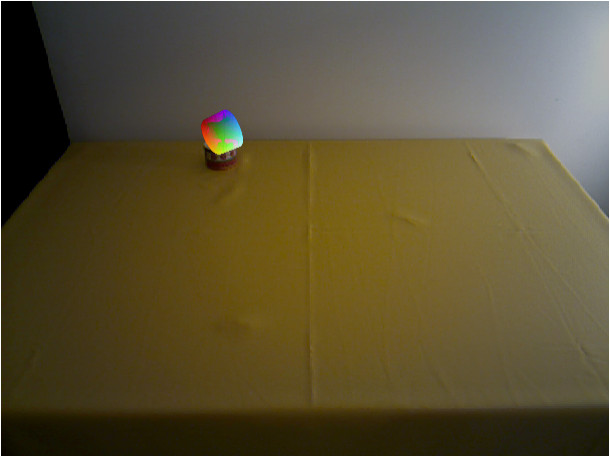}
    \end{overpic}
\end{minipage}
\begin{minipage}{0.195\textwidth}
    \begin{overpic}[width=1\textwidth]{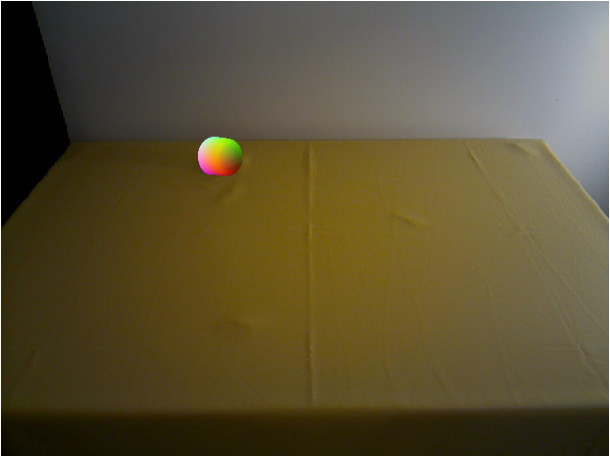}
    \end{overpic}
\end{minipage}
\vspace*{2mm}
\begin{minipage}{\textwidth}
    \centering{(f) Prompt: \texttt{Red and white small mug}}
\end{minipage}

%% file: supp/figures/qualitative/segm_nocs/segm.tex
\hspace*{0.3mm}
\vspace{1mm}
\begin{minipage}{0.155\textwidth}
    \centering{\small{Anchor}}
\end{minipage}
\begin{minipage}{0.155\textwidth}
    \centering{\small{Query}}
\end{minipage}
\begin{minipage}{0.155\textwidth}
    \centering{\small{Anchor}}
\end{minipage}
\begin{minipage}{0.155\textwidth}
    \centering{\small{Query}}
\end{minipage}
\begin{minipage}{0.155\textwidth}
    \centering{\small{Anchor}}
\end{minipage}
\begin{minipage}{0.155\textwidth}
    \centering{\small{Query}}
\end{minipage}
\hspace*{0.3mm}
\begin{minipage}{0.155\textwidth}
    \begin{overpic}[width=1\textwidth]{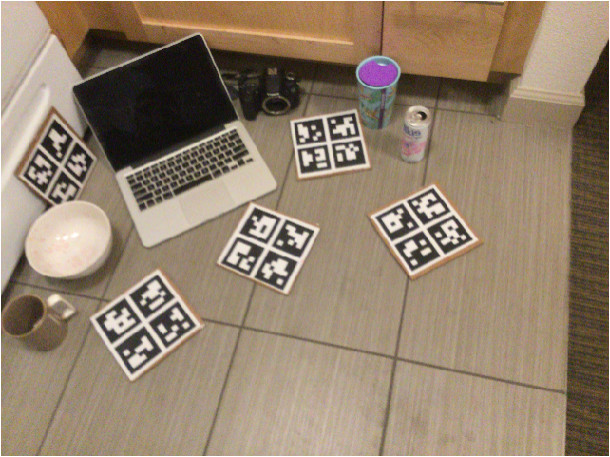}
    \end{overpic}
\end{minipage}
\begin{minipage}{0.155\textwidth}
    \begin{overpic}[width=1\textwidth]{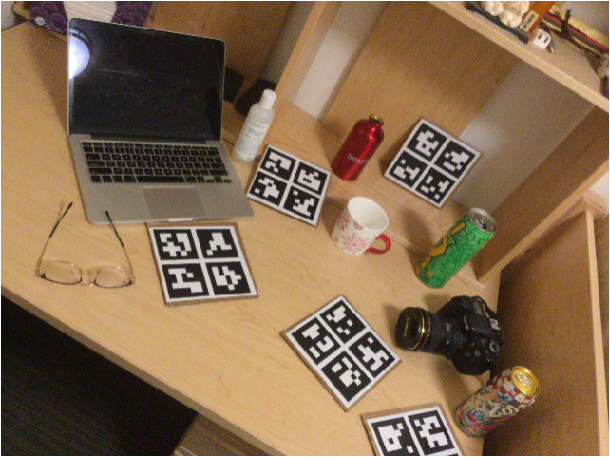}
    \end{overpic}
\end{minipage}
\hspace*{1mm}
\begin{minipage}{0.155\textwidth}
    \begin{overpic}[width=1\textwidth]{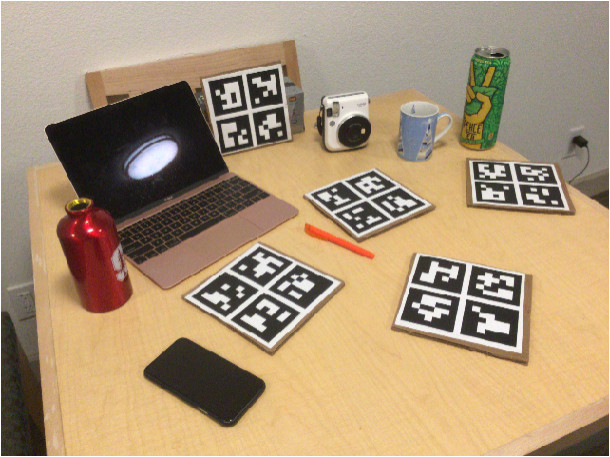}
    \end{overpic}
\end{minipage}
\begin{minipage}{0.155\textwidth}
    \begin{overpic}[width=1\textwidth]{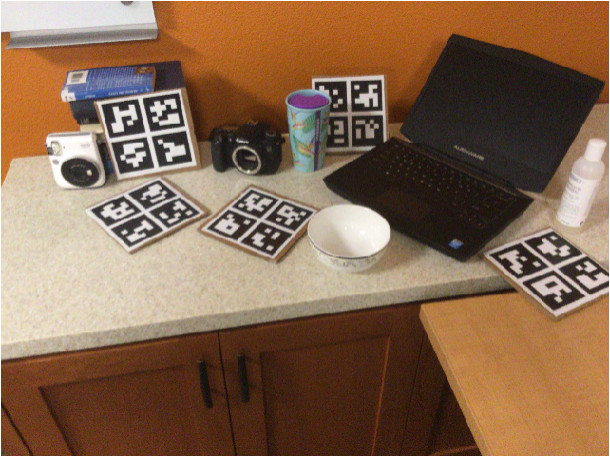}
    \end{overpic}
\end{minipage}
\hspace*{1mm}
\begin{minipage}{0.155\textwidth}
    \begin{overpic}[width=1\textwidth]{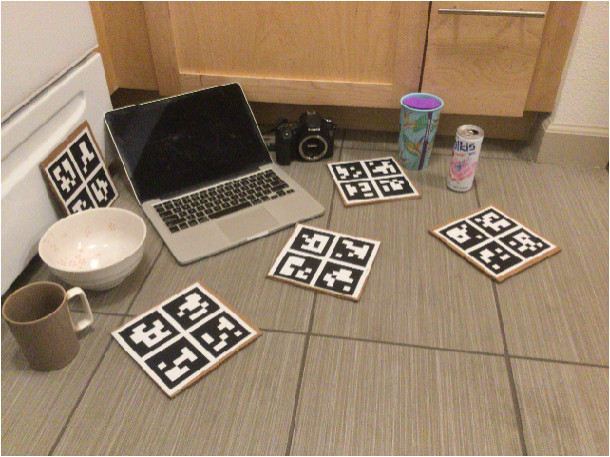}
    \end{overpic}
\end{minipage}
\begin{minipage}{0.155\textwidth}
    \begin{overpic}[width=1\textwidth]{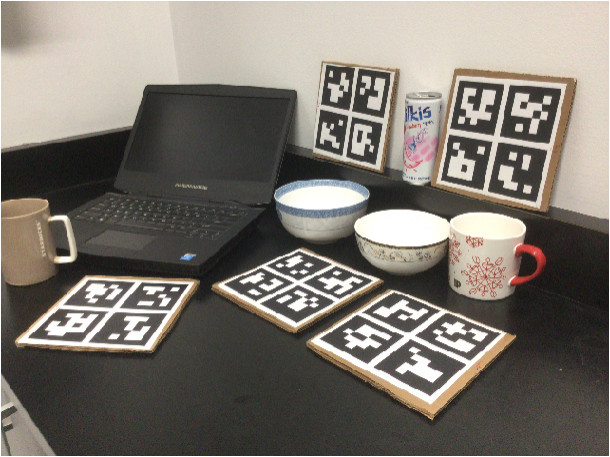}
    \end{overpic}
\end{minipage}
\vspace*{1mm}

\hspace*{0.3mm}
\begin{minipage}{0.155\textwidth}
    \begin{overpic}[width=1\textwidth]{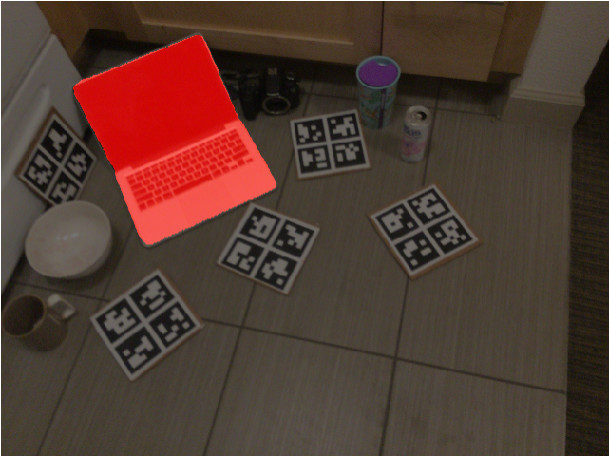}
    \put(-12,8){\small{\rotatebox{90}{Ground truth}}}
    \end{overpic}
\end{minipage}
\begin{minipage}{0.155\textwidth}
    \begin{overpic}[width=1\textwidth]{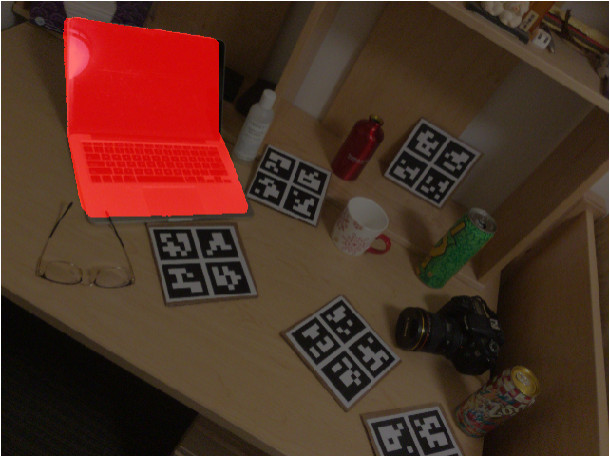}
    \end{overpic}
\end{minipage}
\hspace*{1mm}
\begin{minipage}{0.155\textwidth}
    \begin{overpic}[width=1\textwidth]{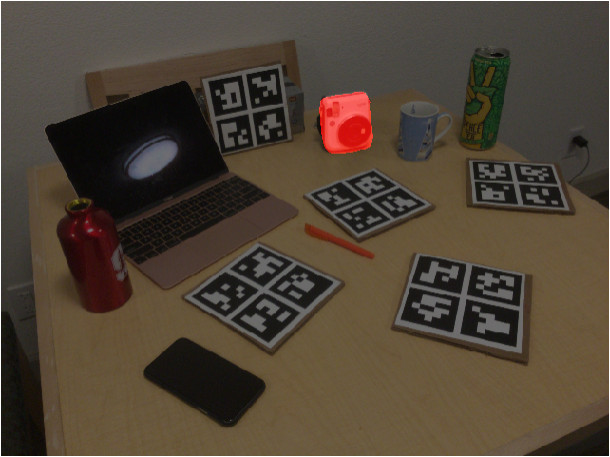}
    \end{overpic}
\end{minipage}
\begin{minipage}{0.155\textwidth}
    \begin{overpic}[width=1\textwidth]{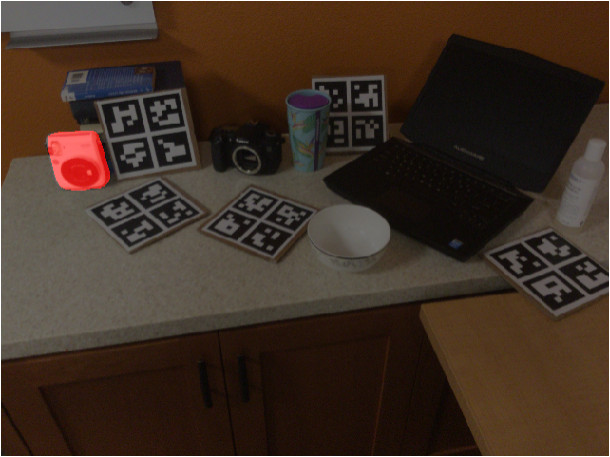}
    \end{overpic}
\end{minipage}
\hspace*{1mm}
\begin{minipage}{0.155\textwidth}
    \begin{overpic}[width=1\textwidth]{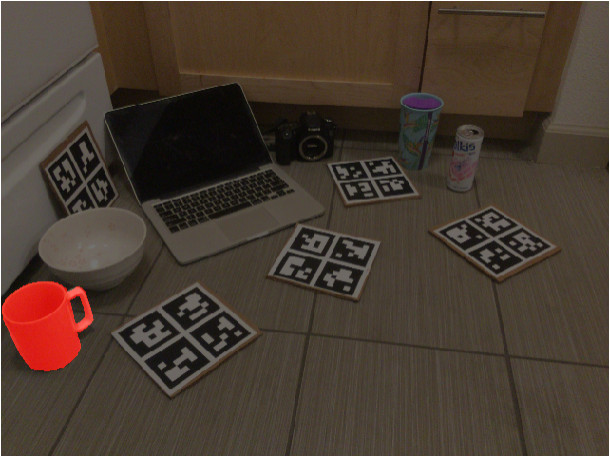}
    \end{overpic}
\end{minipage}
\begin{minipage}{0.155\textwidth}
    \begin{overpic}[width=1\textwidth]{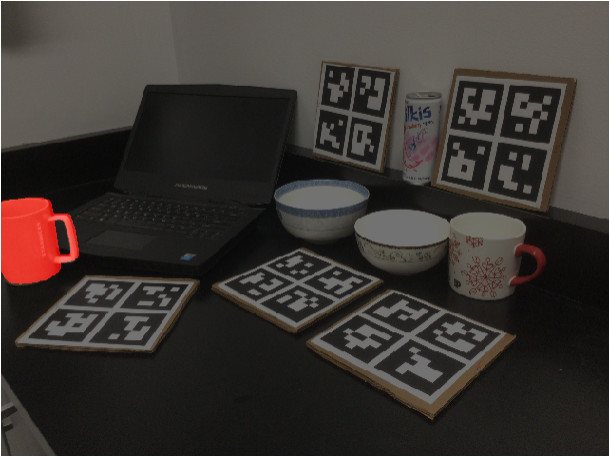}
    \end{overpic}
\end{minipage}
\vspace*{1mm}

\hspace*{0.3mm}
\begin{minipage}{0.155\textwidth}
    \begin{overpic}[width=1\textwidth]{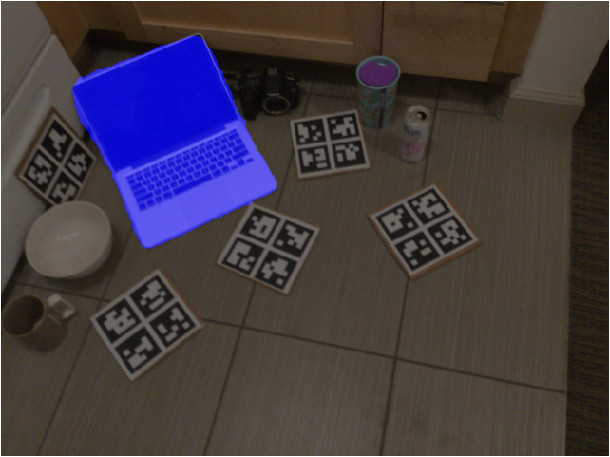}
    \put(-12,10){\small{\rotatebox{90}{OVSeg~\cite{liang2023ovseg}}}}
    \end{overpic}
\end{minipage}
\begin{minipage}{0.155\textwidth}
    \begin{overpic}[width=1\textwidth]{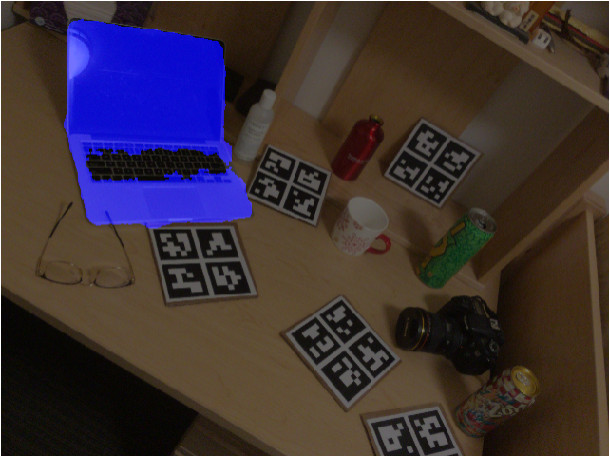}
    \end{overpic}
\end{minipage}
\hspace*{1mm}
\begin{minipage}{0.155\textwidth}
    \begin{overpic}[width=1\textwidth]{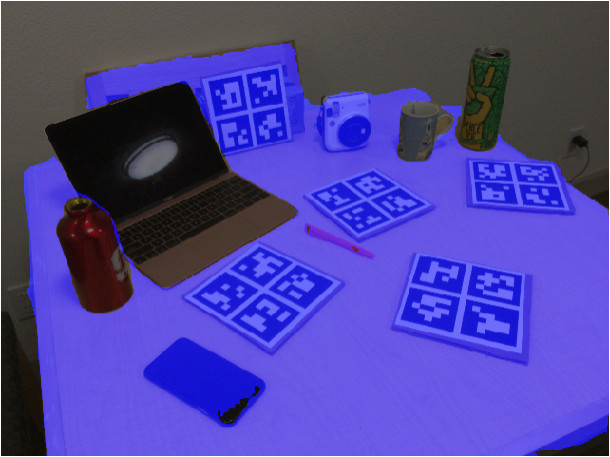}
    \end{overpic}
\end{minipage}
\begin{minipage}{0.155\textwidth}
    \begin{overpic}[width=1\textwidth]{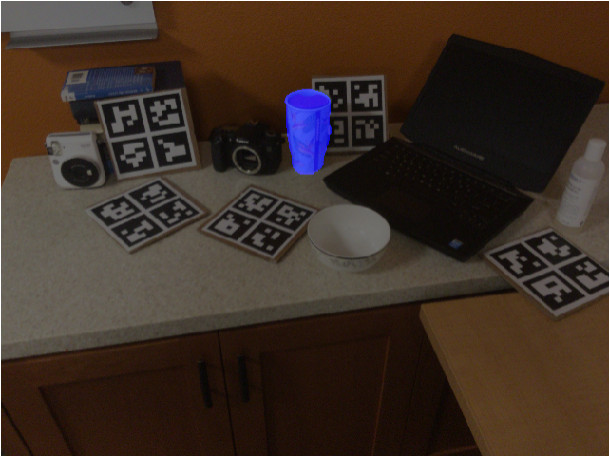}
    \end{overpic}
\end{minipage}
\hspace*{1mm}
\begin{minipage}{0.155\textwidth}
    \begin{overpic}[width=1\textwidth]{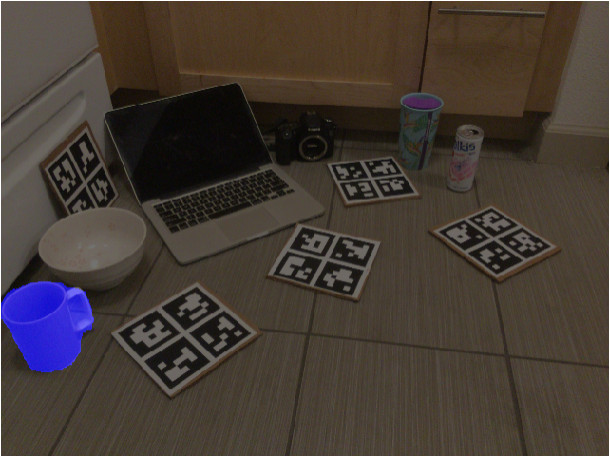}
    \end{overpic}
\end{minipage}
\begin{minipage}{0.155\textwidth}
    \begin{overpic}[width=1\textwidth]{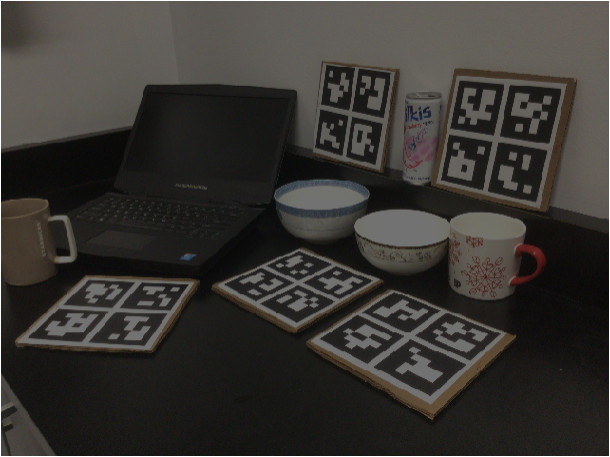}
    \end{overpic}
\end{minipage}
\vspace*{1mm}

\begin{minipage}{0.155\textwidth}
    \begin{overpic}[width=1\textwidth]{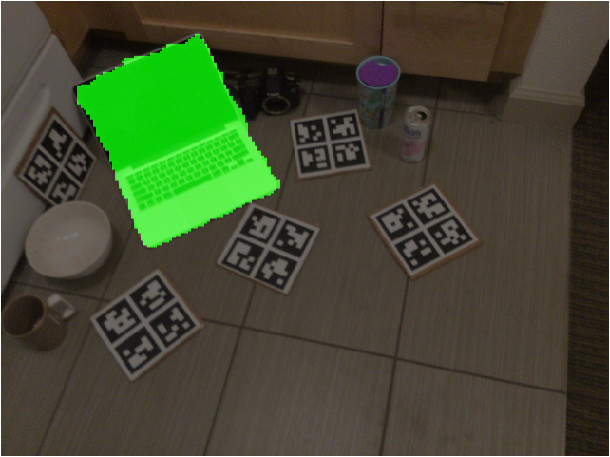}
    \put(-12,18){\small{\rotatebox{90}{\acronym}}}
    \end{overpic}
\end{minipage}
\begin{minipage}{0.155\textwidth}
    \begin{overpic}[width=1\textwidth]{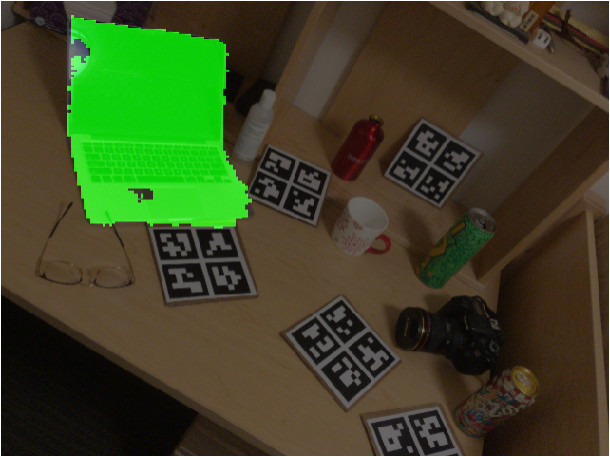}
    \end{overpic}
\end{minipage}
\hspace*{1mm}
\begin{minipage}{0.155\textwidth}
    \begin{overpic}[width=1\textwidth]{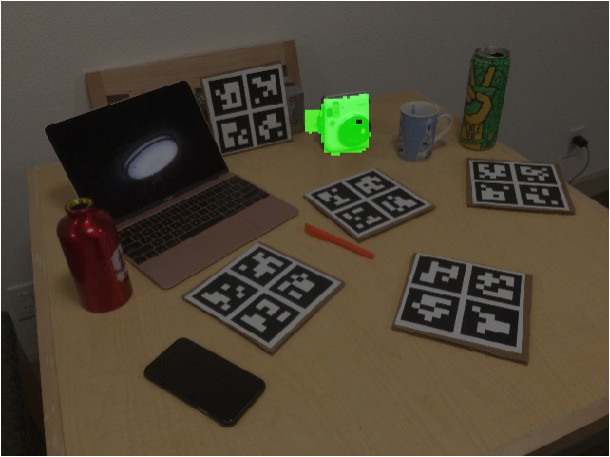}
    \end{overpic}
\end{minipage}
\begin{minipage}{0.155\textwidth}
    \begin{overpic}[width=1\textwidth]{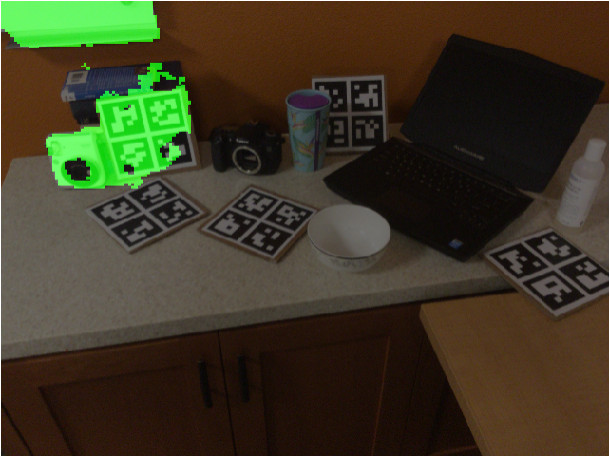}
    \end{overpic}
\end{minipage}
\hspace*{1mm}
\begin{minipage}{0.155\textwidth}
    \begin{overpic}[width=1\textwidth]{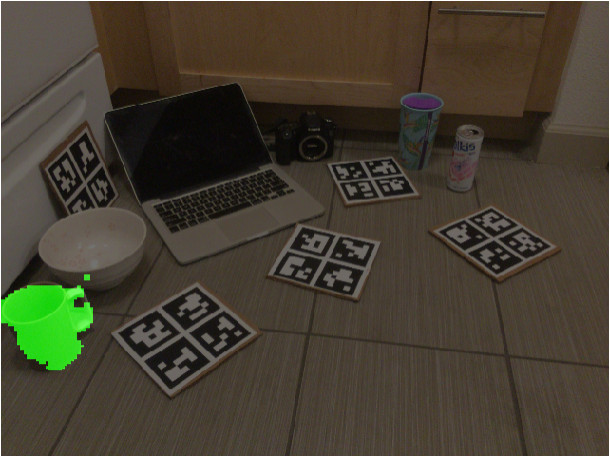}
    \end{overpic}
\end{minipage}
\begin{minipage}{0.155\textwidth}
    \begin{overpic}[width=1\textwidth]{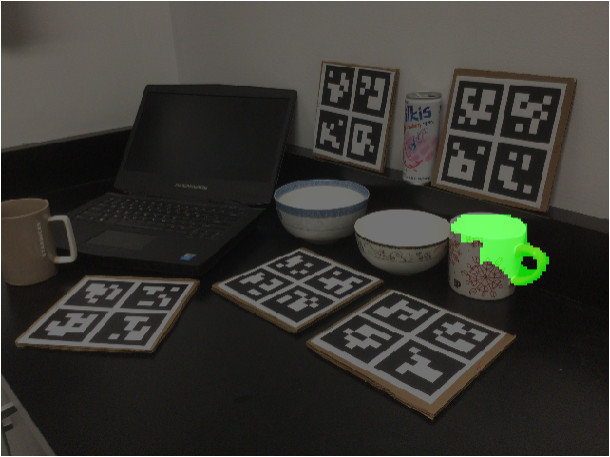}
    \end{overpic}
\end{minipage}
\vspace*{1mm}
\hspace*{0.3mm}
\begin{minipage}{0.31\textwidth}
    \centering (a) Prompt: \texttt{Gray open laptop}
\end{minipage}
\hspace*{1mm}
\begin{minipage}{0.31\textwidth}
    \centering (b) Prompt: \texttt{White small camera}
\end{minipage}
\hspace*{1mm}
\begin{minipage}{0.31\textwidth}
    \centering (c) Prompt: \texttt{Brown mug}
\end{minipage}
\hspace*{1mm}

\hspace*{0.3mm}
\begin{minipage}{0.155\textwidth}
    \begin{overpic}[width=1\textwidth]{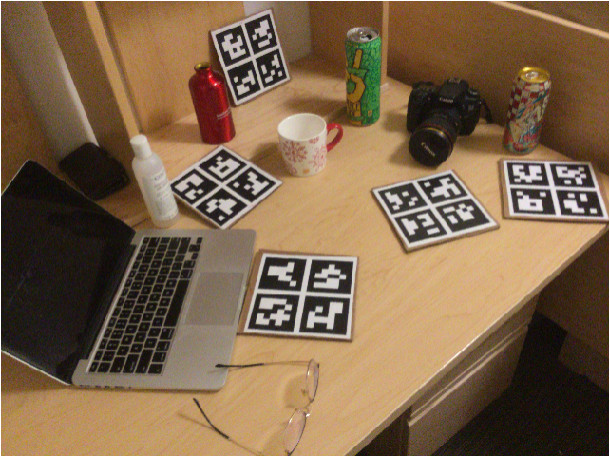}
    \end{overpic}
\end{minipage}
\begin{minipage}{0.155\textwidth}
    \begin{overpic}[width=1\textwidth]{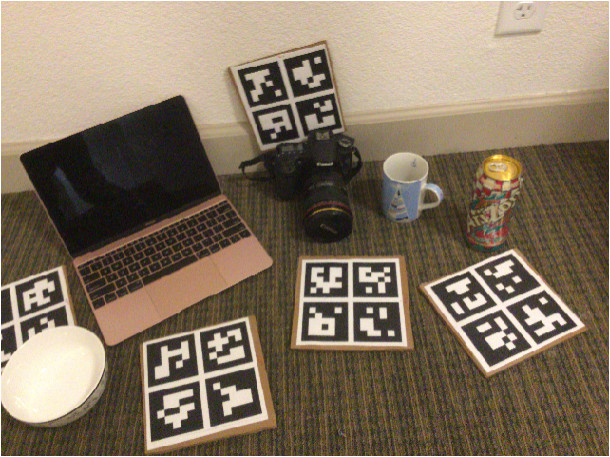}
    \end{overpic}
\end{minipage}
\hspace*{1mm}
\begin{minipage}{0.155\textwidth}
    \begin{overpic}[width=1\textwidth]{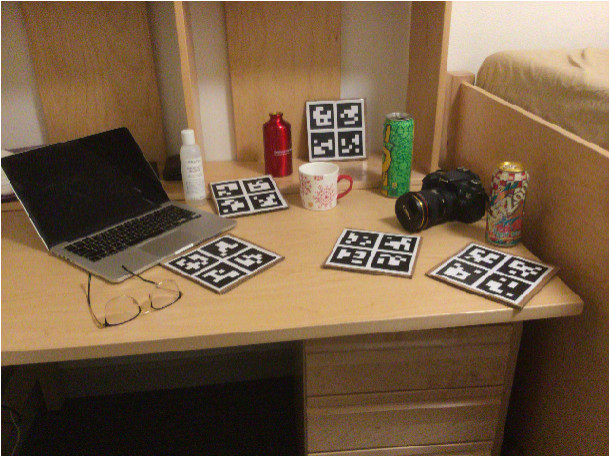}
    \end{overpic}
\end{minipage}
\begin{minipage}{0.155\textwidth}
    \begin{overpic}[width=1\textwidth]{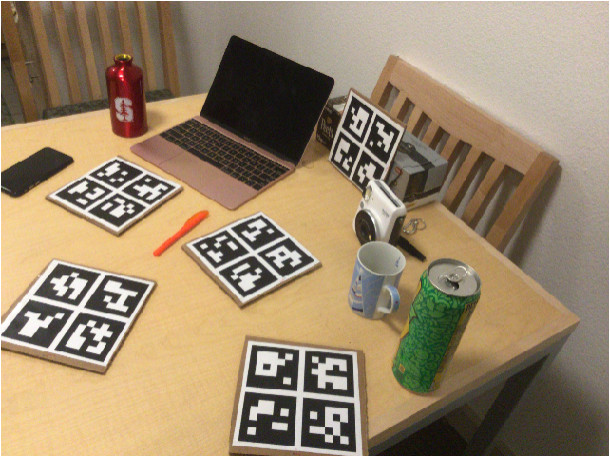}
    \end{overpic}
\end{minipage}
\hspace*{1mm}
\begin{minipage}{0.155\textwidth}
    \begin{overpic}[width=1\textwidth]{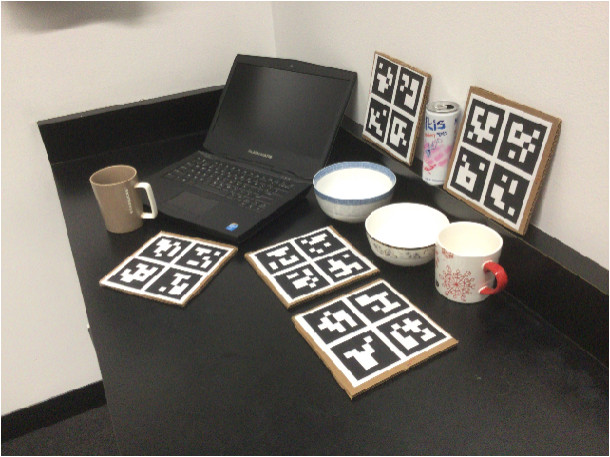}
    \end{overpic}
\end{minipage}
\begin{minipage}{0.155\textwidth}
    \begin{overpic}[width=1\textwidth]{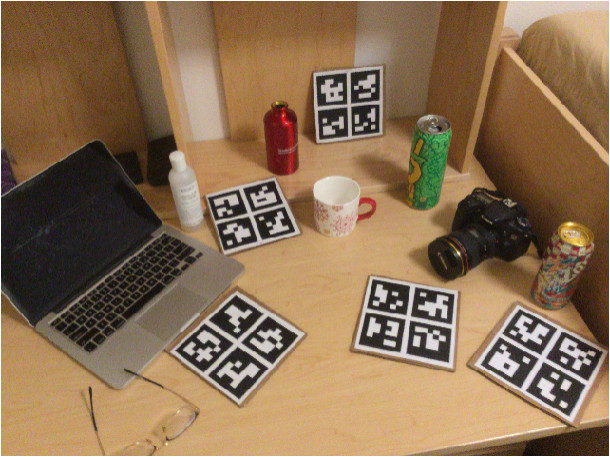}
    \end{overpic}
\end{minipage}
\vspace*{1mm}

\hspace*{0.3mm}
\begin{minipage}{0.155\textwidth}
    \begin{overpic}[width=1\textwidth]{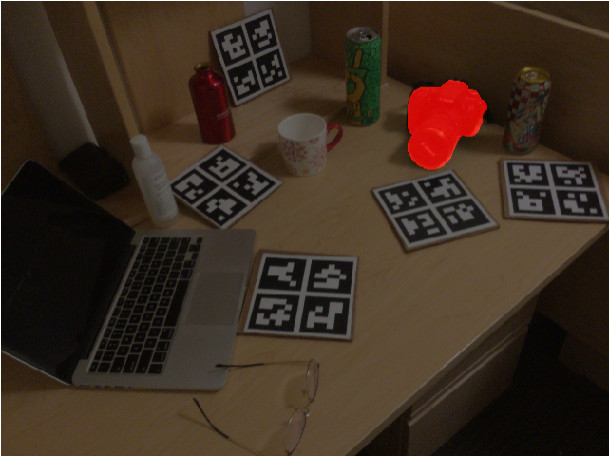}
    \put(-12,8){\small{\rotatebox{90}{Ground truth}}}
    \end{overpic}
\end{minipage}
\begin{minipage}{0.155\textwidth}
    \begin{overpic}[width=1\textwidth]{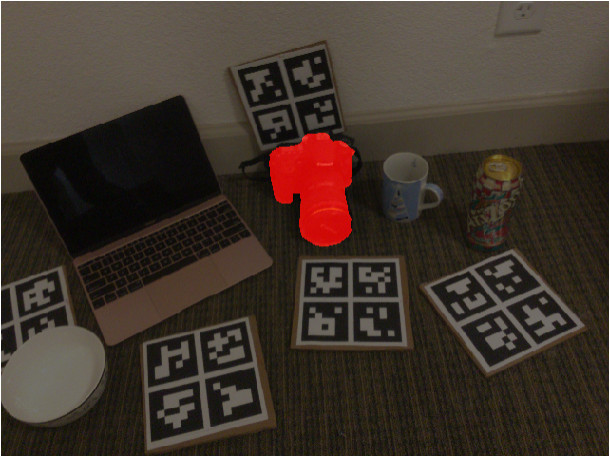}
    \end{overpic}
\end{minipage}
\hspace*{1mm}
\begin{minipage}{0.155\textwidth}
    \begin{overpic}[width=1\textwidth]{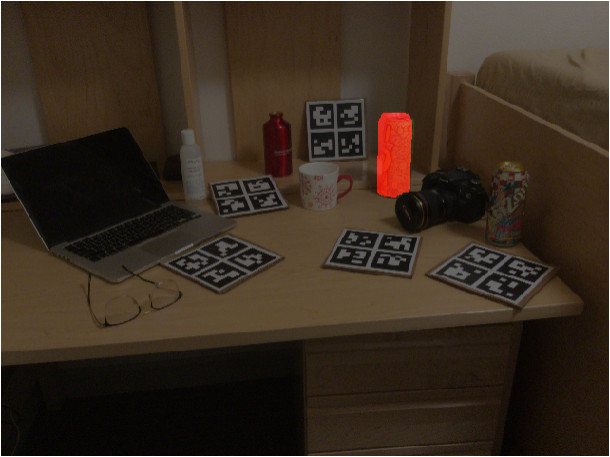}
    \end{overpic}
\end{minipage}
\begin{minipage}{0.155\textwidth}
    \begin{overpic}[width=1\textwidth]{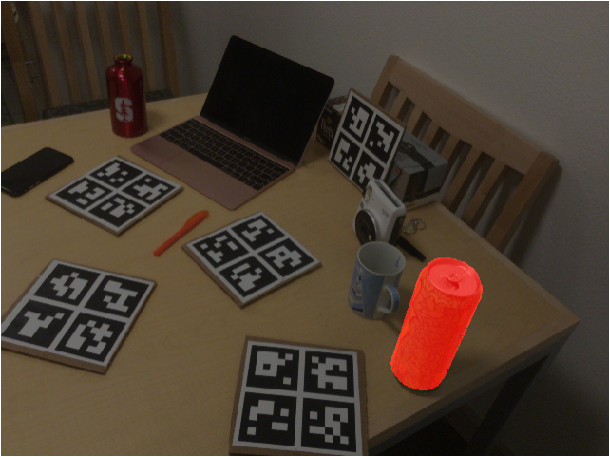}
    \end{overpic}
\end{minipage}
\hspace*{1mm}
\begin{minipage}{0.155\textwidth}
    \begin{overpic}[width=1\textwidth]{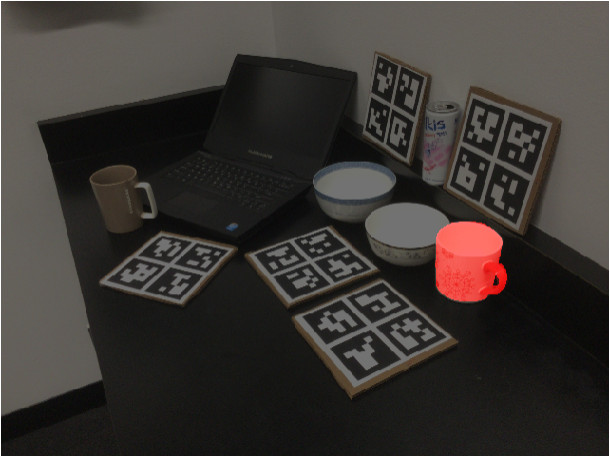}
    \end{overpic}
\end{minipage}
\begin{minipage}{0.155\textwidth}
    \begin{overpic}[width=1\textwidth]{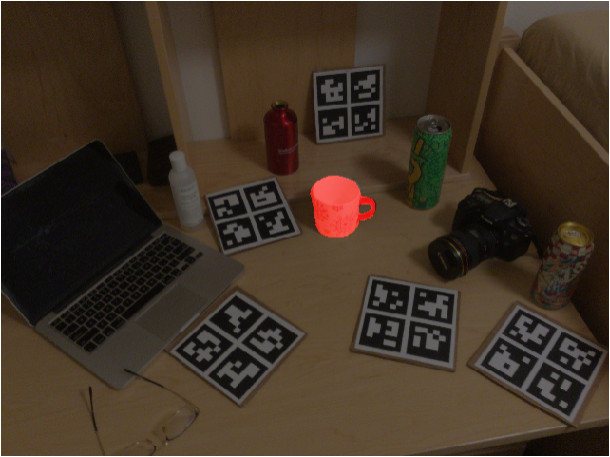}
    \end{overpic}
\end{minipage}
\vspace*{1mm}

\hspace*{0.3mm}
\begin{minipage}{0.155\textwidth}
    \begin{overpic}[width=1\textwidth]{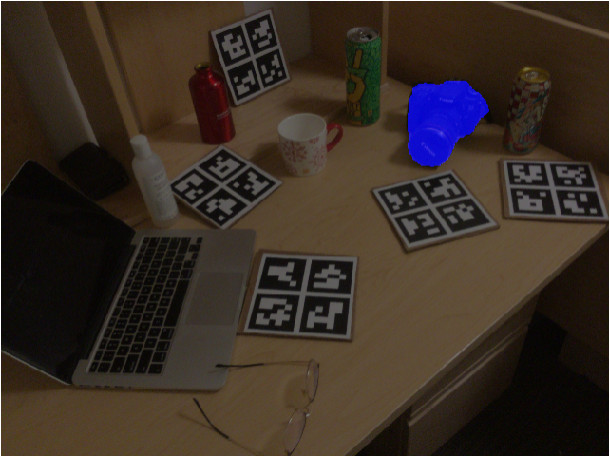}
    \put(-12,10){\small{\rotatebox{90}{OVSeg~\cite{liang2023ovseg}}}}
    \end{overpic}
\end{minipage}
\begin{minipage}{0.155\textwidth}
    \begin{overpic}[width=1\textwidth]{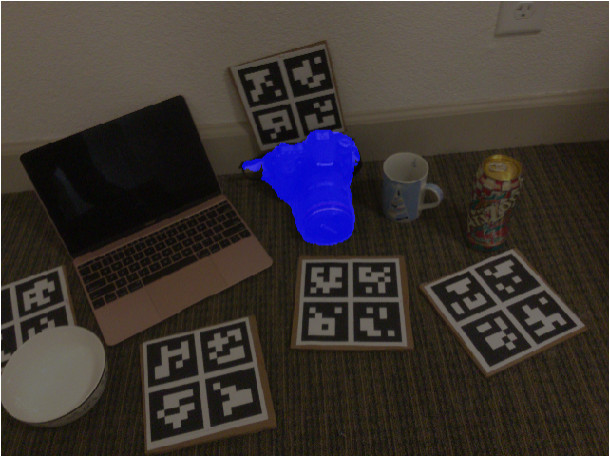}
    \end{overpic}
\end{minipage}
\hspace*{1mm}
\begin{minipage}{0.155\textwidth}
    \begin{overpic}[width=1\textwidth]{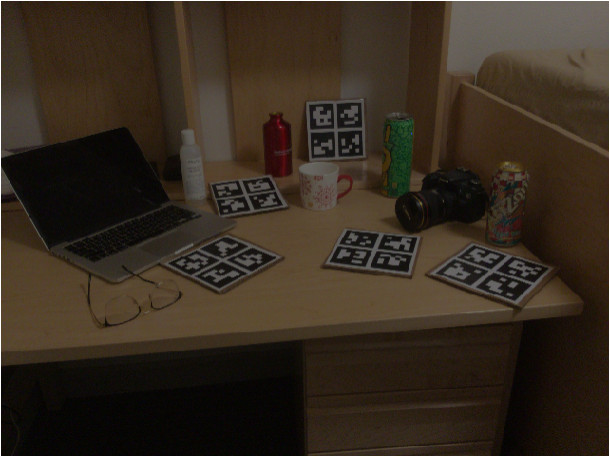}
    \end{overpic}
\end{minipage}
\begin{minipage}{0.155\textwidth}
    \begin{overpic}[width=1\textwidth]{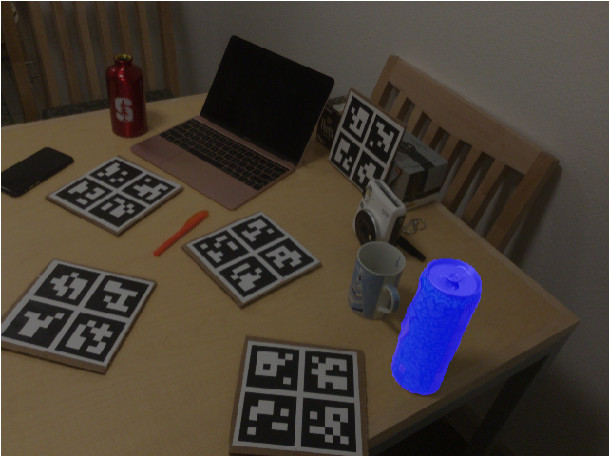}
    \end{overpic}
\end{minipage}
\hspace*{1mm}
\begin{minipage}{0.155\textwidth}
    \begin{overpic}[width=1\textwidth]{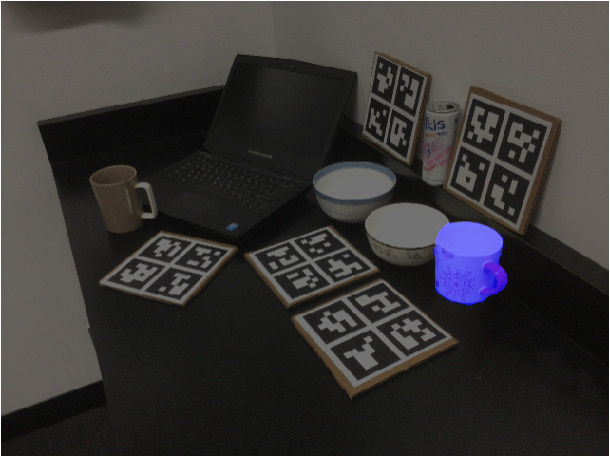}
    \end{overpic}
\end{minipage}
\begin{minipage}{0.155\textwidth}
    \begin{overpic}[width=1\textwidth]{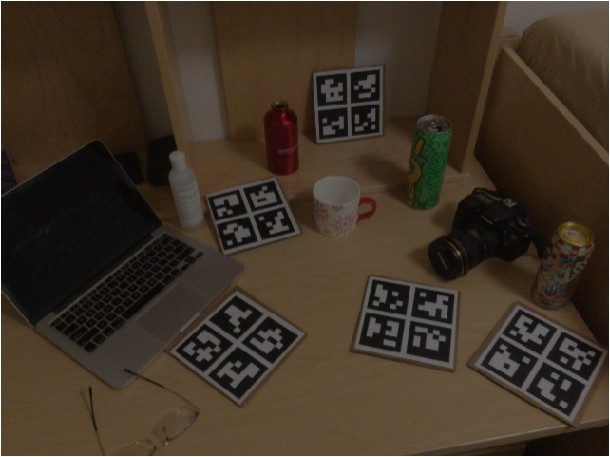}
    \end{overpic}
\end{minipage}
\vspace*{1mm}

\begin{minipage}{0.155\textwidth}
    \begin{overpic}[width=1\textwidth]{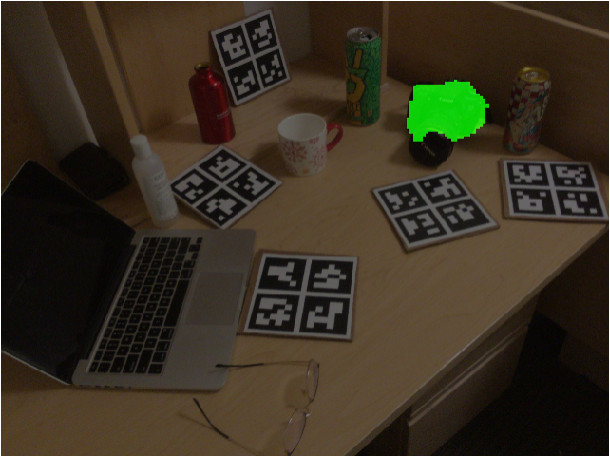}
    \put(-12,18){\small{\rotatebox{90}{\acronym}}}
    \end{overpic}
\end{minipage}
\begin{minipage}{0.155\textwidth}
    \begin{overpic}[width=1\textwidth]{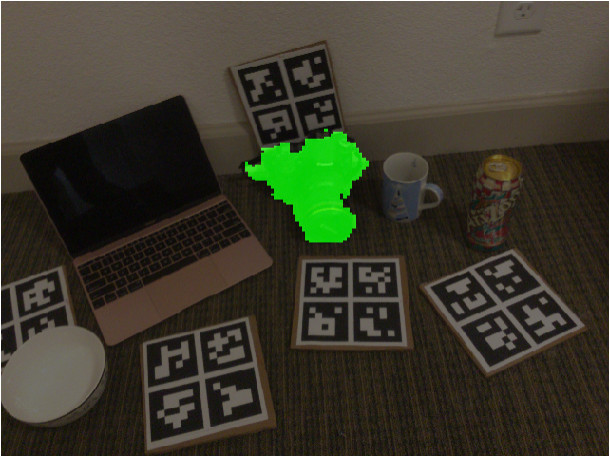}
    \end{overpic}
\end{minipage}
\hspace*{1mm}
\begin{minipage}{0.155\textwidth}
    \begin{overpic}[width=1\textwidth]{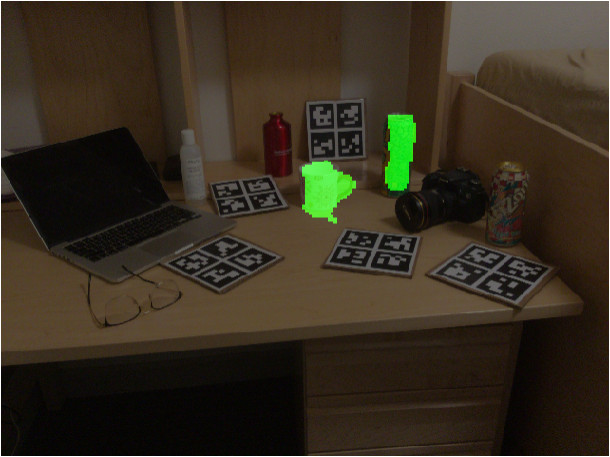}
    \end{overpic}
\end{minipage}
\begin{minipage}{0.155\textwidth}
    \begin{overpic}[width=1\textwidth]{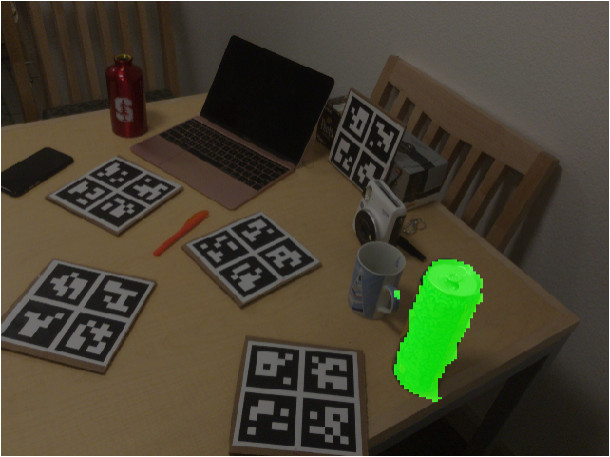}
    \end{overpic}
\end{minipage}
\hspace*{1mm}
\begin{minipage}{0.155\textwidth}
    \begin{overpic}[width=1\textwidth]{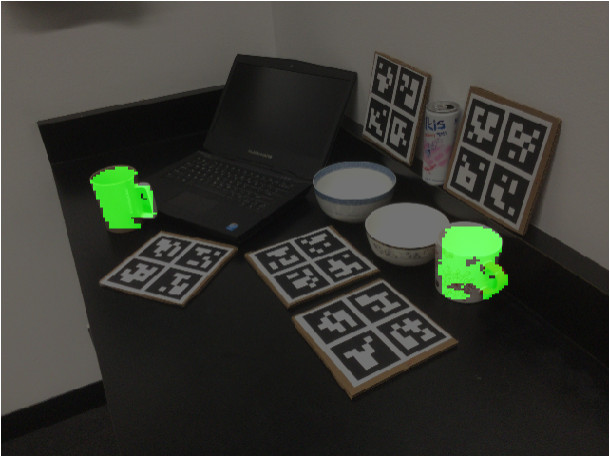}
    \end{overpic}
\end{minipage}
\begin{minipage}{0.155\textwidth}
    \begin{overpic}[width=1\textwidth]{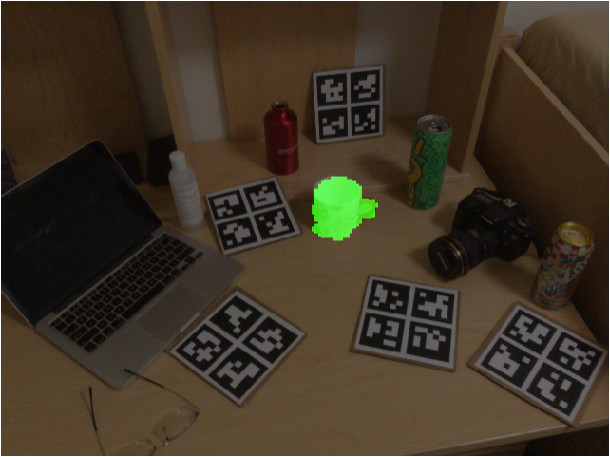}
    \end{overpic}
\end{minipage}
\vspace*{1mm}
\hspace*{0.3mm}
\begin{minipage}{0.31\textwidth}
    \centering (d) Prompt: \texttt{Black lens camera}
\end{minipage}
\hspace*{1mm}
\begin{minipage}{0.31\textwidth}
    \centering (e) Prompt: \texttt{Tall green can}
\end{minipage}
\hspace*{1mm}
\begin{minipage}{0.31\textwidth}
    \centering (f) Prompt: \texttt{White mug}
\end{minipage}
\hspace*{1mm}

%% file: supp/figures/qualitative/segm_toyl/segm.tex
\hspace*{0.3mm}
\vspace{1mm}
\begin{minipage}{0.155\textwidth}
    \centering{\small{Anchor}}
\end{minipage}
\begin{minipage}{0.155\textwidth}
    \centering{\small{Query}}
\end{minipage}
\begin{minipage}{0.155\textwidth}
    \centering{\small{Anchor}}
\end{minipage}
\begin{minipage}{0.155\textwidth}
    \centering{\small{Query}}
\end{minipage}
\begin{minipage}{0.155\textwidth}
    \centering{\small{Anchor}}
\end{minipage}
\begin{minipage}{0.155\textwidth}
    \centering{\small{Query}}
\end{minipage}
\hspace*{0.3mm}
\begin{minipage}{0.155\textwidth}
    \begin{overpic}[width=1\textwidth]{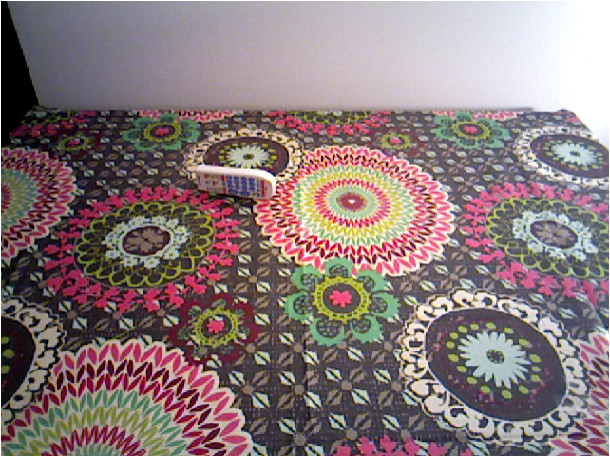}
    \end{overpic}
\end{minipage}
\begin{minipage}{0.155\textwidth}
    \begin{overpic}[width=1\textwidth]{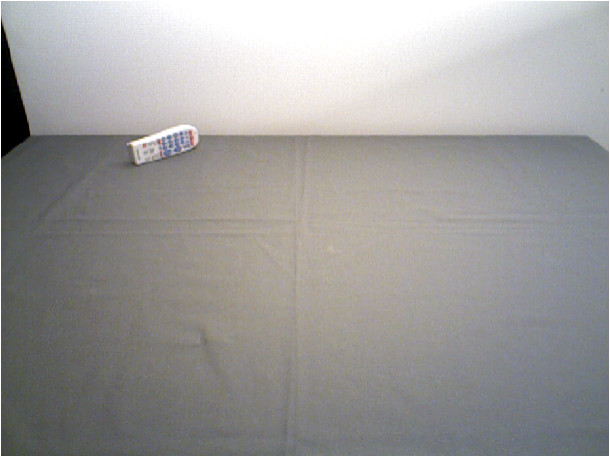}
    \end{overpic}
\end{minipage}
\hspace*{1mm}
\begin{minipage}{0.155\textwidth}
    \begin{overpic}[width=1\textwidth]{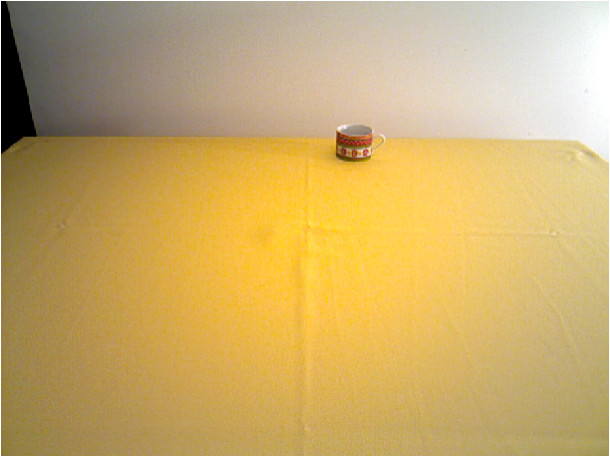}
    \end{overpic}
\end{minipage}
\begin{minipage}{0.155\textwidth}
    \begin{overpic}[width=1\textwidth]{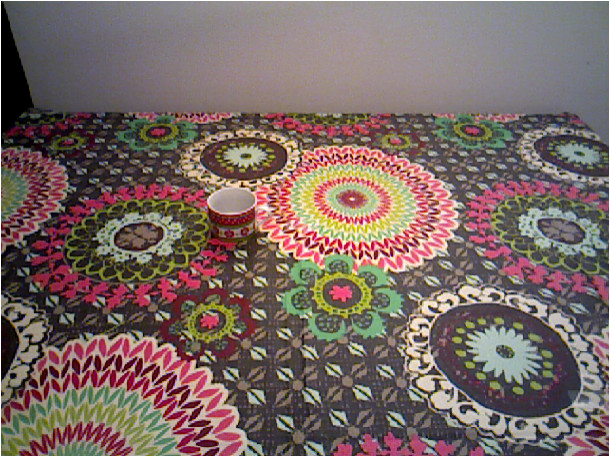}
    \end{overpic}
\end{minipage}
\hspace*{1mm}
\begin{minipage}{0.155\textwidth}
    \begin{overpic}[width=1\textwidth]{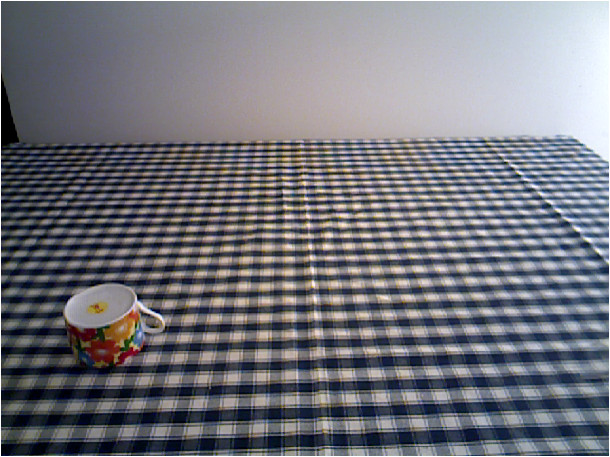}
    \end{overpic}
\end{minipage}
\begin{minipage}{0.155\textwidth}
    \begin{overpic}[width=1\textwidth]{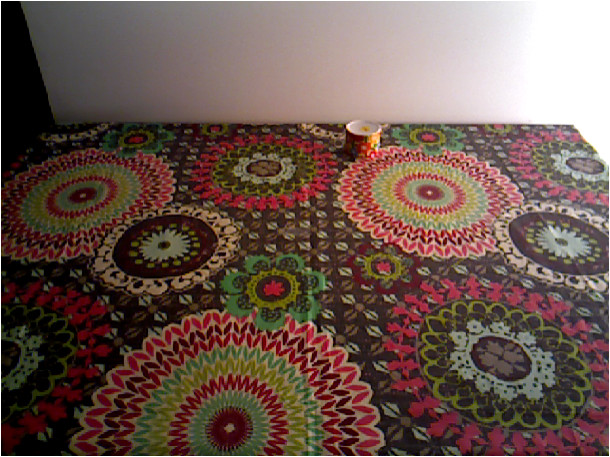}
    \end{overpic}
\end{minipage}
\vspace*{1mm}

\hspace*{0.3mm}
\begin{minipage}{0.155\textwidth}
    \begin{overpic}[width=1\textwidth]{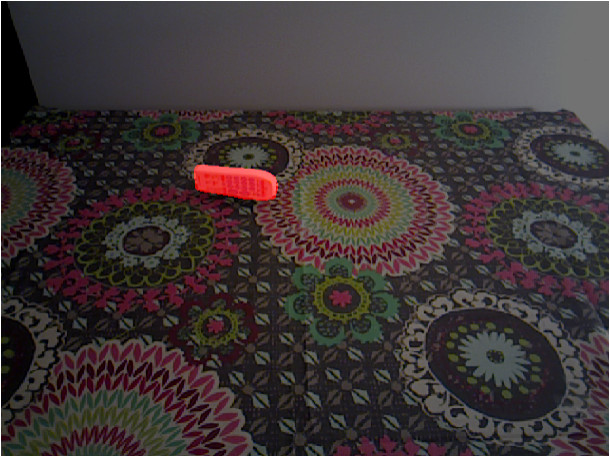}
    \put(-12,8){\small{\rotatebox{90}{Ground truth}}}
    \end{overpic}
\end{minipage}
\begin{minipage}{0.155\textwidth}
    \begin{overpic}[width=1\textwidth]{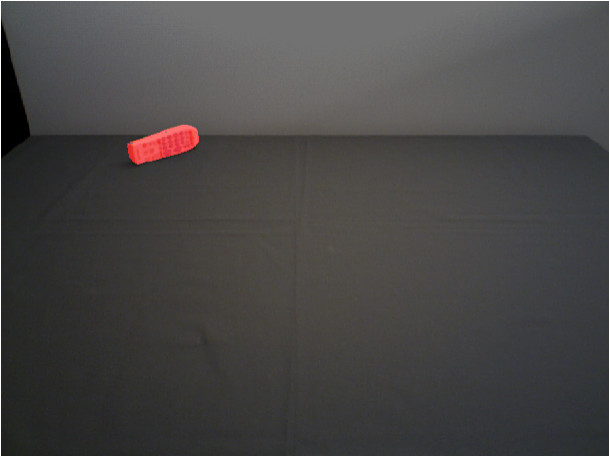}
    \end{overpic}
\end{minipage}
\hspace*{1mm}
\begin{minipage}{0.155\textwidth}
    \begin{overpic}[width=1\textwidth]{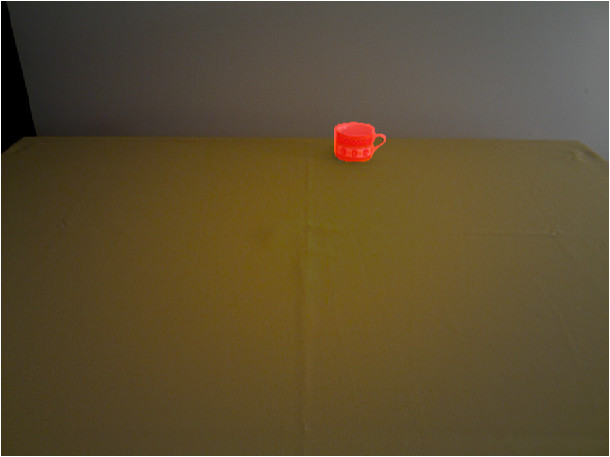}
    \end{overpic}
\end{minipage}
\begin{minipage}{0.155\textwidth}
    \begin{overpic}[width=1\textwidth]{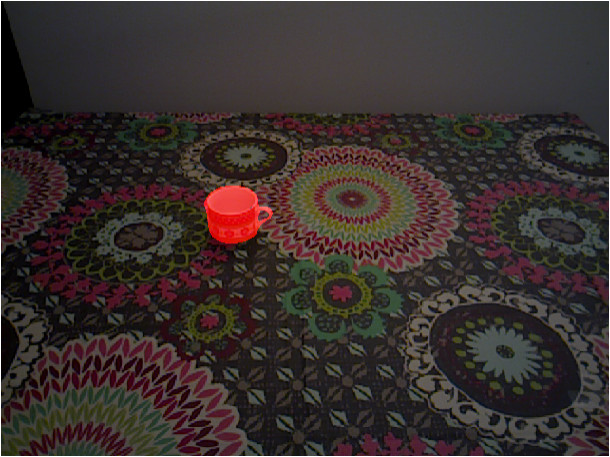}
    \end{overpic}
\end{minipage}
\hspace*{1mm}
\begin{minipage}{0.155\textwidth}
    \begin{overpic}[width=1\textwidth]{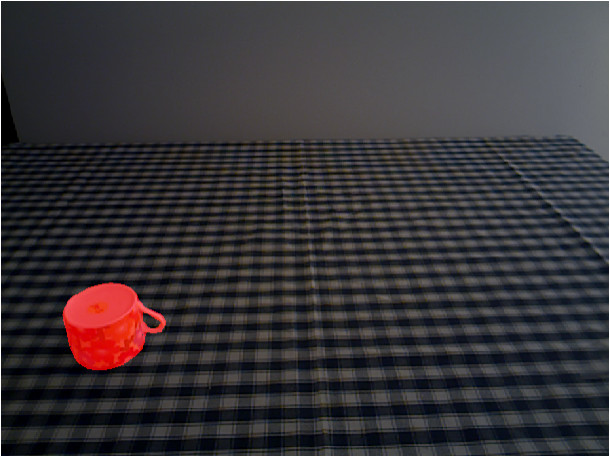}
    \end{overpic}
\end{minipage}
\begin{minipage}{0.155\textwidth}
    \begin{overpic}[width=1\textwidth]{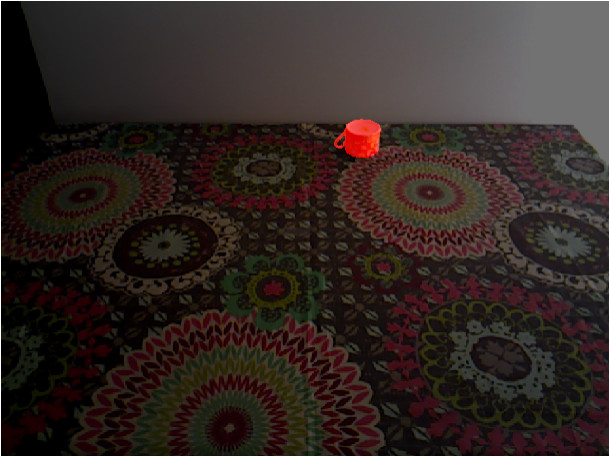}
    \end{overpic}
\end{minipage}
\vspace*{1mm}

\hspace*{0.3mm}
\begin{minipage}{0.155\textwidth}
    \begin{overpic}[width=1\textwidth]{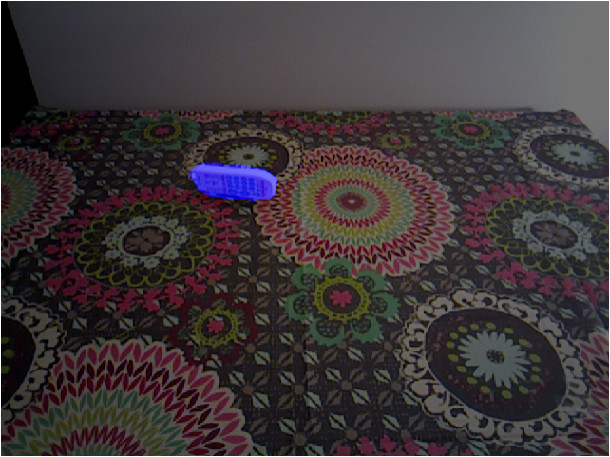}
    \put(-12,10){\small{\rotatebox{90}{OVSeg~\cite{liang2023ovseg}}}}
    \end{overpic}
\end{minipage}
\begin{minipage}{0.155\textwidth}
    \begin{overpic}[width=1\textwidth]{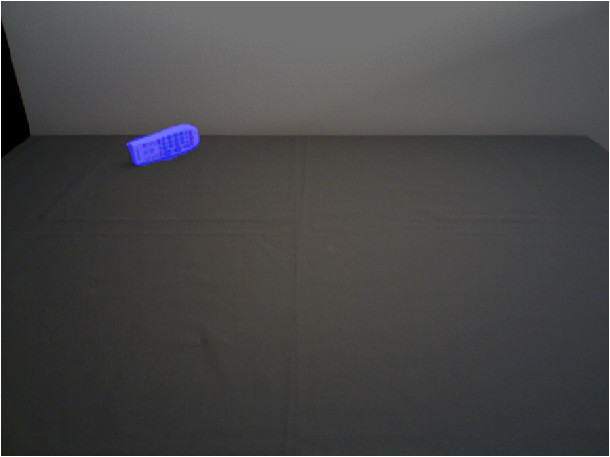}
    \end{overpic}
\end{minipage}
\hspace*{1mm}
\begin{minipage}{0.155\textwidth}
    \begin{overpic}[width=1\textwidth]{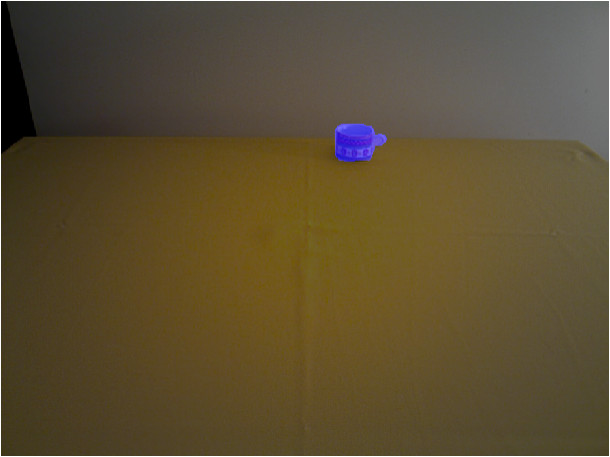}
    \end{overpic}
\end{minipage}
\begin{minipage}{0.155\textwidth}
    \begin{overpic}[width=1\textwidth]{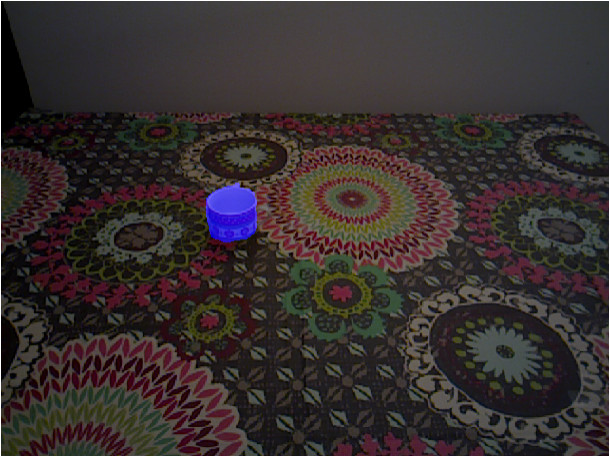}
    \end{overpic}
\end{minipage}
\hspace*{1mm}
\begin{minipage}{0.155\textwidth}
    \begin{overpic}[width=1\textwidth]{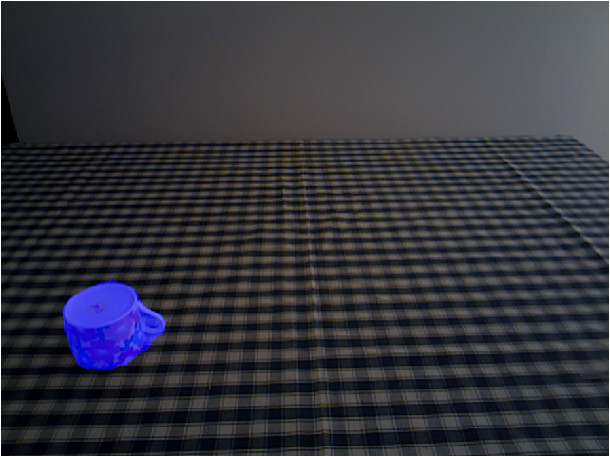}
    \end{overpic}
\end{minipage}
\begin{minipage}{0.155\textwidth}
    \begin{overpic}[width=1\textwidth]{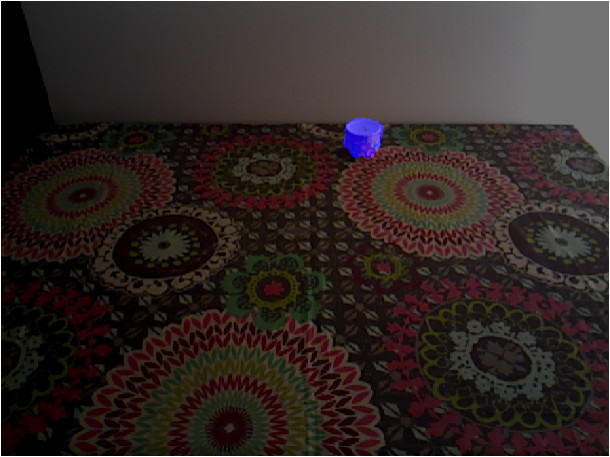}
    \end{overpic}
\end{minipage}
\vspace*{1mm}

\begin{minipage}{0.155\textwidth}
    \begin{overpic}[width=1\textwidth]{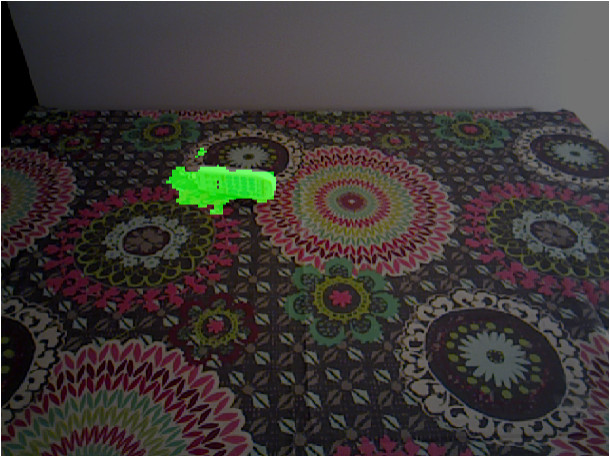}
    \put(-12,18){\small{\rotatebox{90}{\acronym}}}
    \end{overpic}
\end{minipage}
\begin{minipage}{0.155\textwidth}
    \begin{overpic}[width=1\textwidth]{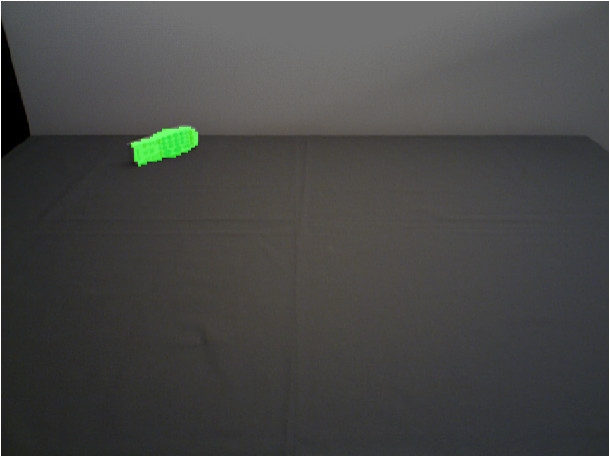}
    \end{overpic}
\end{minipage}
\hspace*{1mm}
\begin{minipage}{0.155\textwidth}
    \begin{overpic}[width=1\textwidth]{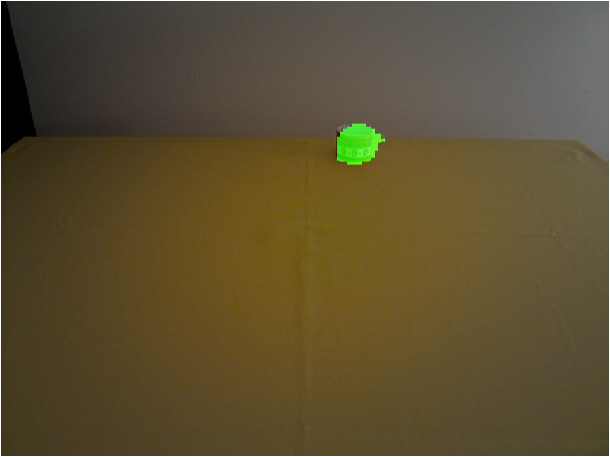}
    \end{overpic}
\end{minipage}
\begin{minipage}{0.155\textwidth}
    \begin{overpic}[width=1\textwidth]{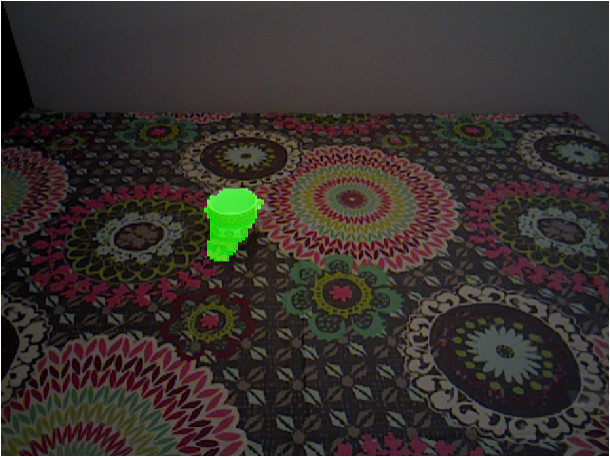}
    \end{overpic}
\end{minipage}
\hspace*{1mm}
\begin{minipage}{0.155\textwidth}
    \begin{overpic}[width=1\textwidth]{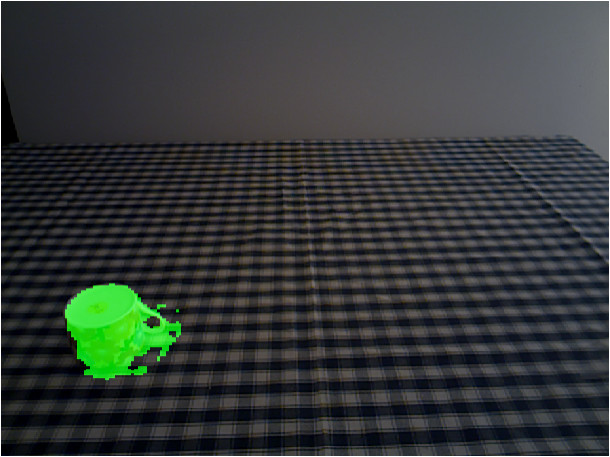}
    \end{overpic}
\end{minipage}
\begin{minipage}{0.155\textwidth}
    \begin{overpic}[width=1\textwidth]{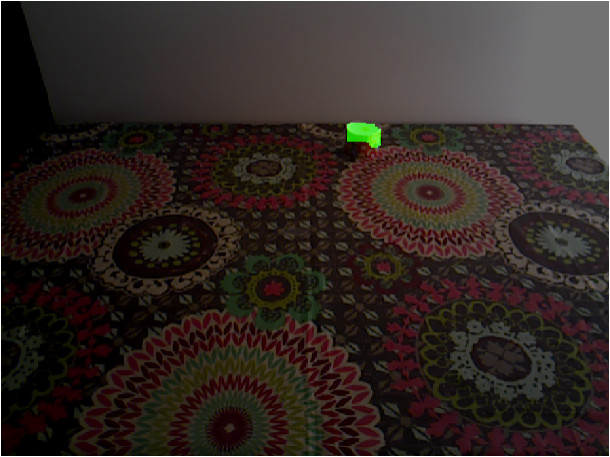}
    \end{overpic}
\end{minipage}
\vspace*{1mm}
\hspace*{0.3mm}
\begin{minipage}{0.31\textwidth}
    \centering (a) Prompt: \texttt{White and blue remote}
\end{minipage}
\hspace*{1mm}
\begin{minipage}{0.31\textwidth}
    \centering (b) Prompt: \texttt{Red and white small mug}
\end{minipage}
\hspace*{1mm}
\begin{minipage}{0.31\textwidth}
    \centering (c) Prompt: \texttt{Large green mug}
\end{minipage}
\hspace*{1mm}

\hspace*{0.3mm}
\begin{minipage}{0.155\textwidth}
    \begin{overpic}[width=1\textwidth]{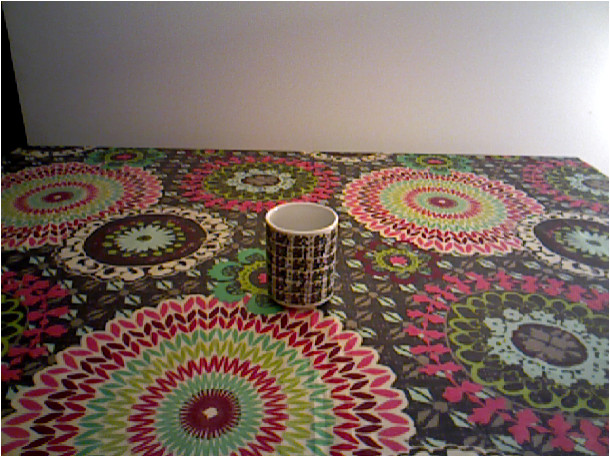}
    \end{overpic}
\end{minipage}
\begin{minipage}{0.155\textwidth}
    \begin{overpic}[width=1\textwidth]{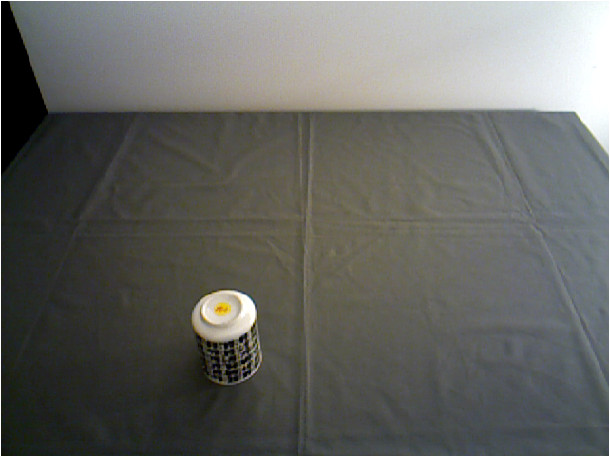}
    \end{overpic}
\end{minipage}
\hspace*{1mm}
\begin{minipage}{0.155\textwidth}
    \begin{overpic}[width=1\textwidth]{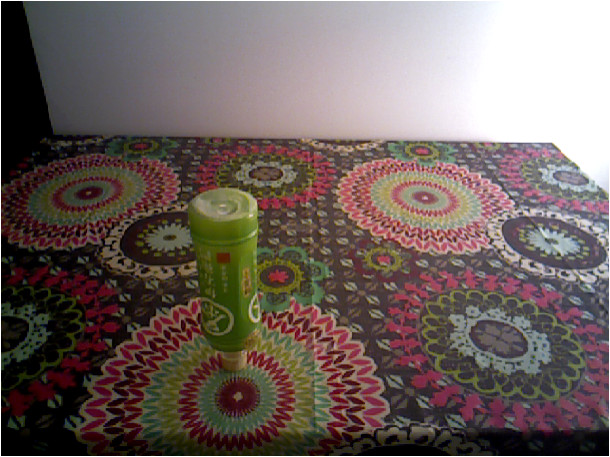}
    \end{overpic}
\end{minipage}
\begin{minipage}{0.155\textwidth}
    \begin{overpic}[width=1\textwidth]{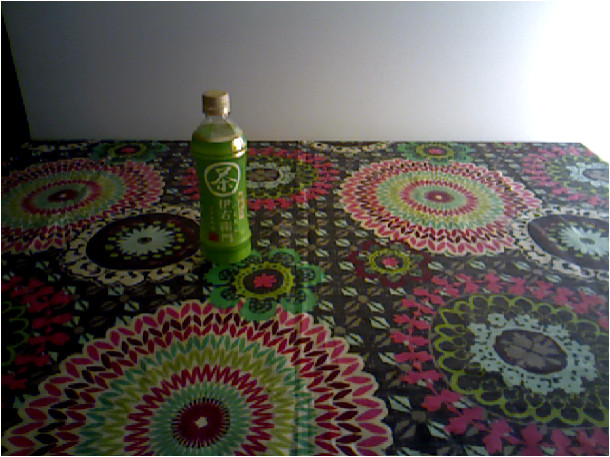}
    \end{overpic}
\end{minipage}
\hspace*{1mm}
\begin{minipage}{0.155\textwidth}
    \begin{overpic}[width=1\textwidth]{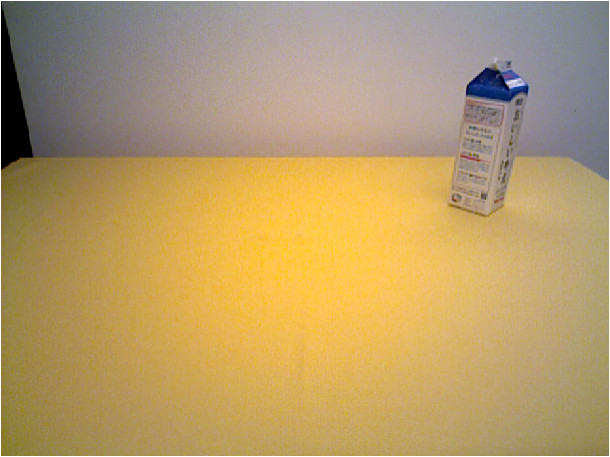}
    \end{overpic}
\end{minipage}
\begin{minipage}{0.155\textwidth}
    \begin{overpic}[width=1\textwidth]{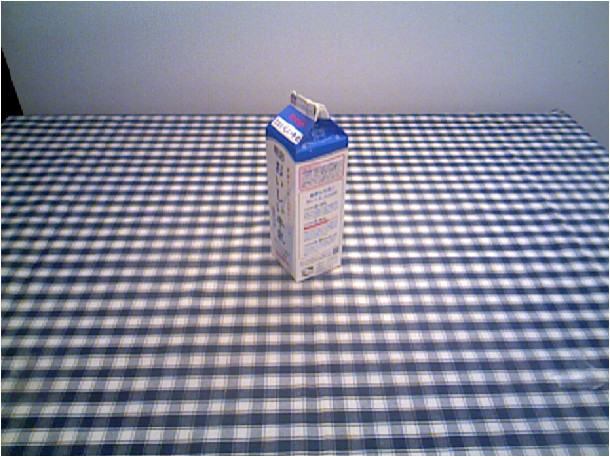}
    \end{overpic}
\end{minipage}
\vspace*{1mm}

\hspace*{0.3mm}
\begin{minipage}{0.155\textwidth}
    \begin{overpic}[width=1\textwidth]{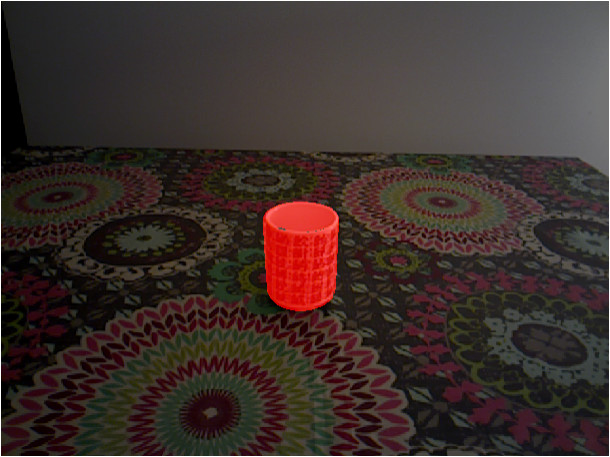}
    \put(-12,8){\small{\rotatebox{90}{Ground truth}}}
    \end{overpic}
\end{minipage}
\begin{minipage}{0.155\textwidth}
    \begin{overpic}[width=1\textwidth]{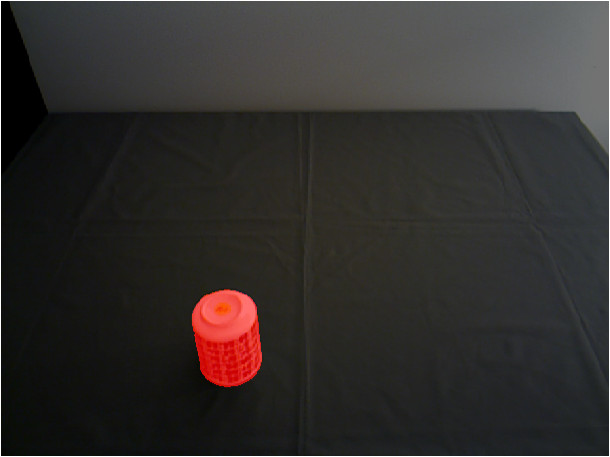}
    \end{overpic}
\end{minipage}
\hspace*{1mm}
\begin{minipage}{0.155\textwidth}
    \begin{overpic}[width=1\textwidth]{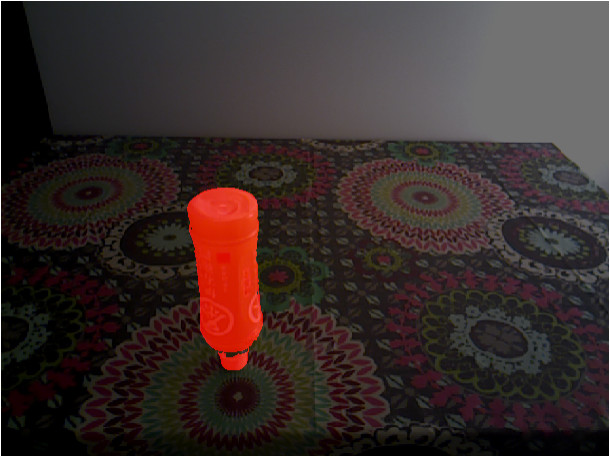}
    \end{overpic}
\end{minipage}
\begin{minipage}{0.155\textwidth}
    \begin{overpic}[width=1\textwidth]{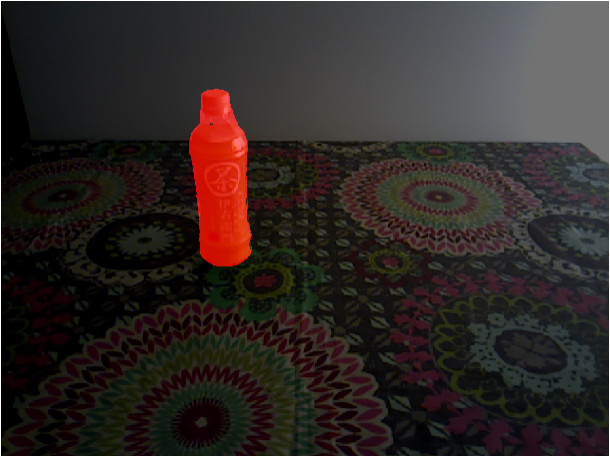}
    \end{overpic}
\end{minipage}
\hspace*{1mm}
\begin{minipage}{0.155\textwidth}
    \begin{overpic}[width=1\textwidth]{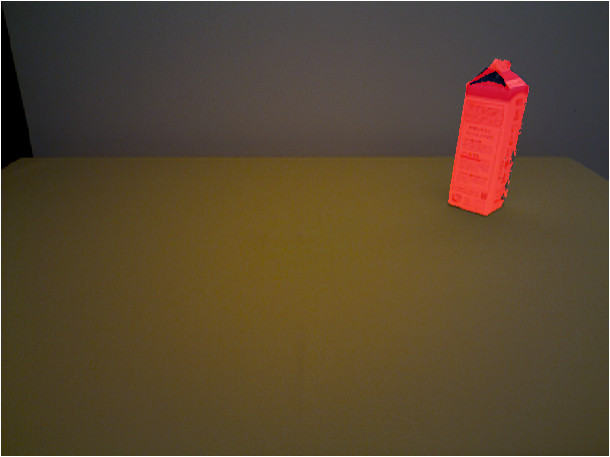}
    \end{overpic}
\end{minipage}
\begin{minipage}{0.155\textwidth}
    \begin{overpic}[width=1\textwidth]{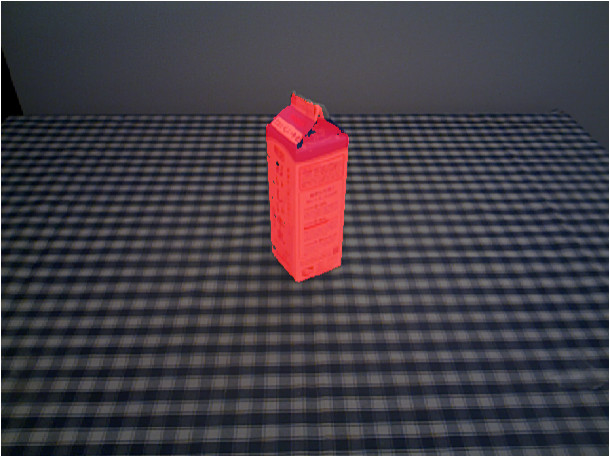}
    \end{overpic}
\end{minipage}
\vspace*{1mm}

\hspace*{0.3mm}
\begin{minipage}{0.155\textwidth}
    \begin{overpic}[width=1\textwidth]{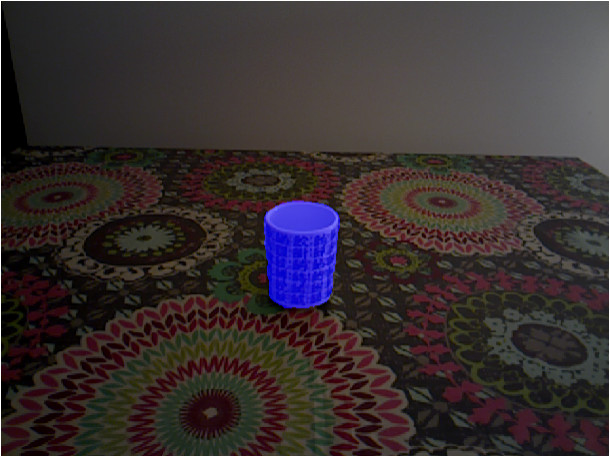}
    \put(-12,10){\small{\rotatebox{90}{OVSeg~\cite{liang2023ovseg}}}}
    \end{overpic}
\end{minipage}
\begin{minipage}{0.155\textwidth}
    \begin{overpic}[width=1\textwidth]{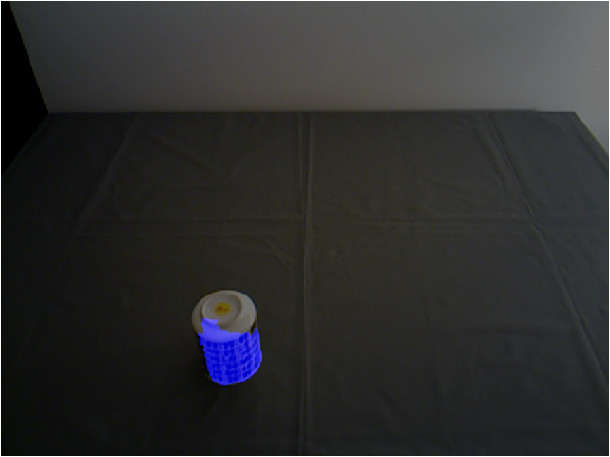}
    \end{overpic}
\end{minipage}
\hspace*{1mm}
\begin{minipage}{0.155\textwidth}
    \begin{overpic}[width=1\textwidth]{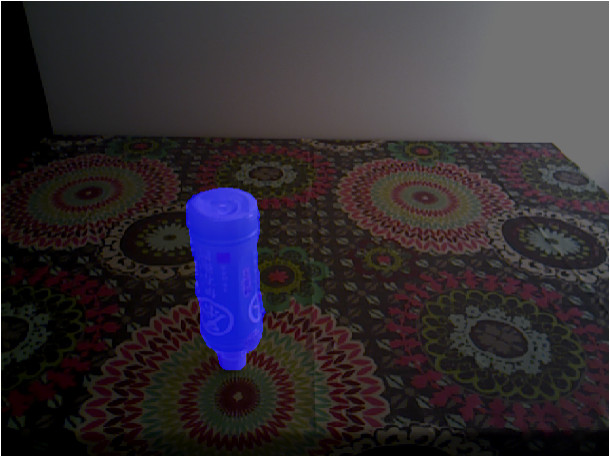}
    \end{overpic}
\end{minipage}
\begin{minipage}{0.155\textwidth}
    \begin{overpic}[width=1\textwidth]{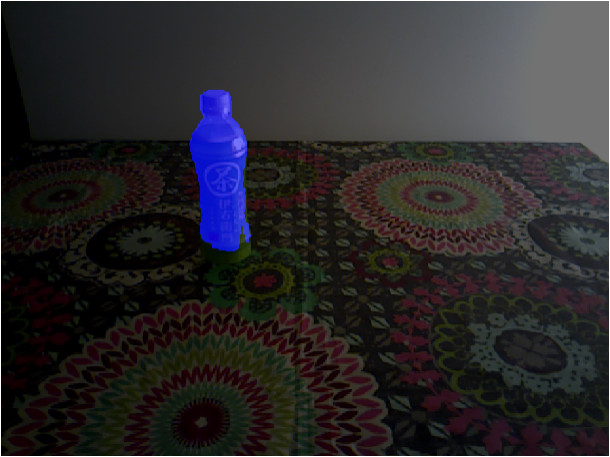}
    \end{overpic}
\end{minipage}
\hspace*{1mm}
\begin{minipage}{0.155\textwidth}
    \begin{overpic}[width=1\textwidth]{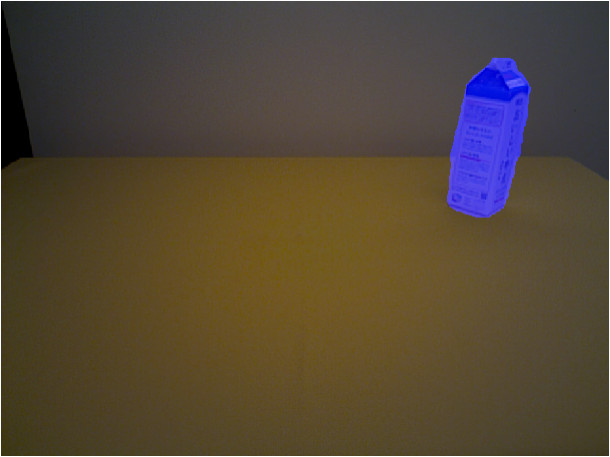}
    \end{overpic}
\end{minipage}
\begin{minipage}{0.155\textwidth}
    \begin{overpic}[width=1\textwidth]{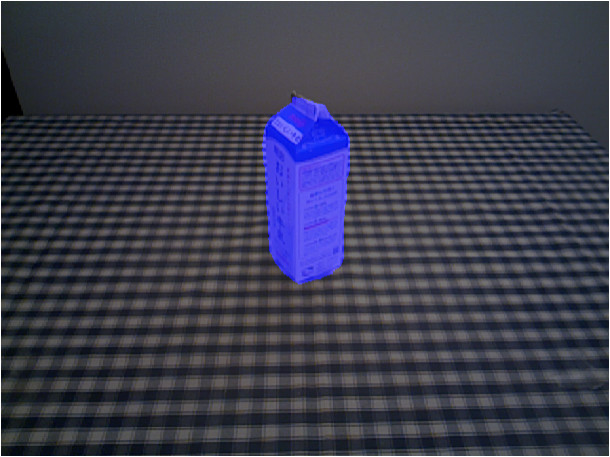}
    \end{overpic}
\end{minipage}
\vspace*{1mm}

\begin{minipage}{0.155\textwidth}
    \begin{overpic}[width=1\textwidth]{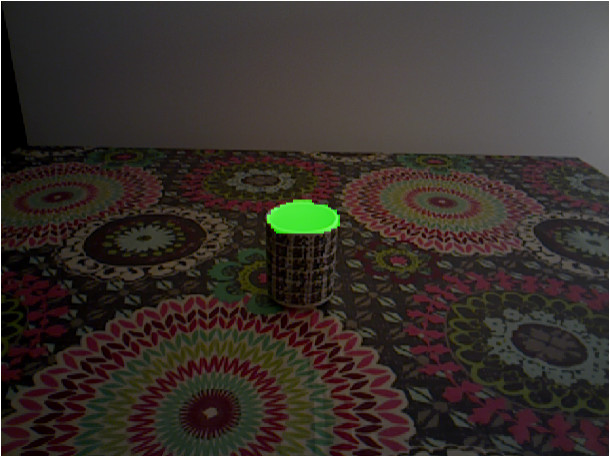}
    \put(-12,18){\small{\rotatebox{90}{\acronym}}}
    \end{overpic}
\end{minipage}
\begin{minipage}{0.155\textwidth}
    \begin{overpic}[width=1\textwidth]{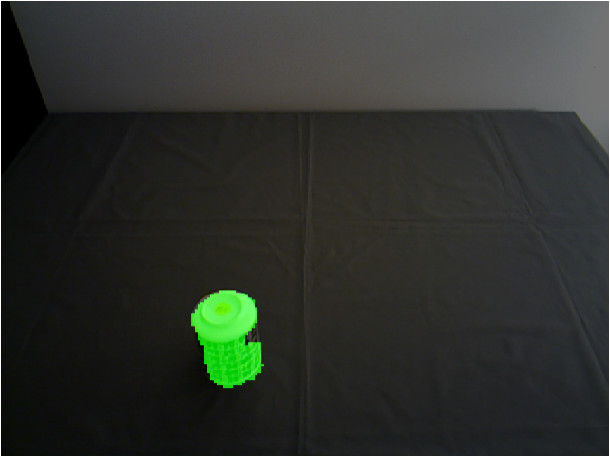}
    \end{overpic}
\end{minipage}
\hspace*{1mm}
\begin{minipage}{0.155\textwidth}
    \begin{overpic}[width=1\textwidth]{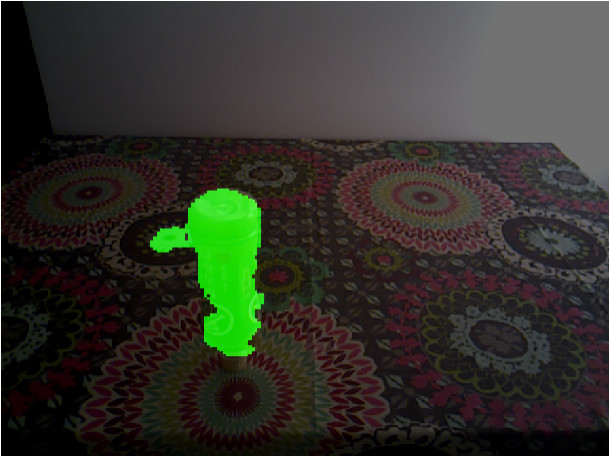}
    \end{overpic}
\end{minipage}
\begin{minipage}{0.155\textwidth}
    \begin{overpic}[width=1\textwidth]{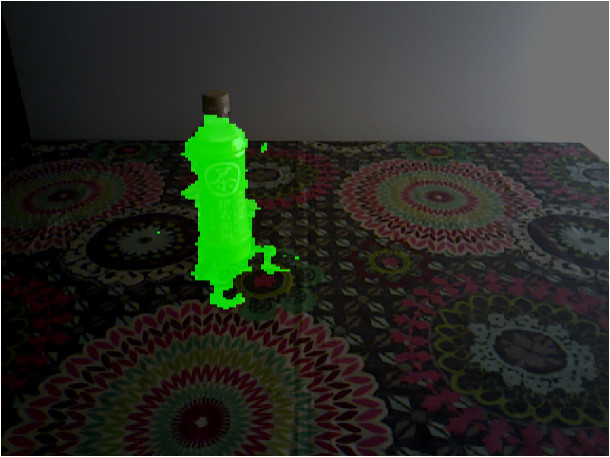}
    \end{overpic}
\end{minipage}
\hspace*{1mm}
\begin{minipage}{0.155\textwidth}
    \begin{overpic}[width=1\textwidth]{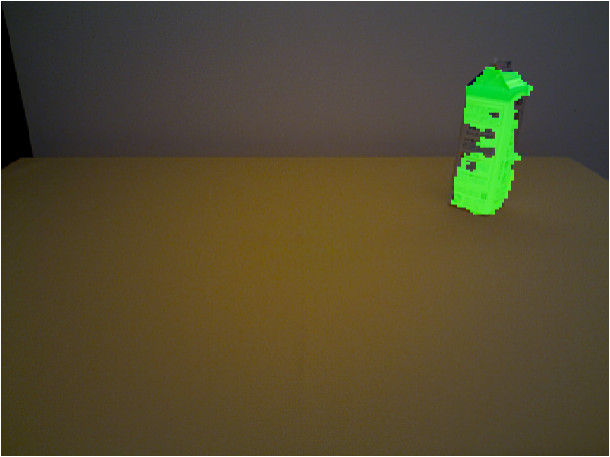}
    \end{overpic}
\end{minipage}
\begin{minipage}{0.155\textwidth}
    \begin{overpic}[width=1\textwidth]{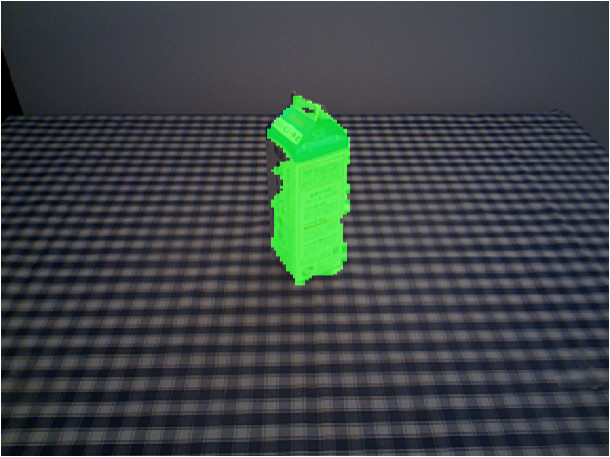}
    \end{overpic}
\end{minipage}
\vspace*{1mm}
\hspace*{0.3mm}
\begin{minipage}{0.31\textwidth}
    \centering (d) Prompt: \texttt{Black and white mug}
\end{minipage}
\hspace*{1mm}
\begin{minipage}{0.31\textwidth}
    \centering (e) Prompt: \texttt{Green plastic bottle}
\end{minipage}
\hspace*{1mm}
\begin{minipage}{0.31\textwidth}
    \centering (f) Prompt: \texttt{White and blue milk cartoon}
\end{minipage}
\hspace*{1mm}

%% file: supp/figures/qualitative/feat_nocs/feats.tex
\hspace*{0.3mm}
\vspace{1mm}
\begin{minipage}{0.155\textwidth}
    \centering{\small{Anchor}}
\end{minipage}
\begin{minipage}{0.155\textwidth}
    \centering{\small{Query}}
\end{minipage}
\begin{minipage}{0.155\textwidth}
    \centering{\small{Anchor}}
\end{minipage}
\begin{minipage}{0.155\textwidth}
    \centering{\small{Query}}
\end{minipage}
\begin{minipage}{0.155\textwidth}
    \centering{\small{Anchor}}
\end{minipage}
\begin{minipage}{0.155\textwidth}
    \centering{\small{Query}}
\end{minipage}
\hspace*{0.3mm}
\begin{minipage}{0.155\textwidth}
    \begin{overpic}[width=1\textwidth]{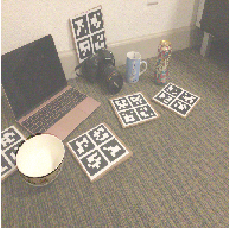}
    \put(6,28){\tiny{\color{black!10!green}{$\bullet$}}}
    \end{overpic}
\end{minipage}
\begin{minipage}{0.155\textwidth}
    \begin{overpic}[width=1\textwidth]{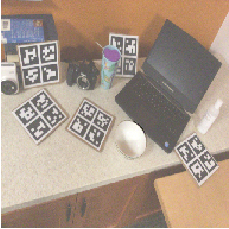}
    \end{overpic}
\end{minipage}
\hspace*{1mm}
\begin{minipage}{0.155\textwidth}
    \begin{overpic}[width=1\textwidth]{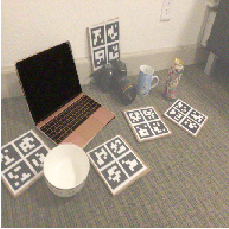}
    \put(25,72){\tiny{\color{black!10!green}{$\bullet$}}}
    \end{overpic}
\end{minipage}
\begin{minipage}{0.155\textwidth}
    \begin{overpic}[width=1\textwidth]{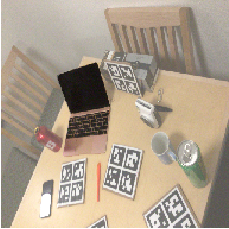}
    \end{overpic}
\end{minipage}
\hspace*{1mm}
\begin{minipage}{0.155\textwidth}
    \begin{overpic}[width=1\textwidth]{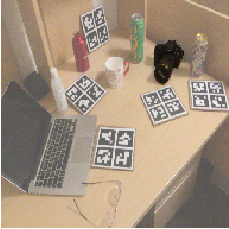}
    \put(70,75){\tiny{\color{black!10!green}{$\bullet$}}}
    \end{overpic}
\end{minipage}
\begin{minipage}{0.155\textwidth}
    \begin{overpic}[width=1\textwidth]{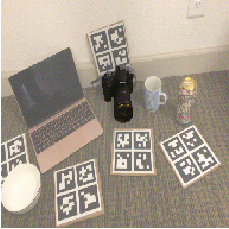}
    \end{overpic}
\end{minipage}
\vspace*{1mm}

\hspace*{0.3mm}
\begin{minipage}{0.155\textwidth}
    \begin{overpic}[width=1\textwidth]{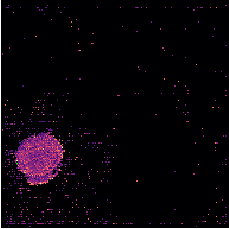}
    \put(-12,9){\small{\rotatebox{90}{Standard prompt}}}
    \put(6,28){\tiny{\color{black!10!green}{$\bullet$}}}
    
    \end{overpic}
\end{minipage}
\begin{minipage}{0.155\textwidth}
    \begin{overpic}[width=1\textwidth]{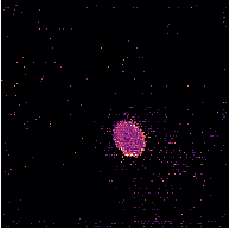}
    \end{overpic}
\end{minipage}
\hspace*{1mm}
\begin{minipage}{0.155\textwidth}
    \begin{overpic}[width=1\textwidth]{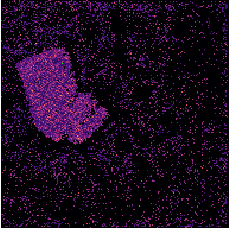}
    \put(25,72){\tiny{\color{black!10!green}{$\bullet$}}}
    \end{overpic}
\end{minipage}
\begin{minipage}{0.155\textwidth}
    \begin{overpic}[width=1\textwidth]{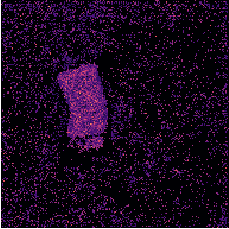}
    \end{overpic}
\end{minipage}
\hspace*{1mm}
\begin{minipage}{0.155\textwidth}
    \begin{overpic}[width=1\textwidth]{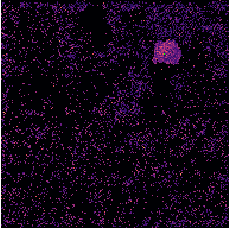}
    \put(70,75){\tiny{\color{black!10!green}{$\bullet$}}}
    \end{overpic}
\end{minipage}
\begin{minipage}{0.155\textwidth}
    \begin{overpic}[width=1\textwidth]{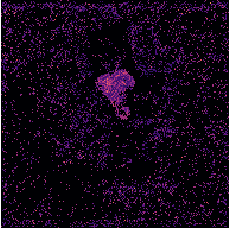}
    \end{overpic}
\end{minipage}
\vspace*{1mm}

\hspace*{0.3mm}
\begin{minipage}{0.155\textwidth}
    \begin{overpic}[width=1\textwidth]{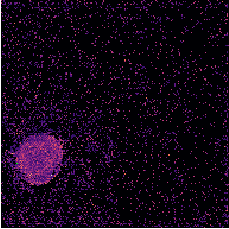}
    \put(-12,21){\small{\rotatebox{90}{Name only}}}
    \put(6,28){\tiny{\color{black!10!green}{$\bullet$}}}
    \end{overpic}
\end{minipage}
\begin{minipage}{0.155\textwidth}
    \begin{overpic}[width=1\textwidth]{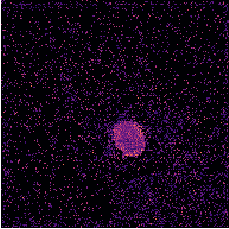}
    \end{overpic}
\end{minipage}
\hspace*{1mm}
\begin{minipage}{0.155\textwidth}
    \begin{overpic}[width=1\textwidth]{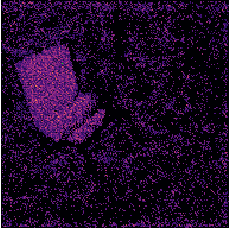}
    \put(25,72){\tiny{\color{black!10!green}{$\bullet$}}}
    \end{overpic}
\end{minipage}
\begin{minipage}{0.155\textwidth}
    \begin{overpic}[width=1\textwidth]{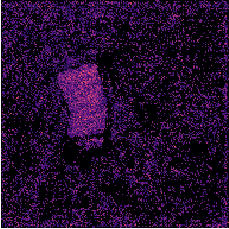}
    \end{overpic}
\end{minipage}
\hspace*{1mm}
\begin{minipage}{0.155\textwidth}
    \begin{overpic}[width=1\textwidth]{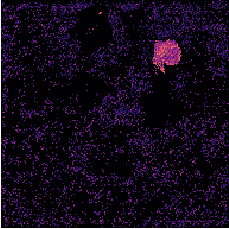}
    \put(70,75){\tiny{\color{black!10!green}{$\bullet$}}}
    \end{overpic}
\end{minipage}
\begin{minipage}{0.155\textwidth}
    \begin{overpic}[width=1\textwidth]{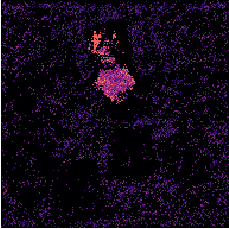}
    \end{overpic}
\end{minipage}
\vspace*{1mm}

\hspace*{0.3mm}
\begin{minipage}{0.155\textwidth}
    \begin{overpic}[width=1\textwidth]{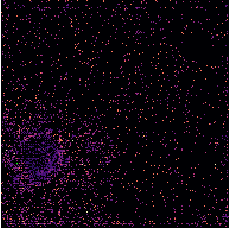}
    \put(-12,12){\small{\rotatebox{90}{Misleading desc.}}}
    \put(6,28){\tiny{\color{black!10!green}{$\bullet$}}}
    \end{overpic}
\end{minipage}
\begin{minipage}{0.155\textwidth}
    \begin{overpic}[width=1\textwidth]{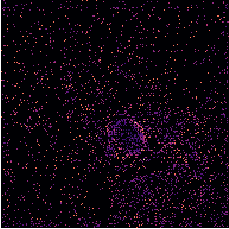}
    \end{overpic}
\end{minipage}
\hspace*{1mm}
\begin{minipage}{0.155\textwidth}
    \begin{overpic}[width=1\textwidth]{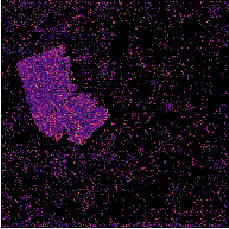}
    \put(25,72){\tiny{\color{black!10!green}{$\bullet$}}}
    \end{overpic}
\end{minipage}
\begin{minipage}{0.155\textwidth}
    \begin{overpic}[width=1\textwidth]{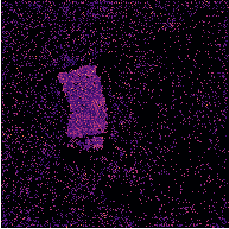}
    \end{overpic}
\end{minipage}
\hspace*{1mm}
\begin{minipage}{0.155\textwidth}
    \begin{overpic}[width=1\textwidth]{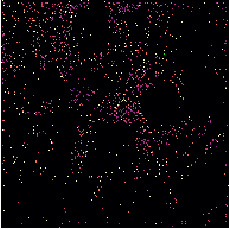}
    \put(70,75){\tiny{\color{black!10!green}{$\bullet$}}}
    \end{overpic}
\end{minipage}
\begin{minipage}{0.155\textwidth}
    \begin{overpic}[width=1\textwidth]{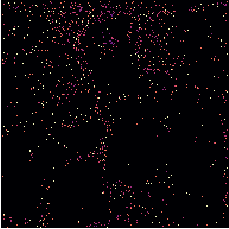}
    \end{overpic}
\end{minipage}
\vspace*{1mm}

\hspace*{0.3mm}
\begin{minipage}{0.31\textwidth}
    \centering (a) Prompt: \texttt{White small bowl}
\end{minipage}
\hspace*{1mm}
\begin{minipage}{0.31\textwidth}
    \centering (b) Prompt: \texttt{Open brown laptop}
\end{minipage}
\hspace*{1mm}
\begin{minipage}{0.31\textwidth}
    \centering (c) Prompt: \texttt{Black lens camera}
\end{minipage}
\hspace*{1mm}